\documentclass{article}

 \usepackage[preprint]{neurips_2026}

\usepackage[utf8]{inputenc} % allow utf-8 input
\usepackage[T1]{fontenc}    % use 8-bit T1 fonts
\usepackage{hyperref}       % hyperlinks
\usepackage{url}            % simple URL typesetting
\usepackage{booktabs}       % professional-quality tables
\usepackage{amsfonts}       % blackboard math symbols
\usepackage{nicefrac}       % compact symbols for 1/2, etc.
\usepackage{microtype}      % microtypography
\usepackage{xcolor}         % colors

\usepackage{microtype}
\usepackage{graphicx}
\usepackage{subcaption}
\usepackage{booktabs} % for professional tables
\usepackage{tikz}
\usepackage{arydshln}
\usepackage{paracol}
\usepackage{caption}
\usepackage{bm}
\usepackage[section]{placeins}

\usepackage{xcolor}         % colors
\usepackage{amsmath}
\usepackage{amssymb}
\usepackage{stmaryrd}
\usepackage{multirow}
\usepackage{blindtext}
\usepackage{multicol}
\usepackage{fvextra} % Needed for verbatim in multicol
\usepackage{algorithm}
\usepackage{algorithmic}
\usepackage{amsthm}
\newtheorem{definition}{Definition}

\definecolor{keywordcolor}{rgb}{0.7, 0.1, 0.1}   % red
\definecolor{tacticcolor}{rgb}{0.0, 0.1, 0.6}    % blue
\definecolor{commentcolor}{rgb}{0.4, 0.4, 0.4}   % grey
\definecolor{symbolcolor}{rgb}{0.0, 0.1, 0.6}    % blue
\definecolor{sortcolor}{rgb}{0.1, 0.5, 0.1}      % green
\definecolor{attributecolor}{rgb}{0.7, 0.1, 0.1} % red

\usepackage{listings}

\lstdefinestyle{leanfullproof}{
  language=lean,
  basicstyle=\tiny\ttfamily,
  breaklines=true,
  breakatwhitespace=false,
  columns=fullflexible,
  mathescape=false,
  literate=
    {ℚ}{{$\mathbb{Q}$}}1 {ℤ}{{$\mathbb{Z}$}}1 {ℕ}{{$\mathbb{N}$}}1
    {∀}{{$\forall$}}1 {∃}{{$\exists$}}1 {→}{{$\to$}}1 {↔}{{$\leftrightarrow$}}1
    {∨}{{$\vee$}}1 {∧}{{$\wedge$}}1 {≤}{{$\le$}}1 {≥}{{$\ge$}}1
    {⊆}{{$\subseteq$}}1 {∣}{{$\mid$}}1 {≠}{{$\ne$}}1 {≡}{{$\equiv$}}1 {⊢}{{$\vdash$}}1
    {←}{{$\leftarrow$}}1 {↑}{{$\uparrow$}}1 {▸}{{$\triangleright$}}1
    {λ}{{$\lambda$}}1 {φ}{{$\varphi$}}1 {·}{{$\cdot$}}1
    {⟨}{{$\langle$}}1 {⟩}{{$\rangle$}}1 {‹}{{$\langle$}}1 {›}{{$\rangle$}}1
    {⌈}{{$\lceil$}}1 {⌉}{{$\rceil$}}1 {⌊}{{$\lfloor$}}1 {⌋}{{$\rfloor$}}1
}

\usepackage[utf8]{inputenc}
\usepackage{amsmath, amssymb}
\usepackage{tikz}
\usetikzlibrary{shapes, arrows.meta, positioning, calc, fit, backgrounds, quotes}

\usepackage{hyperref}

\newcommand{\ImProver}{ImProver 2}
\title{ImProver 2: Iteratively Self-Improving LMs for Neurosymbolic Proof Optimization}

\author{%
Riyaz Ahuja \\
Carnegie Mellon University \\
\texttt{riyaza@andrew.cmu.edu} \\
\And
Tate Rowney \\
Carnegie Mellon University \\
\texttt{trowney@andrew.cmu.edu} \\
\And
Jeremy Avigad \\
Carnegie Mellon University \\
\texttt{avigad@andrew.cmu.edu} \\
\And
Sean Welleck \\
Carnegie Mellon University \\
\texttt{swelleck@andrew.cmu.edu}
}

\makeatletter
\def\ps@firstpagenotice{%
  \let\@mkboth\@gobbletwo
  \def\@oddhead{}%
  \def\@evenhead{}%
  \def\@oddfoot{\hfil\footnotesize\@noticestring\hfil}%
  \let\@evenfoot\@oddfoot
}
\renewcommand{\@notice}{\thispagestyle{firstpagenotice}}
\makeatother

\begin{document}

\maketitle

\begin{abstract}
Formal mathematics libraries are rapidly expanding, creating a growing need to refactor verified proofs for maintainability and to improve training data quality for neural provers. However, scalable proof optimization is hindered by heterogeneous and heuristically specified objectives, scarce data, and high training and inference costs. To overcome these challenges, we introduce \ImProver, a neurosymbolic framework for automated proof optimization in Lean 4. \ImProver \ combines a data-efficient expert-iteration pipeline with a scaffold that exposes formal structure alongside lightweight informal abstractions. We further introduce a suite of metrics capturing structural proof properties. Using \ImProver, we train a 7B-parameter model that outperforms orders-of-magnitude larger models within the same model family, and is competitive with mid-tier frontier models across metrics. We additionally demonstrate that our neurosymbolic scaffold significantly improves performance across both small and frontier models. We show that with proper scaffolding and training, small models can effectively restructure research-level proofs over complex and varied metrics, matching substantially larger systems and establishing proof optimization as a scalable, learnable task.
\end{abstract}

\section{Introduction}
Formal proof assistants such as Lean \citep{LeanTheoremProver}, Rocq \citep{RocqTheoremProver}, and Isabelle \citep{IsabelleTheoremProver} have transformed mathematical practice by making the correctness of proofs explicit and mechanized. Community libraries such as Lean's Mathlib \citep{Mathlib} now grow at a pace driven both by increasing human contributions and by recent advances in neural theorem proving and autoformalization \citep{achim2025aristotleimolevelautomatedtheorem, AlphaProof}.

This rapid expansion raises multiple concerns about issues of data quality. First, this expansion stresses the maintainability, coherence, and long-term usability of libraries due to proofs that are often heterogeneous in style and clarity \citep{Mathlib}. Second, low library quality decreases its utility as training data: modern theorem provers and autoformalizers increasingly train on these very corpora, so the structure and readability of proofs directly shape downstream prover performance \citep{gu2025proofoptimizer}.
% We corroborate this second point empirically by showing that provers trained on structure-optimized corpora outperform those trained on unoptimized data.
Unfortunately, the growth rate of formal libraries already exceeds what human reviewers and maintainers can reliably curate; and moreover, this discrepancy is only expected to widen due to increasing volumes of machine-generated proofs which, even when guaranteed to be correct, do not carry similar guarantees as to the proof's quality, modularity, or understandability \citep{SeedProver}.

This motivates automated proof optimization: given a verified proof, produce a formally correct rewrite that scores better under a user-defined objective, such as shorter length, higher modularity, or fewer explicit dependencies \citep{ImProver}. Because objectives vary by use case and formal context, a practical method must scale across arbitrary metrics and research-level theorems. Small specialized models are especially attractive for this setting because relevant data is scarce in general corpora, library-scale deployment can require millions of samples, and local open-weight models are more accessible to formalization projects.

We address these problems with \textbf{\ImProver}, a self-improving pipeline that trains small language models (\textit{SLM}s) to optimize Lean proofs under a wide-ranging class of metrics.
Our core idea is to leverage \textit{iterative preference optimization}: iterating between generating proof candidates, scoring them according to correctness and the desired optimization criterion, and learning from the resulting pairs of higher and lower scoring proofs.
% In turn, the model generates its own training data, which is
% We build upon iterative reasoning preference optimization (IRPO) \citep{IRPO}, a preference-optimization algorithm, to iteratively improve these models by generating their own training data
% via inference and metric evaluation
% on a dataset of research-level theorems (a process referred to as \textit{expert iteration}).
In particular, we extend the Iterative Reasoning Preference Optimization (IRPO)~\citep{IRPO} algorithm with
a new replay buffer that balances old and newly generated data for use in the next round of training, preventing model collapse and allowing for monotonic improvement over many rounds. We additionally give the model access to rich information from the Lean theorem proving environment, including goal states, informalized summaries, lemma context, and examples, which we term \textit{neurosymbolic augmentation}.
We use \textbf{\ImProver} to train models for three different metrics: the \textit{length} of the proof, \textit{modularity} (the ability to divide the proof into a series of smaller lemmas), and the explicit \textit{dependencies} used by the theorem, and show that our trained models can substantially improve over their base model and remain competitive with much larger unscaffolded systems on research-level theorems.
In summary, our contributions are:

\begin{enumerate}
    \item \emph{Self-improving SLMs for proof refactoring.} We show that iterative preference optimization can bootstrap small language models on proof optimization in specialized research libraries, making them competitive with much larger models in several unscaffolded comparisons.

    \item \emph{Structural optimization metrics.} Beyond proof length~\citep{ImProver,gu2025proofoptimizer}, we study \textit{modularity} and \textit{dependency} metrics that leverage proof structure and the surrounding library.
    % We define a suite of formal metrics (length, modularity/declarativity via spawned-goal counts, dependency footprint) \iffalse and a stable informal metric (LLM-as-judge readability)\fi designed for reliable, pairwise evaluation under randomization and consensus.
    These metrics were chosen with formal mathematics experts and target distinct proof-refactoring objectives.
    % \item \emph{A highly stable iterative RL pipeline.} We present a robust expert-iteration loop utilizing IRPO (DPO with explicit NLL regularization), enabling data-free self-improvement with strong stability and sample efficiency.
    \item \emph{Neurosymbolic augmentation for research mathematics.} We extract relevant lemmas or definitions, goal-state traces, and automatic informalizations of target proofs. This augmentation boosts both small and large models on proof optimization.
\end{enumerate}
We additionally open-source our code and data\footnote{\href{https://github.com/riyazahuja/improver}{Github}}.

\section{Related Work}

Interest in neurosymbolic theorem proving--the use of deep learning to create or manipulate verified mathematical proofs in languages such as Lean 4 \citep{LeanTheoremProver}--has seen significant advancements in recent years \citep{lu-etal-2023-survey, Li-etal-survey}. In particular, much research has focused on generating formal proofs given their statements \citep{GPT-f}, with recent systems achieving high performance on nontrivial benchmarks and internationally renowned mathematics competitions \citep{AlphaProof, achim2025aristotleimolevelautomatedtheorem, SeedProver}. Many systems additionally utilize neurosymbolic augmentation, providing a generative prover model with information gathered from within the proof environment \citep{leandojo, ImProver, GodelProver}. However, both formal and informal proofs generated by current LLM-based systems often suffer from stylistic irregularities even when they are sound, including redundant steps or a structure which does not clearly represent the broader logical argument \citep{frieder2025datamathematicalcopilotsbetter}.

% JA: "numerous stylistic issues": I deleted the word "numerous". I am not sure what "stylistic issues" means; is it the best description?
% (Tate) I added a bit more description about the issues they mention in the paper

Previous work by \citep{ImProver, gu2025proofoptimizer} has attempted to rectify these issues by creating LLM-based agents to refactor formal proofs. \citet{ImProver} created a system capable of optimizing towards multiple metrics of improvement; however, it relied on general-purpose closed-source models, leading to substantial deployment costs and limited ability to improve performance beyond these models' baseline. \citet{gu2025proofoptimizer} focused solely on optimizing the token count of proofs according to a complex tokenizer intended to reduce the time required to compile them; they do not examine other metrics, leaving out important use cases and limiting its utility to research mathematicians. Furthermore, the works above do not fully utilize the information available through working in an interactive theorem proving environment via goal-state extraction \citep{GPT-f}, premise retrieval \citep{leandojo}, or auto-informalization \citep{autoinformalization}; both of the above overlook at least one of these aspects.

\section{Proof Optimization}
\label{sec:proof_optimization}

Given a verified proof, a proof optimization agent synthesizes a semantically equivalent proof that is ``better'' under a user-specified objective while remaining correct according to the Lean kernel. We follow the setup of \citet{ImProver, gu2025proofoptimizer} and introduce two additional structural metrics: \textit{modularity} and \textit{dependencies}.

\begin{figure*}[tbp]
% Mathlib.Order.Atoms
% SetLike.isCoatom_iff
% 5.0
% 2.0
% -3.0
% \setlength{\columnsep}{0.2cm}
\begin{minipage}[t]{0.33\textwidth}
\textbf{Original}
\begin{lstlisting}[basicstyle=\tiny\ttfamily]

theorem isCoatom_iff [OrderTop A] {K : A} :
    IsCoatom K ↔ K ≠ T ∧ ∀ H g, K ≤ H → g ∉ K → g ∈ H → H = T := by
  simp_rw [IsCoatom, lt_iff_le_not_le, SetLike.not_le_iff_exists,
    and_comm (a := _ ≤ _), and_imp, exists_imp, ← and_imp, and_comm]
\end{lstlisting}

\end{minipage}\begin{minipage}[t]{0.33\textwidth}
\textbf{Original}
\begin{lstlisting}[basicstyle=\tiny\ttfamily]
theorem mem_cross_iff (x y : TSet γ) :
    ∀ a, a ∈' cross hβ hγ hδ x y ↔ ∃ b c, a = ⟨b, c⟩' ∧ b ∈' x ∧ c ∈' y := by
  intro a
  rw [cross, mem_inter_iff, vCross_spec]
  constructor
  · rintro ⟨h₁, b, c, rfl, h₂⟩
    simp only [op_mem_converse_iff, vCross_spec, op_inj] at h₁
    obtain ⟨b', c', ⟨rfl, rfl⟩, h₁⟩ := h₁
    exact ⟨b, c, rfl, h₁, h₂⟩
  · rintro ⟨b, c, rfl, h₁, h₂⟩
    simp only [op_mem_converse_iff, vCross_spec, op_inj]
    exact ⟨⟨c, b, ⟨rfl, rfl⟩, h₁⟩, ⟨b, c, ⟨rfl, rfl⟩, h₂⟩⟩
\end{lstlisting}
\end{minipage}\begin{minipage}[t]{0.33\textwidth}
\textbf{Original}
\begin{lstlisting}[basicstyle=\tiny\ttfamily,literate={≤ₛ}{{$\leq_s$}}2 {⊆}{{$\subseteq$}}1 {α}{{$\alpha$}}1 {Phi}{{$\varphi$}}1 {→}{{$\rightarrow$}}1 {₁}{{$_1$}}1 {₂}{{$_2$}}1 {dollar}{{$ \$ $}}1]
lemma KD_weakerThan_KDB : (Hilbert.KD α) ≤ₛ (Hilbert.KDB α) := normal_weakerThan_of_subset dollar by intro; aesop;
\end{lstlisting}
\end{minipage}

\begin{minipage}[t]{0.33\textwidth}
\textbf{Optimized (Dependency)}
\begin{lstlisting}[basicstyle=\tiny\ttfamily]
theorem isCoatom_iff [OrderTop A] {K : A} :
    IsCoatom K ↔ K ≠ T ∧ ∀ H g, K ≤ H → g ∉ K → g ∈ H → H = T := by
  constructor <;> intro h
  <;> simp_all [IsCoatom, lt_iff_le_not_le, SetLike.not_le_iff_exists]
  <;> tauto
\end{lstlisting}
\end{minipage}\begin{minipage}[t]{0.33\textwidth}
\textbf{Optimized (Length)}
\begin{lstlisting}[basicstyle=\tiny\ttfamily]
theorem mem_cross_iff (x y : TSet γ) :
    ∀ a, a ∈' cross hβ hγ hδ x y ↔ ∃ b c, a = ⟨b, c⟩' ∧ b ∈' x ∧ c ∈' y := by
  simp_all [cross, mem_inter_iff, vCross_spec,
    op_mem_converse_iff, op_inj]
  <;> aesop
\end{lstlisting}
\end{minipage}
\begin{minipage}[t]{0.33\textwidth}
\textbf{Optimized (Modularity)}
\begin{lstlisting}[basicstyle=\tiny\ttfamily,literate={≤ₛ}{{$\leq_s$}}2 {⊆}{{$\subseteq$}}1 {α}{{$\alpha$}}1 {Phi}{{$\varphi$}}1 {→}{{$\rightarrow$}}1 {₁}{{$_1$}}1 {₂}{{$_2$}}1 {·}{{$\cdot$}}1]
lemma KD_weakerThan_KDB : (Hilbert.KD α) ≤ₛ (Hilbert.KDB α) := by
  have h₁ : (LO.Modal.Hilbert.KD α).axioms ⊆ (LO.Modal.Hilbert.KDB α).axioms → (Hilbert.KD α) ≤ₛ (Hilbert.KDB α) := by
    intro h
    apply normal_weakerThan_of_subset
    apply h
  have h₂ : (LO.Modal.Hilbert.KD α).axioms ⊆ (LO.Modal.Hilbert.KDB α).axioms := by
    intro Phi hPhi
    cases' hPhi with hPhi hPhi
    · simp_all [LO.Modal.Hilbert.KD]
    · simp_all [LO.Modal.Hilbert.KDB]
  exact h₁ h₂
\end{lstlisting}
\end{minipage}
\captionof{figure}{\ImProver \ automatically optimizes human-written proofs to reduce explicit dependencies, minimize length, or maximize proof modularity, while maintaining formal correctness. }
\label{fig:main-demo}
\end{figure*}

\subsection{Setup and notation}
\label{sec:setup_notation}
Let $\mathcal{C}$ denote proof contexts (imports, local declarations, module metadata, etc.), $\mathcal{X}$ theorem statements, and $\mathcal{Y}$ proofs. We consider $(c,x,y)\in \mathcal{C}\times\mathcal{X}\times\mathcal{Y}$, where $y$ is a purported proof of $x$ in $c$ which may or may not be sound.

We define a \textit{verifier} as a computable function
\[
\text{v}(c,x,y) : \mathcal{C} \times \mathcal{X} \times \mathcal{Y} \to  \{0,1\},
\]
This function outputs $1$ if $y$ is a syntactically correct and sound proof of $x$ in $c$. We also define
\[
\mathcal{F}(c,x)=\{y\in\mathcal{Y}:\text{v}(c,x,y)=1\}.
\]
Two proofs $y,y'\in\mathcal{F}(c,x)$ are said to be \emph{semantically equivalent}.

We use the Lean 4 language \citep{LeanTheoremProver} as our verifier; the language's kernel/type-checker provides a strong guarantee of correspondence with a type-theoretic model of modern mathematics.

\subsection{Optimization Objective}
\label{sec:metrics}

Two semantically equivalent proofs may have significant syntactic differences, and moreover certain characteristics may make them more or less desirable for use in practice. To quantify this, we define an \textit{optimization metric} as a computable function
\[
\mu:\mathcal{C}\times\mathcal{X}\times\mathcal{Y}\to\mathbb{R},
\]
Given an initial theorem and proof $(c, x, y_0)$, we aim to find a verifiably correct proof that maximizes metric score:
\begin{equation}
\label{eq:po_score}
{\arg\max}_{\substack{y \in \mathcal{Y}\\v(c, x, y)=1}} \quad \mu(c, x, y)
\end{equation}

In practice, we approximate this via language model-based Lean 4 code generation: by independently generating multiple variants, we may pick the one with greatest improvement, if it surpasses the original.

\subsubsection{Metrics of Interest}
   \newcommand{\Unassigned}{\text{Unassigned}}
   \newcommand{\dom}{\text{dom}}
   \newcommand{\inst}{\text{inst}}
   \newcommand{\Children}{\text{Children}}
   \newcommand{\Step}{\text{Step}}
   \newcommand{\goalbefore}{\text{goalsbefore}}
   \newcommand{\goalsafter}{\text{goalsafter}}
   \newcommand{\Deps}{\text{Deps}}
   \newcommand{\Current}{\text{Current}}
   \newcommand{\Spawned}{\text{Spawned}}
   \newcommand{\prodmap}{\text{prodmap}}
   \newcommand{\F}{\mathcal{F}}
   \newcommand{\A}{\mathcal{A}}
   \newcommand{\gb}{\bar{g}}

In this work, we focus and evaluate on a collection of three metrics, which are designed to be practical and interpretable structural objectives for formal proof optimization. These metrics are proxies rather than reviewer-validated measures of subjective proof quality: they support automatic evaluation at scale, but they do not by themselves establish human maintainer preference or downstream theorem-prover utility.
\begin{itemize}
   \item \textbf{Length:} We aim to minimize the length of proofs, measured by the number of tactics used. In practice, proof shortening (or ``golfing") is a common activity in formal mathematics \citep{Mathlib}, as shorter proofs are often easier to read and maintain, and reduce overhead during compilation. As such, we define the length metric $\mu_{len}$ as the negative of the number of tactics.

   \item \textbf{Dependencies:} We aim to minimize the explicit dependency footprint of proofs, measured by the number of unique external lemmas explicitly named in the proof. This metric does not measure semantic independence from the library: a proof using \texttt{simp} or \texttt{omega} may still rely on many facts internally. This encourages self-contained proofs that do not require the use or memorization of large numbers of dependency names, improving maintainability.\footnote{We would like to thank Dr. Heather Macbeth of Imperial College London for reaching out to suggest the idea behind this metric.}.

   More specifically, given a theorem and proof $(c, x, y)$, we compute $\text{Deps}_{c, x, y}$, the set of all theorems explicitly named in the proof $y$ (see Section \ref{app:context_extraction}). The metric is then defined as $\mu_{dep}(c, x, y) := -\left|\text{Deps}_{c, x, y}\right|$.

   \item \textbf{Modularity:} We aim to maximize the \textit{modularity} of proofs, which is intuitively understood to be the number of independent subproofs in our proof. This is a structural objective motivated by proof decomposition, lemma-generation workflows, and maintainable proof organization.

   % In addition, we note that such proofs -- which break down complex arguments into smaller, reusable subproofs -- are more similar to the style of recent autoformalized and machine-generated proofs, which often generate proofs in a lemma-first manner. As such, we hypothesize that optimizing for modularity will improve the utility of proofs as training data for downstream theorem provers and autoformalizers and hope to explore this empirically in future work.
   To quantify modularity, we postprocess Lean proofs to deconstruct them into a tree of tactics, with edges representing the logical dependencies and flow between tactics in the proof. In this tree, we mark a subset of edges as ``spawned" if they correspond to goals that are introduced by the proof but are not direct subgoals of the current tactic, which naturally arise from tactics like \texttt{have}, \texttt{calc}, etc., and correspond with the informal notion of "modular" independent subproofs.

    We therefore define the modularity metric as
    $\mu_{mod}(c,x,y) = |\{\text{effective spawned goals in } y\}|$.
    The designation of which spawned goals are non-trivial and ``effective" requires the use of several heuristics; a complete definition is provided in Appendix \ref{app:modularity_metric}. These safeguards reduce trivial goal-spawning and duplicate-wrapper artifacts, but they do not eliminate all stylistically debatable decompositions.

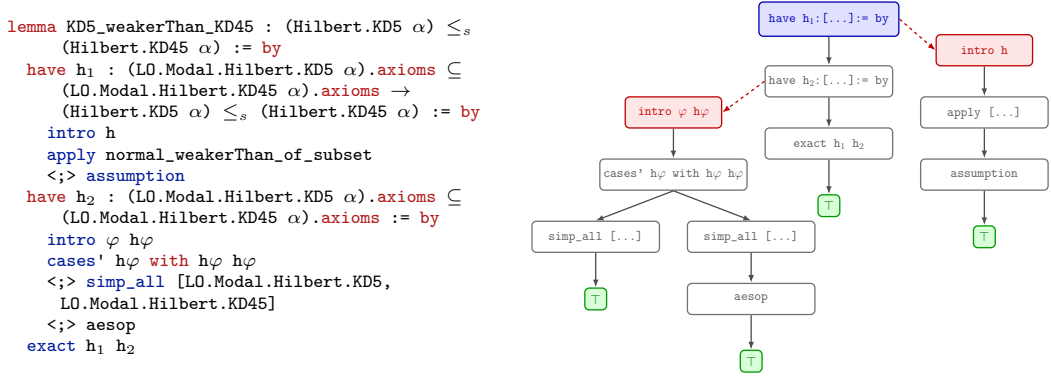
\begin{figure}[ht]
    \centering
\textbf{Example Proof Tree}

\begin{minipage}[t]{0.49\textwidth}
\vspace{0pt}
    \begin{scriptsize}
    \begin{lstlisting}[basicstyle=\scriptsize\ttfamily,literate={≤ₛ}{{$\leq_s$}}2 {⊆}{{$\subseteq$}}1 {α}{{$\alpha$}}1 {Phi}{{$\varphi$}}1 {→}{{$\rightarrow$}}1 {₁}{{$_1$}}1 {₂}{{$_2$}}1]
lemma KD5_weakerThan_KD45 : (Hilbert.KD5 α) ≤ₛ (Hilbert.KD45 α) := by
  have h₁ : (LO.Modal.Hilbert.KD5 α).axioms ⊆ (LO.Modal.Hilbert.KD45 α).axioms → (Hilbert.KD5 α) ≤ₛ (Hilbert.KD45 α) := by
    intro h
    apply normal_weakerThan_of_subset
    <;> assumption
  have h₂ : (LO.Modal.Hilbert.KD5 α).axioms ⊆ (LO.Modal.Hilbert.KD45 α).axioms := by
    intro Phi hPhi
    cases' hPhi with hPhi hPhi
    <;> simp_all [LO.Modal.Hilbert.KD5, LO.Modal.Hilbert.KD45]
    <;> aesop
  exact h₁ h₂
    \end{lstlisting}
    \end{scriptsize}

  \end{minipage}
  \begin{minipage}[t]{0.49\textwidth}
\vspace{0pt}

    \resizebox{\columnwidth}{!}{%
\begin{tikzpicture}[
    font=\sffamily,
    >=Latex,
    % every node/.style={inner sep=0pt}
]

% Styles
\tikzset{
  rootbox/.style={
    draw=blue!70!black,
    fill=blue!12,
    rounded corners=4pt,
    very thick,
    minimum width=4.5cm,
    minimum height=1.1cm,
    text=blue!70!black,
    font=\large,%\bfseries\large,
    align=center
  },
  normalbox/.style={
    draw=black!55,
    fill=white,
    rounded corners=4pt,
    very thick,
    minimum width=4.1cm,
    minimum height=1.0cm,
    text=black!65,
    font=\large,%\bfseries\large,
    align=center
  },
  effectivebox/.style={
    draw=red!75!black,
    fill=red!10,
    rounded corners=4pt,
    very thick,
    minimum width=3.1cm,
    minimum height=1.0cm,
    text=red!75!black,
    font=\large,%\bfseries\large,
    align=center
  },
  termbox/.style={
    draw=green!60!black,
    fill=green!15,
    rounded corners=4pt,
    very thick,
    minimum width=0.7cm,
    minimum height=0.7cm,
    text=green!40!black,
    font=\large
  },
  solidarrow/.style={
    draw=black!70,
    very thick,
    -{Latex[length=2.5mm]}
  },
  dashedarrow/.style={
    draw=red!75!black,
    dashed,
    very thick,
    -{Latex[length=2.5mm]}
  },
  idlabel/.style={
    font=\scriptsize,
    text=black!45
  }
}

% Nodes
\node[rootbox]      (n0)  at (0,0)        {\texttt{have h$_1$:[\ldots]:= by}};
\node[effectivebox] (n1)  at (5,-1)   {\texttt{intro h}};

\node[normalbox]    (n4)  at (0,-2)     {\texttt{have h$_2$:[\ldots]:= by}};
\node[effectivebox] (n5)  at (-5,-3)  {\texttt{intro $\varphi$ h$\varphi$}};

\node[normalbox]    (n2)  at (5,-3)   {\texttt{apply [\ldots]}};
\node[normalbox]    (n3)  at (5,-5)   {\texttt{assumption}};

\node[normalbox]    (n10) at (0,-4)    {\texttt{exact h$_1$ h$_2$}};

\node[normalbox]    (n6)  at (-5,-5) {\texttt{cases' h$\varphi$ with h$\varphi$ h$\varphi$}};
\node[normalbox]    (n7)  at (-7.5,-7)  {\texttt{simp\_all [\ldots]}};
\node[normalbox]    (n8)  at (-2.5,-7)  {\texttt{simp\_all [\ldots]}};
\node[normalbox]    (n9)  at (-2.5,-9)  {\texttt{aesop}};

\node[termbox]      (t3)  at (5,-7)   {$\top$};
\node[termbox]      (t10) at (0,-6)    {$\top$};
\node[termbox]      (t7)  at (-7.5,-9)  {$\top$};
\node[termbox]      (t9)  at (-2.5,-11) {$\top$};

% Edges
\draw[solidarrow]  (n0.south) -- (n4.north);
\draw[dashedarrow] (n0.east) -- (n1.west);

\draw[solidarrow]  (n1.south) -- (n2.north);
\draw[solidarrow]  (n2.south) -- (n3.north);
\draw[solidarrow]  (n3.south) -- (t3.north);

\draw[dashedarrow]  (n4.west) -- (n5.east);
\draw[solidarrow]  (n4.south) -- (n10.north);
\draw[solidarrow]  (n10.south) -- (t10.north);

\draw[solidarrow]  (n5.south) -- (n6.north);
\draw[solidarrow]  (n6.south) -- (n7.north);
\draw[solidarrow]  (n6.south) -- (n8.north);
\draw[solidarrow]  (n7.south) -- (t7.north);
\draw[solidarrow]  (n8.south) -- (n9.north);
\draw[solidarrow]  (n9.south) -- (t9.north);

\end{tikzpicture}%
}
\end{minipage}

    \caption{Example Lean proof (top) and its generated proof tree (bottom). Note the root node (in blue), and the two effective spawned goals coming from $h_1$ and $h_2$ (in red).}
    \label{fig:kd5-kd45-proof}
\end{figure}

\end{itemize}

\section{\ImProver}
\label{improver2}
We present \ImProver, a training pipeline that bootstraps small language models (\textit{SLM}s) for a target proof-optimization metric. It combines IRPO training (\ref{sec:irpo}), a replay buffer balancing old and newly generated data (\ref{sec:training_replay}), and neurosymbolic augmentation for generation (\ref{sec:generation}).

% --- Color Palette ---
\definecolor{cGen}{RGB}{50, 50, 120}      % Deep Navy
\definecolor{cPrompt}{RGB}{220, 220, 220} % Light Gray
\definecolor{cBuf}{RGB}{80, 160, 140}     % Teal

% Data Flow Gradient
\definecolor{cDtrain}{RGB}{100, 149, 237} % Cornflower Blue
\definecolor{cDre}{RGB}{138, 110, 215}    % Medium Purple
\definecolor{cDfil}{RGB}{186, 85, 211}    % Medium Orchid
\definecolor{cDirpo}{RGB}{219, 112, 147}  % Pale Violet Red

% Score Color
\definecolor{cScore}{RGB}{255, 235, 175}  % Pastel Gold/Yellow

% Backgrounds
\definecolor{bgInf}{RGB}{240, 248, 255}   % Light Blue
\definecolor{bgBuf}{RGB}{235, 250, 245}   % Light Teal
\definecolor{bgShape}{RGB}{248, 240, 252} % Light Lavender
\definecolor{bgTrain}{RGB}{255, 245, 240} % Light Orange/Peach

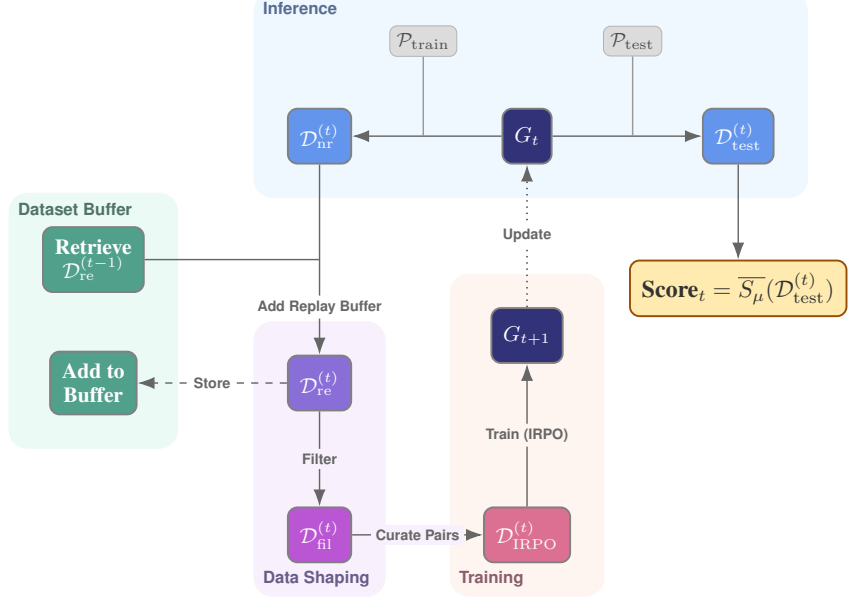
\begin{figure*}[t]
\centering
\pgfdeclarelayer{bg}
\pgfsetlayers{bg,main}

\resizebox{0.8\linewidth}{!}{%
\begin{tikzpicture}[
    font=\small\sffamily,
    node distance=14mm and 22mm,
    >={Latex[length=2.5mm, width=1.8mm]},
    % --- Styles ---
    % Main Node Style
    box/.style={
        draw=black!50,
        thick,
        rounded corners=4pt,
        inner sep=5pt,
        align=center,
        text=white,
        font=\small\bfseries,
        minimum height=2.3em,
        % drop shadow={opacity=0.15, shadow xshift=1pt, shadow yshift=-1pt}
    },
    % Input/Prompt Style
    pbox/.style={
        draw=black!30,
        fill=cPrompt,
        text=black!70,
        rounded corners=3pt,
        inner sep=3pt,
        font=\footnotesize
    },
    % Score Box Style
    sbox/.style={
        draw=orange!60!black, 
        thick,
        fill=cScore,
        text=black!80,
        rounded corners=4pt,
        inner sep=5pt,
        font=\bfseries
    },
    % Arrow Style
    conn/.style={
        ->,
        semithick,
        draw=black!60,
        rounded corners=5pt
    },
    % Tag Style (Default)
    tag/.style={
        font=\tiny\sffamily\bfseries,
        text=black!60,
        inner sep=2pt,
        rounded corners=2pt
    },
    % Background Regions
    region/.style={
        inner sep=14pt,
        rounded corners=8pt,
        draw=none
    },
    % Section Labels
    lbl/.style={
        font=\scriptsize\bfseries\sffamily,
        text opacity=0.8,
        inner sep=4pt
    }
]

% ---------------------------------------------------------
% 1. NODES
% ---------------------------------------------------------

% Center Column
\node[box, fill=cGen] (Gt) {$G_t$};

% Inference Section
\node[box, fill=cDtrain, left=of Gt] (Dtrain) {$\mathcal{D}_{\mathrm{nr}}^{(t)}$};
\node[box, fill=cDtrain, right=of Gt] (Dtest) {$\mathcal{D}_{\mathrm{test}}^{(t)}$};

% Score
\node[sbox, below=1.4cm of Dtest] (score) 
    {Score$_t$ $= \overline{S_\mu}(\mathcal{D}^{(t)}_{\mathrm{test}})$};

% Prompts
\node[pbox] (Ptrain) at ($(Dtrain)!0.5!(Gt) + (0, 1.4cm)$) {$\mathcal{P}_{\mathrm{train}}$};
\node[pbox] (Ptest)  at ($(Gt)!0.5!(Dtest) + (0, 1.4cm)$) {$\mathcal{P}_{\mathrm{test}}$};

% Data Shaping (Vertical)
\node[box, fill=cDre, below=2.8cm of Dtrain] (Dre) {$\mathcal{D}_{\mathrm{re}}^{(t)}$};
\node[box, fill=cDfil, below=of Dre] (Dfil) {$\mathcal{D}_{\mathrm{fil}}^{(t)}$};

% Buffer (Aligned Horizontally)
\node[box, fill=cBuf, left=2.2cm of Dre] (add) {Add to\\Buffer};

% Calculate intersection point for the horizontal line
\coordinate (midFlow) at ($(Dtrain.south)!0.5!(Dre.north)$);
\node[box, fill=cBuf] (retrieve) at (add |- midFlow) {Retrieve\\$\mathcal{D}_{\mathrm{re}}^{(t-1)}$};

% Training Section
\node[box, fill=cDirpo] (Dirpo) at (Gt |- Dfil) {$\mathcal{D}_{\mathrm{IRPO}}^{(t)}$};
\node[box, fill=cGen] (Gt1)   at ($(Dirpo)!0.5!(Gt)$) {$G_{t+1}$};

% ---------------------------------------------------------
% 2. CONNECTIONS & LABELS
% ---------------------------------------------------------

% Gt -> Dtrain (No Tag)
\draw[conn] (Gt) -- (Dtrain);

% Gt -> Dtest (No Tag)
\draw[conn] (Gt) -- (Dtest);

% Dtest -> Score (No Tag)
\draw[conn] (Dtest) -- (score);

% Prompts Merge
\draw[semithick, black!40] (Ptrain) -- ($(Dtrain)!0.5!(Gt)$);
\draw[semithick, black!40] (Ptest)  -- ($(Gt)!0.5!(Dtest)$);

% Dtrain -> Dre (With Add Replay Buffer Tag)
% The tag uses fill=white because this gap is white (outside the colored boxes)
% Positioned at 0.75 so it is strictly below the intersection point
\draw[conn] (Dtrain) -- node[tag, fill=white, pos=0.75] {Add Replay Buffer} (Dre);

% Retrieve -> Flow (Horizontal Line)
\draw[semithick, black!60] (retrieve.east) -- (midFlow);

% Dre -> Dfil (Filter)
% Uses fill=bgShape so it blends into the purple background
\draw[conn] (Dre) -- node[tag, fill=bgShape] {Filter} (Dfil);

% Dre -> Add (Store)
% Uses fill=white as it crosses the whitespace gap mostly
\draw[conn, dashed] (Dre) -- node[tag, fill=white] {Store} (add);

% Dfil -> Dirpo (Curate Pairs)
% Located in the gap between sections, so fill=white
\draw[conn] (Dfil) -- node[tag, fill=bgShape] {Curate Pairs} (Dirpo);

% Dirpo -> Gt+1 (Train)
% Inside the Orange box, so uses fill=bgTrain
\draw[conn] (Dirpo) -- node[tag, fill=bgTrain] {Train (IRPO)} (Gt1);

% Gt+1 -> Gt (Update)
% In whitespace, fill=white
\draw[conn, dotted, thick] (Gt1) -- node[tag, fill=white] {Update} (Gt);

% ---------------------------------------------------------
% 3. BACKGROUNDS
% ---------------------------------------------------------
\begin{pgfonlayer}{bg}
    % Inference
    \node[region, fill=bgInf, fit=(Gt)(Dtrain)(Dtest)(Ptrain)(Ptest)] (infBox) {};
    \node[lbl, text=cDtrain!50!black, anchor=north west] at (infBox.north west) {Inference};
    
    % Buffer
    \node[region, fill=bgBuf, fit=(retrieve)(add)] (bufBox) {};
    \node[lbl, text=cBuf!50!black, anchor=north west] at (bufBox.north west) {Dataset Buffer};
    
    % Data Shaping
    \node[region, fill=bgShape, fit=(Dre)(Dfil)] (shapeBox) {};
    \node[lbl, text=cDre!50!black, anchor=south west] at (shapeBox.south west) {Data Shaping};
    
    % Training
    \node[region, fill=bgTrain, fit=(Dirpo)(Gt1)] (trainBox) {};
    \node[lbl, text=cDirpo!50!black, anchor=south west] at (trainBox.south west) {Training};
\end{pgfonlayer}

\end{tikzpicture}
}
\caption{\textbf{ImProver\textsuperscript{2} training loop.} The diagram illustrates the iterative process of generation, retrieval, filtering, and training. Node colors represent the evolution of data from initial sampling (Blue) through processing (Purple) to training (Magenta).}
\label{fig:flow_pipeline}
\end{figure*}

\subsection{Overview}
\label{sec:improver_spec}

Given an un-modified base language model $G_0$, at the $t$-th iteration \ImProver's core loop aims to train the model $G_{t+1}$ from $G_t$ as follows (also depicted in Figure \ref{fig:flow_pipeline}):
\begin{enumerate}
    \item For some budget hyperparameter $n \in \mathbb{N}$, generate $n$ candidate proofs per problem in the training set using the current model $G_{t}$ (\ref{sec:generation}), providing the model with neurosymbolic augmentation (\ref{sec:ns_aug}) and a description of the metric to assist in generation. These new potential proofs form a new dataset (denoted $\mathcal{D}^{(t)}_{\mathrm{nr}}$).

    \item The previous iteration's dataset, $\mathcal{D}^{(t-1)}_{\mathrm{re}}$ (our \textit{replay buffer}), is interleaved with $\mathcal{D}^{(t)}_{\mathrm{nr}}$ to get $\mathcal{D}^{(t)}_{\mathrm{re}}$ (see \ref{sec:training_replay}).

    \item The model $G_t$ is trained to obtain $G_{t+1}$ (\ref{sec:irpo}). Our reinforcement learning policy of choice, known as Iterative Reasoning Preference Optimization or IRPO \citep{IRPO}, relies on preference pairs of desirable/undesirable proofs, so $\mathcal{D}^{(t)}_{\mathrm{re}}$ is filtered to remove low-quality solutions and pairs are created to form $\mathcal{D}^{(t)}_{\mathrm{IRPO}}$ (again described in \ref{sec:training_replay}).

    \item $G_{t+1}$ is evaluated with $n$ samples on the test set, again similarly to \ref{sec:generation}. The improvement in metric score of the new proofs over the originals is calculated.
\end{enumerate}

The loop is repeated until convergence of the average improvement score, or exhaustion of the compute budget.

\subsection{Generation}
\label{sec:generation}

A central feature of \ImProver \ is self-generation of training data: at each round, the current model $G^{(t)}$ receives the theorem statement, original proof, target metric, and proof-environment context, then samples $n$ candidate rewrites per theorem.

\subsubsection{Neurosymbolic Augmentation}

\label{sec:ns_aug}
Formal proof environments provide substantial opportunities to obtain relevant information about a proof. We prompt our language model with additional neurosymbolic context that exposes the structure and dependencies of a problem at both a formal and informal level. This augmentation comes from three sources: a context slice to find relevant lemmas or definitions, goal-state traces to highlight the exact effect of each tactic on the progress of the proof, and auto-informalization to provide a higher-level natural language description of the proof in question. Each of these sources is described in detail below.

% Explicitly, this augmentation -- as modelled by the map $\Psi$ -- extracts and serializes three additional channels of information from the raw input triplet $I=(c,x,y_0)$:
% \[
% \Psi:\ I\ \mapsto\ \Big(I,\underbrace{\Psi_{\text{ctx}}(I)}_{\text{context slice}},\ \underbrace{\Psi_{\text{cos}}(I)}_{\text{goal-state traces}},\ \underbrace{\Psi_{\text{inf}}(I)}_{\text{auto-informalization}}\Big).
% \]
% The generator (LLM) sees only the serialization of $(\Psi,\mu)$, so its conditional distribution effectively factors through these channels. Now, we describe each channel in turn.

\paragraph{Context}
\label{sec:context_extraction}
When working with proofs in high-dependency environments, it is likely that a given proof $y_0$ relies on many lemmas, definitions, and other formal objects defined in the context $c$. We aim to extract and serialize a minimal set of these objects to better inform our generation process: the signatures of all definitions and theorems that are directly referenced by name in the theorem statement $x$ or the original proof $y_0$ are collected and provided provided to the model, along with any associated documentation comments. The process of collecting and filtering these is outlined in Appendix \ref{app:context_extraction}.

\paragraph{Chain-of-States (CoS)}
Chain-of-states prompting \citep{ImProver} provides a language model with the explicit state and remaining goals of a proof following the application of each tactic/step, providing richer information about the proof's structure than is normally available. We utilize Lean's \texttt{InfoTree} structures to obtain and serialize these states, and interleave them into the original proof as comments. This process is described formally in Appendix \ref{app:annotation_cos}.

\paragraph{Auto-informalization}
Utilizing natural language to guide the generation of formal proofs has been shown to significantly increase the capabilities of LLM-based systems on formal mathematical tasks \citep{DraftSketchProve}. We therefore expose natural-language sketches of each target proof to the model, providing a fuzzy layer of abstraction that captures the ``meaning'' of formal items while being robust to syntactic variation and surface-level noise.

More concretely, we prompt a language model using the proof's chain-of-states information as described above to translate a Lean proof into natural language by explaining the effect of each tactic on the proof state and providing this to the proof optimizer model. We emphasize that this informalization is a secondary channel for the generator and serves simply as an additional representation of the target theorem; outputs are still prompted to be generated formally, and correctness is still judged formally.

\subsection{Training}
\label{sec:training}

% JA: I changed "firstly" to "first".

We adapt IRPO \citep{IRPO} for proof optimization by ranking candidates using both correctness and metric improvement, yielding denser preference signals than binary success alone. We also mix new and old samples through replay, and train only on the final proof output rather than the reasoning trace.

\subsubsection{Datasets and Replay Buffer}
\label{sec:training_replay}
Indiscriminate self-training can collapse model output distributions \citep{modelCollapse}. Our replay buffer filters new samples, combines them with existing data, and sorts them by improvement before training.

% Specifically, following the generation of samples $\mathcal{D}_\text{nr}^{(t)}$ at iteration $t$ (see \ref{sec:generation}), we utilize this and the previous iteration's dataset $\mathcal{D}_\text{re}^{(t-1)}$ to obtain the new dataset $\mathcal{D}_\text{re}^{(t)}$ via the following steps:
% \begin{enumerate}
%     \item Filter $\mathcal{D}_\text{nr}^{(t)}$ for ``replay-eligible" examples (those that contain improved, compiling solutions from previous iterations).
%     \item For each proof in the filtered set, find its counterpart in $\mathcal{D}_\text{re}^{(t-1)}$, and combine the sets of proof candidates from each one, creating our new dataset $\mathcal{D}_\text{re}^{(t)}$.
%     \item Filter $\mathcal{D}_\text{re}^{(t)}$ by removing problems that had a high overall improvement rate (i.e. those that represent trivial tasks for the model).

%     \item For each theorem $T$ in $\mathcal{D}_\text{re}^{(t)}$, partition its set of proofs into a ``winners" or ``losers" category (denoted $W_T^{(t)}$ and $L_T^{(t)})$, with ``winners" being those which compile and have a sufficiently high improvement score, and ``losers" containing all other proofs. We call this filtered and partitioned dataset $\mathcal{D}_\text{fil}^{(t)}$.
% \end{enumerate}
After generating $\mathcal{D}_\text{nr}^{(t)}$, we combine candidates for each eligible problem with prior candidates, remove trivially easy problems, and partition proofs into ``winners'' (compiling, high-improvement proofs) and ``losers'' (all others). Appendix \ref{app:replay_buffer} gives details.

\paragraph{IRPO Dataset}
We form two preference-pair types: winner--winner pairs ordered by metric improvement, and winner--loser pairs preferring valid improving proofs over failures. The construction and hyperparameters $W,L$ appear in Algorithm~\ref{alg:pref_pair_creation}.

\subsubsection{IRPO (Iterative Reasoning Preference Optimization)}
\label{sec:irpo}

With this dataset $\mathcal{D}_{\text{IRPO}}^{(t)}$, we calculate the IRPO loss on each item $T$ as the (weighted) sum of the DPO loss of the preference pair and the negative log-likelihood (NLL) loss over the winner:
\begin{align*}
\mathcal{L}_{\text{IRPO}}(T) =& \mathcal{L}_{\text{DPO}}(y_{T,\ell},y_{T,w} \mid \Psi(c_T,x_T,y_{T,0}),\mu) \\&+ \alpha\mathcal{L}_{\text{NLL}}(y_{T,\ell},y_{T,w} \mid \Psi(c_T,x_T,y_{T,0}),\mu)
\end{align*}

In the above, we represent the relevant neurosymbolic augmentation with the function $\Psi$. After training with respect to this objective for one epoch, we obtain $G_{t+1}$.

\section{Experiments}
\label{experiments}

\subsection{Setup}
\label{sec:exp_setup}

We evaluate \ImProver\ on all three metrics using public research-level
mathematics repositories, benchmarking against open-source and closed-source
baselines at varying parameter counts.

\paragraph{Dataset and split}

We utilize Lean proofs from several open-source projects formalizing research-level mathematics across multiple domains \citep{Mathlib, HEPLean, PFR, Carleson, ConNF, FLT, Foundation, Seymour}. We hold out all theorems in miniCTX-v2 as the test set. To prevent
data leakage, we additionally exclude from training every theorem
that appears in the same source file as a miniCTX-v2 theorem.
Training and validation sets are drawn from all remaining files
in an $80\%/20\%$ split, although we pre-filter the Mathlib portion of the dataset to a uniformly sampled subset of 37 files, due to its significantly larger scale.

We use miniCTX-v2 as a proxy for deployment on human-written research-level mathematics. Although \ImProver\ could in principle be applied to other domains, such as long machine-generated proofs, these settings fall outside the scope of our main evaluation. As a qualitative test, however, we also evaluate \ImProver\ on machine-generated AlphaProof proofs from the 2024 International Mathematical Olympiad in Appendix~\ref{app:qualitative}.

\paragraph{Evaluation protocol}

All evaluations use Lean v4.17.0. For all main results, we evaluate with best@16 sampling (\ref{sec:generation}) and report the improvement score $\mu(c, x, y)-\mu(c, x, y_0)$ for all problems in the test set and $\mu \in \{\mu_{len},\mu_{dep},\mu_{mod}\}$. With this we compute the mean of the improvement scores, as well as ancillary metrics such as the compilation accuracy $\mathcal{A}(\mathcal{D}_{test}^{(t)})$ and improved accuracy $\mathcal{A}^+_\mu(\mathcal{D}_{test}^{(t)})$, defined as the percentage of compiling theorems that have a strictly positive improvement score. We operate with a base model of $G_0 = $\texttt{DeepSeek-R1-Distill-Qwen-7B} \citep{DeepseekR1}. All models are evaluated, prompted -- and for $G_0$, trained -- with the hyperparameters and configuration described in Appendix~\ref{app:config}.

\paragraph{Systems compared}

Our main evaluation compares \ImProver\ against three classes of baselines on MiniCTX-v2 under best@16 sampling: (i) the \texttt{DeepSeek-R1} family at multiple scales, to study parameter scaling within a fixed model family; (ii) frontier GPT-based and open-weight systems, including \texttt{GPT-5-high} (a full-size high-reasoning variant), \texttt{GPT-5-chat}, \texttt{GPT-5-mini}, \texttt{GPT-5-nano}, and \texttt{GPT-oss-120B}; and (iii) the prior \texttt{ImProver} system and its base model \texttt{GPT-4o}. We also report current API inference costs in Table~\ref{tab:frontier_and_intrafamily}.

We also evaluate all generators with and without the neurosymbolic scaffold $\Psi$ to isolate the effect of augmentation from the effect of training.
For each metric, we train IRPO until validation improvement regresses and report the best checkpoint for that objective.

\vspace{0.5em}

\subsection{Main Results}
\label{sec:main_results}

\begin{table*}
\centering
\small
\begin{minipage}{0.54\textwidth}
\centering
\captionof{table}{\textbf{Frontier \& Intra-family Evaluations.} Comparison with frontier models, intra-family baselines, and prior proof optimization systems. Mean improvement at best@16 on MiniCTX-v2. Input/output cost per 1M tokens also listed for API models.}
\addtolength{\tabcolsep}{-0.2em}
\begin{tabular}{l|llll}
\hline
\textbf{Model}  & \textbf{Length} & \textbf{Mod.} & \textbf{Dep.} & \textbf{Cost} \\
\hline
DS-R1 7B     & 0.118 & 0.003 & 0.050 & Local \\
DS-R1 14B      & 0.140 & 0.037 & 0.093 & Local \\
DS-R1 671B & 0.308 & 0.055 & 0.153 & \$0.70/ \$2.50\\
\hdashline
GPT-4o & 0.336 & 0.034 & 0.050 & \$2.50/\$10 \\
GPT-oss-120B & 0.321 & 0.075 & 0.181 & Local \\
GPT-5-nano  & 0.087 & 0.065 & 0.106 & \$0.05/\$0.40\\
GPT-5-mini   & 0.330 & 0.109 & 0.203 & \$0.25/\$2 \\
GPT-5-chat  & 0.346 & 0.118 & 0.046 & \$1.25/\$10 \\
GPT-5-high  & \textbf{0.660} & 0.120 & \textbf{0.208} & \$1.25/\$10 \\
\hdashline
ImProver & 0.355 & 0.088 & 0.047 & \$2.50/\$10 \\
\ImProver  & 0.330 & \textbf{0.143} & 0.206 & Local  \\
\hline
\end{tabular}
\label{tab:frontier_and_intrafamily}
\end{minipage}\hfill
\begin{minipage}{0.44\textwidth}
\centering
\captionof{table}{\textbf{Per-iteration improvements.} Progression of mean improvement at best@16 on MiniCTX-v2 across IRPO training iterations across all three metrics.}
\begin{tabular}{lccc}
\hline
\textbf{Iteration} & \textbf{Length}  & \textbf{Mod.}  & \textbf{Dep.}  \\
\hline
Base     & 0.118 & 0.003 & 0.050  \\
\textit{+ Scaffold} & 0.236 & 0.007 & 0.056  \\
\hdashline
1   & 0.265 & 0.062 & 0.137  \\
2   & 0.318  & 0.134 & \textbf{0.206}  \\
3   & \textbf{0.330}  & \textbf{0.143} & 0.165  \\
4   & 0.299 & 0.096 & N/A \\
\hline
\end{tabular}
\label{tab:iter}
\end{minipage}
\end{table*}

Table~\ref{tab:frontier_and_intrafamily} shows that \ImProver\ improves the \texttt{DeepSeek-R1 7B} base model on all three objectives, leads all evaluated unscaffolded systems on modularity, and is competitive with frontier models on dependency and length. Namely, after IRPO training, the model improves from $0.118$ to $0.330$ on length, $0.003$ to $0.143$ on modularity, and $0.050$ to $0.206$ on dependency. These gains also exceed the larger \texttt{DeepSeek-R1 14B} and \texttt{DeepSeek-R1 671B} baselines on every metric, suggesting that task-specific training can compensate for substantial generic scale within this model family.

Against frontier and prior systems, \ImProver\ is strongest on the structural metrics in the unscaffolded comparison. It leads all evaluated unscaffolded systems on modularity and is effectively tied with \texttt{GPT-5-high} on dependency ($0.206$ vs.\ $0.208$). On length, it matches \texttt{GPT-5-mini}, exceeds unscaffolded \texttt{GPT-oss-120B}, and trails the high-reasoning \texttt{GPT-5-high} and prior multi-step \texttt{GPT-4o}-based \texttt{ImProver} system. When frontier models receive the same scaffold (Table~\ref{tab:scaffold}), several outperform \ImProver; we therefore interpret these results as evidence for effective specialization rather than dominance over the strongest scaffolded frontier systems.

 We next unpack these aggregate results through three complementary analyses: intra-family parameter scaling, best@$n$ comparisons against frontier and prior systems, and the evolution of performance across IRPO iterations. We then isolate the contribution of the neurosymbolic scaffold across model families.

% \subsubsection{Baseline and Frontier Comparison}

% Against frontier and prior systems, \ImProver\ is strongest on the structural metrics in the unscaffolded comparison and remains competitive with mid-tier frontier models on length optimization despite its relatively small size and cost. Namely, it leads all unscaffolded systems on modularity and is effectively tied with \texttt{GPT-5-high} on dependency ($0.206$ vs.\ $0.208$). On length, it matches \texttt{GPT-5-mini}, exceeds unscaffolded \texttt{GPT-oss-120B}, and trails the high-reasoning \texttt{GPT-5-high} and prior multi-step \texttt{GPT-4o}-based \texttt{ImProver} system. When frontier models receive the same scaffold (Table~\ref{tab:scaffold}), several of them outperform \ImProver; we therefore interpret the SLM result as evidence for effective specialization, not as dominance over the strongest scaffolded frontier systems.

% Indeed, in \ref{sec:scaffold_power} we further analyze the impact of scaffolding across base models. Further best@$n$ and scaling breakdowns appear in Appendix~\ref{app:statistical_analyses}.

\subsubsection{Performance and Parameter Scaling}
\label{sec:perf_param}

\begin{figure*}
\centering
\begin{subfigure}{0.49\linewidth}
    \centering
    \includegraphics[width=1\linewidth]{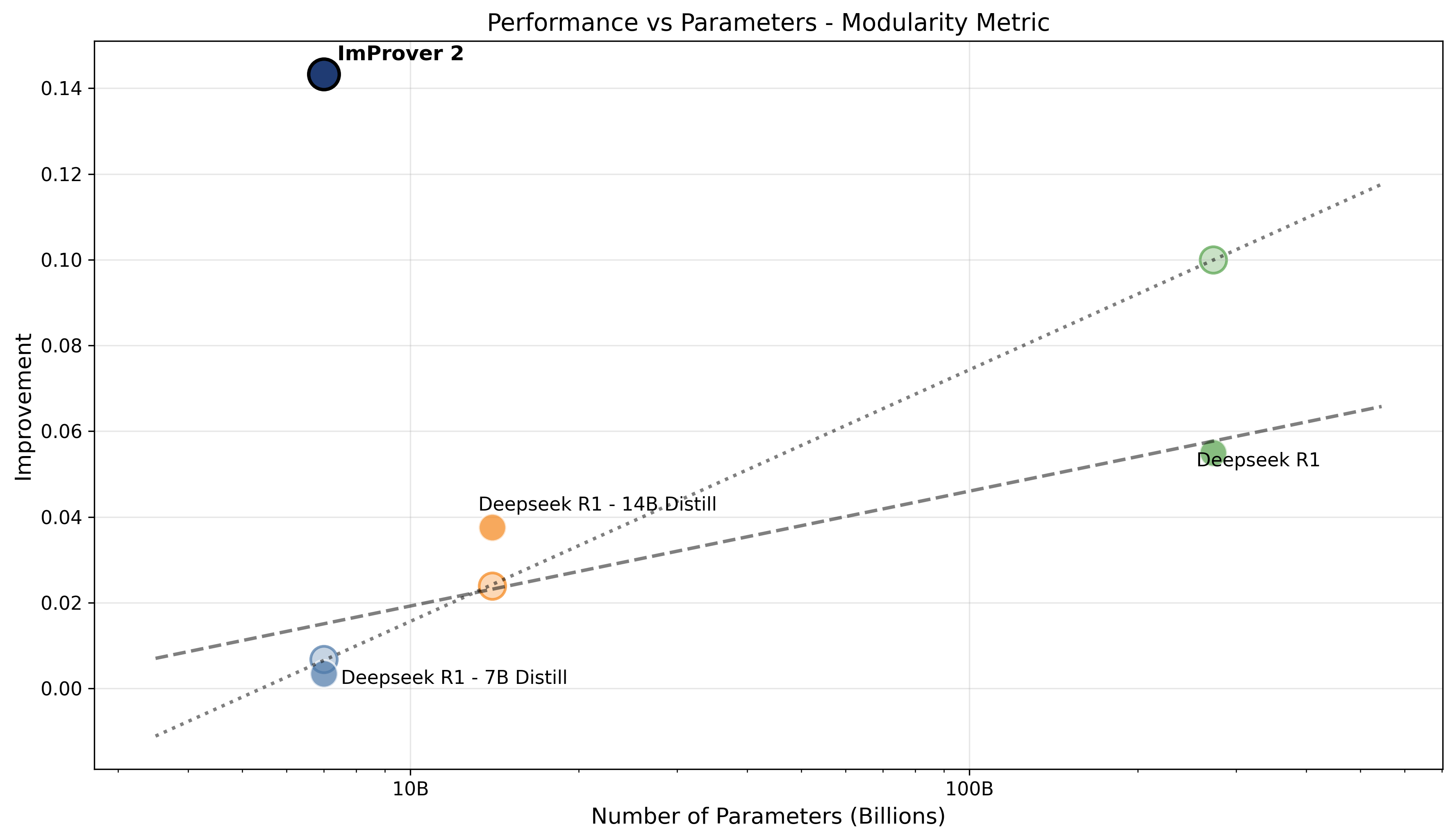}

\end{subfigure}
% \hfill
\begin{subfigure}{0.49\linewidth}
    \centering
    \includegraphics[width=1\linewidth]{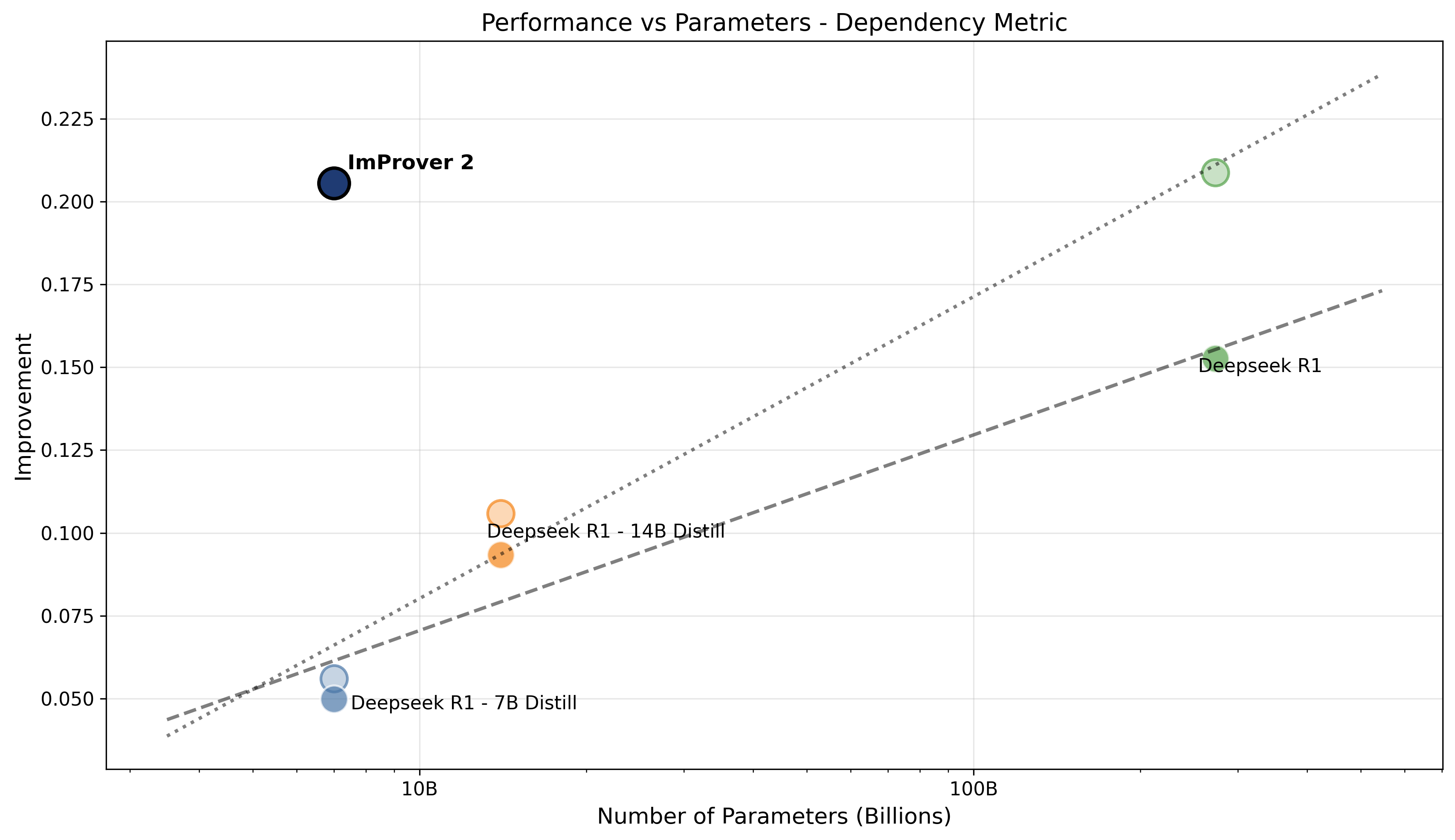}

\end{subfigure}
\hfill

\begin{subfigure}{0.49\linewidth}
    \centering
    \includegraphics[width=1\linewidth]{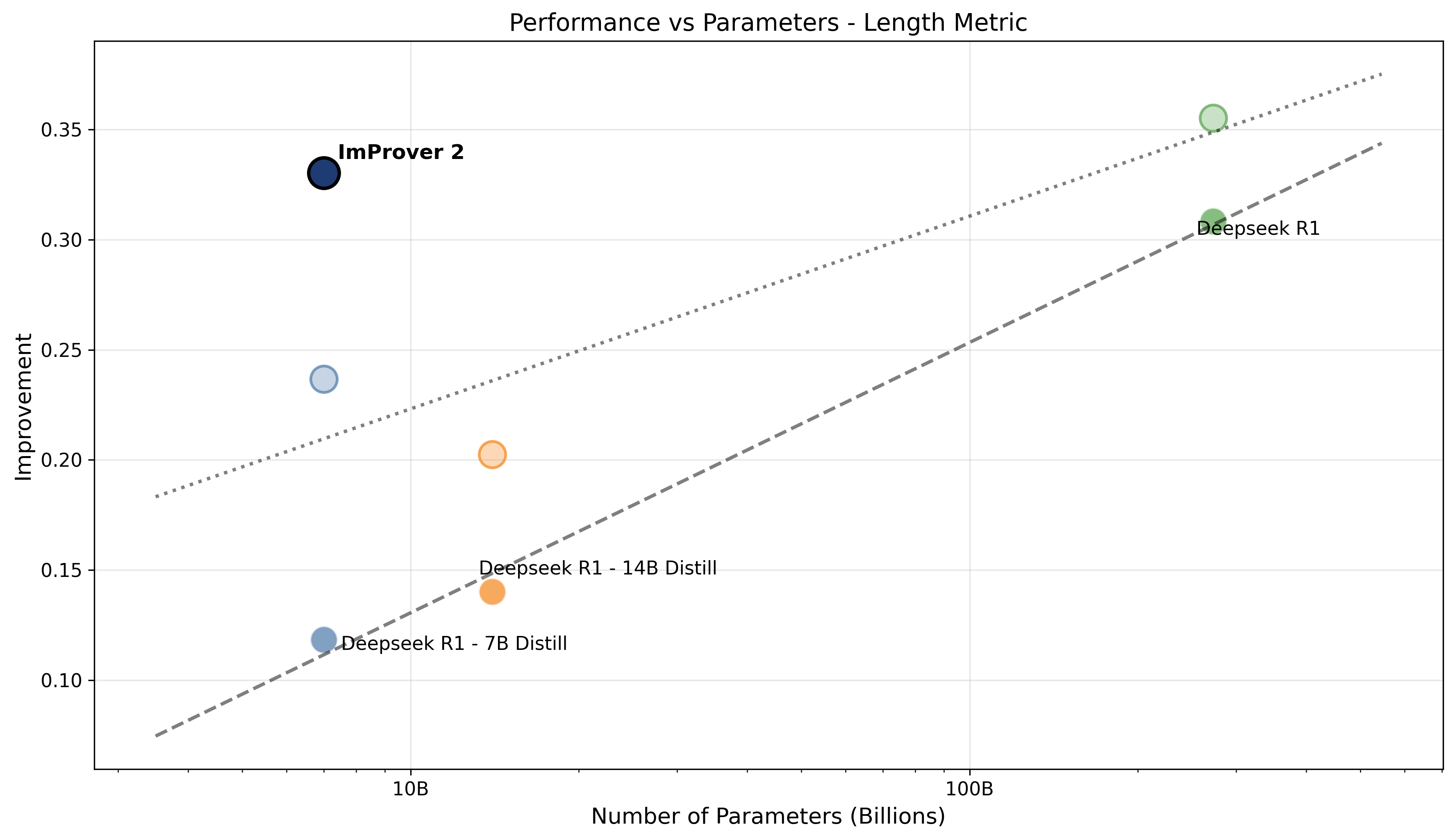}

    \end{subfigure}
        \caption{Effect of parameter count on model performance on mean improvement at best@16 across all three metrics, with \ImProver \ marked.}
        \label{fig:param_perf}
\end{figure*}

A central question is whether proof optimization performance is primarily driven by scale or by task-specific specialization, when all else is equal. Figure~\ref{fig:param_perf} shows that within the \texttt{DeepSeek-R1} model family, larger models generally achieve higher mean improvement under a fixed prompting and sampling protocol. This trend is clearest for length and dependency, indicating that generic reasoning capacity does help with proof refactoring.

At the same time, scale is not the full story. \ImProver, trained from a 7B base model, outperforms the much larger 671B DeepSeek model on all three objectives. The gap is especially pronounced for modularity ($0.143$ vs.\ $0.055$) and dependency ($0.206$ vs.\ $0.153$), suggesting that for structural proof optimization, the relevant bottleneck is not only reasoning capacity but also whether the model has been adapted to the structure of the formal task.

This interpretation is reinforced by the accuracy metrics. Namely, within the same model family, larger general-purpose models often maintain stronger raw compilation rates, but \ImProver\ is more likely to produce compiling proofs that are also improved under the target metric.

We therefore interpret the scaling results as evidence that proof optimization is only partly a scale problem; it is also a specialization problem, and one for which iterative preference optimization is particularly effective.

\subsubsection{Comparison to Frontier and Prior Systems}
\label{sec:frontier_comparison}

\begin{figure*}
    \centering

\begin{subfigure}{0.49\linewidth}
    \centering
    \includegraphics[width=1\linewidth]{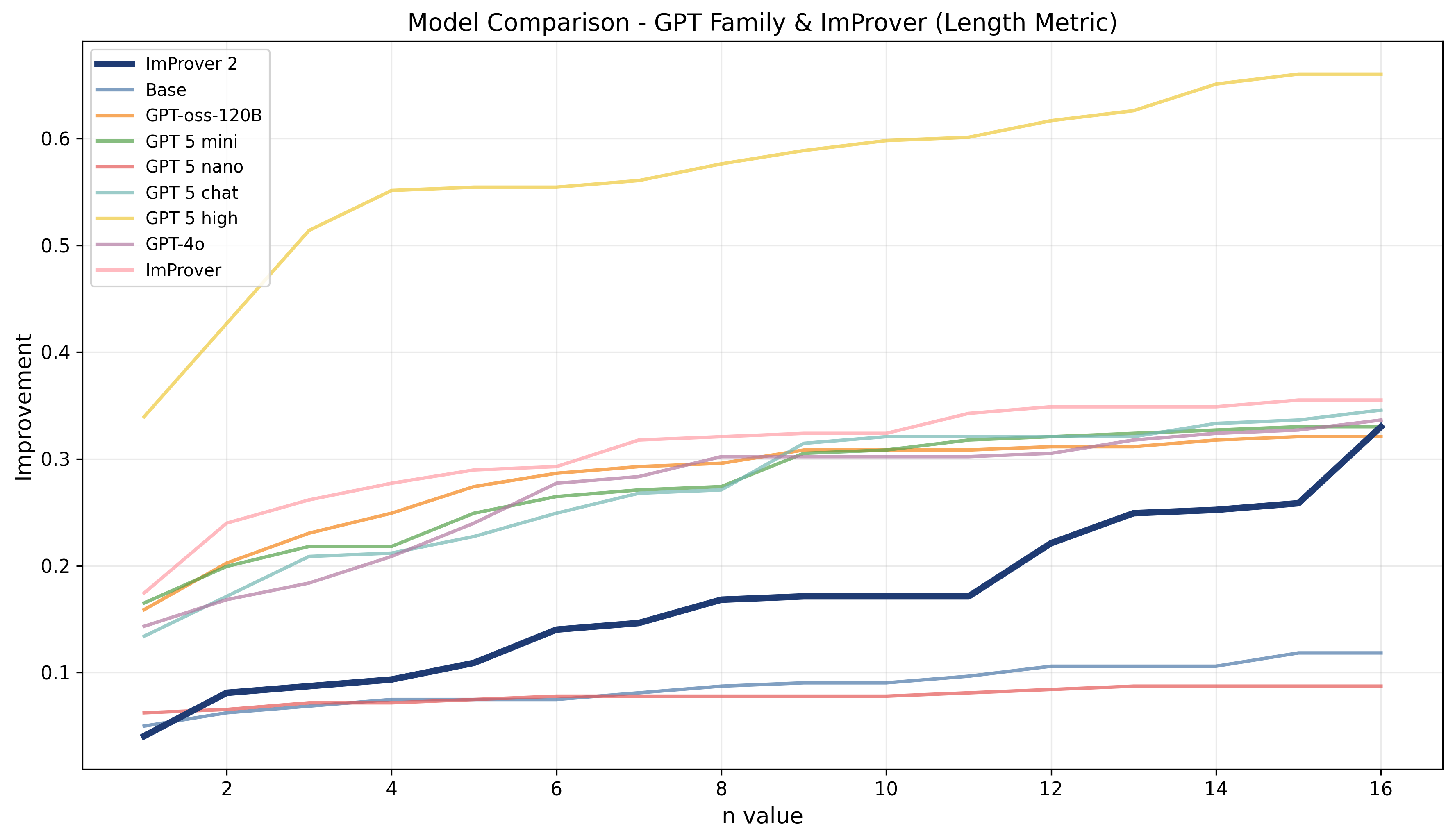}
    \label{fig:length_gpt_improver_comparison}
\end{subfigure}
\hfill
\begin{subfigure}{0.49\linewidth}
    \centering
    \includegraphics[width=1\linewidth]{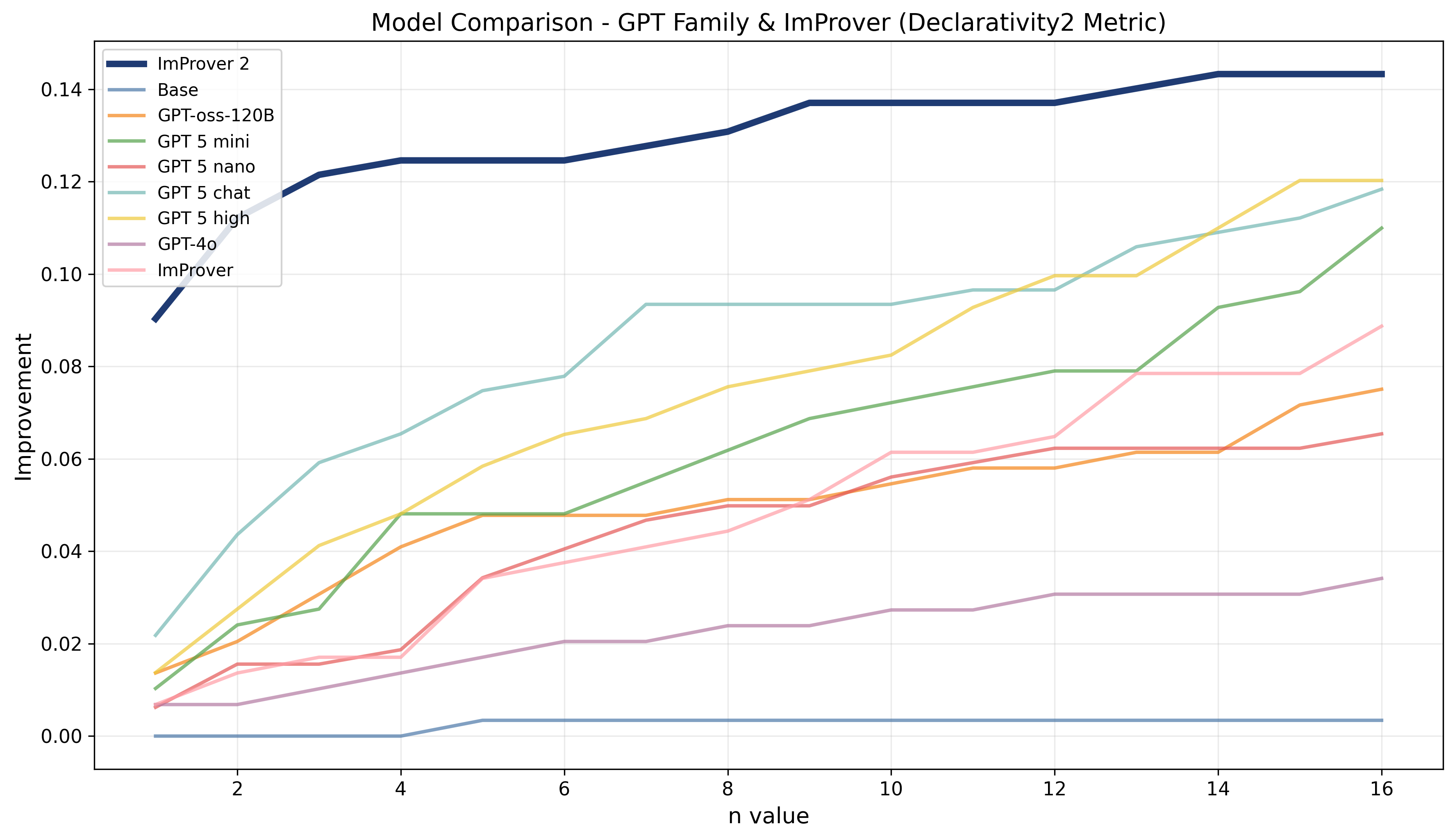}
    \label{fig:declarativity2_gpt_improver_comparison}
\end{subfigure}
\hfill
\begin{subfigure}{0.49\linewidth}
    \centering
    \includegraphics[width=1\linewidth]{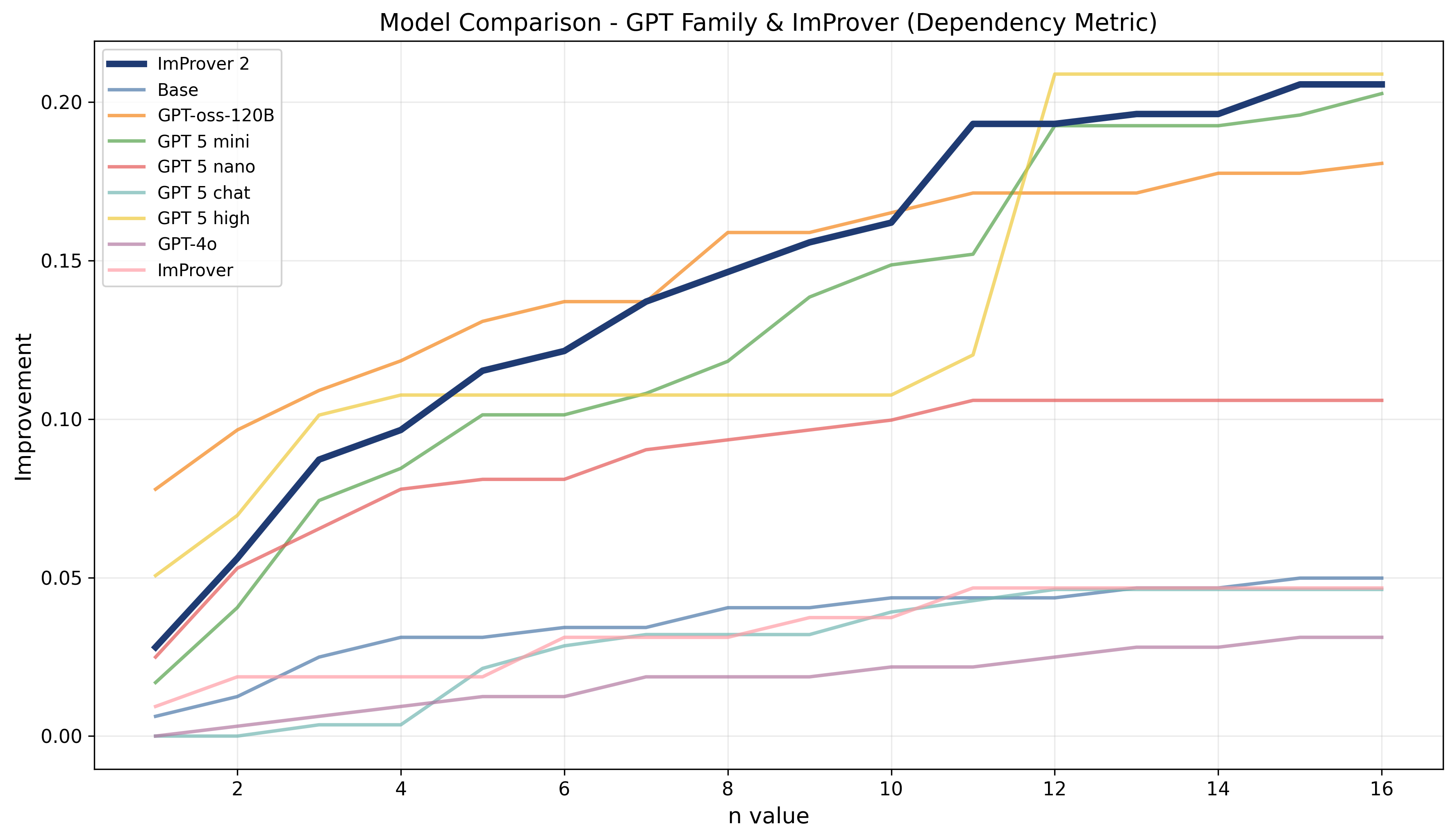}
    \label{fig:dependency_gpt_improver_comparison}
    \end{subfigure}
    \caption{Comparison of \ImProver \ against frontier GPT-based models and ImProver, evaluated on mean improvement at best@$n$ on all metrics, from $n=1$ to $n=16$.}
        \label{fig:gpt_improver_comparison}
\end{figure*}

We compare \ImProver\ against frontier closed-source models, a large open-weight baseline, and the prior \texttt{ImProver} system.
The resulting picture is mixed but informative: frontier systems are already strong proof rewriters, yet their strengths differ by objective, while \ImProver\ is most competitive on the structural metrics.

Among the frontier baselines evaluated without scaffolding, \ImProver\ achieves the strongest modularity score ($0.143$), showing that iterative task-specific training can produce structural rewrites that generalist frontier models do not always find under the same single-shot, unscaffolded protocol. On dependency, \ImProver\ ($0.206$) is competitive with the strongest frontier model evaluated, \texttt{GPT-5-high} ($0.208$), and leads \texttt{GPT-5-mini} ($0.203$); we treat all three as effectively tied at the level of mean performance. On length, \ImProver\ ($0.330$) matches \texttt{GPT-5-mini} and exceeds unscaffolded \texttt{GPT-oss-120B} ($0.321$), trailing only \texttt{GPT-5-high} ($0.660$), which represents a qualitatively different operating point in terms of inference cost.

Relative to large open-weight baselines, \ImProver\ is also strong in the unscaffolded comparison.
It matches or exceeds unscaffolded \texttt{GPT-oss-120B} on all three metrics despite starting
from a smaller base model. We note that \texttt{GPT-oss-120B} is a sparse
mixture-of-experts model activating approximately 5B parameters per token, so the
raw parameter count comparison overstates the computational gap. Moreover, when \texttt{GPT-oss-120B} receives the same scaffold, it substantially outperforms \ImProver\ on length and dependency (Table~\ref{tab:scaffold}). We therefore interpret the result as evidence that metric-specific iterative self-improvement can make a small dense model useful and competitive in some settings, not as evidence that it dominates the strongest scaffolded open-weight baselines.

The comparison with the prior \texttt{ImProver} system is also revealing. The original ImProver system leads \ImProver\ on length ($0.355$ vs.\ $0.330$), consistent
with its use of a multi-step prompting strategy with a strong proprietary base model. By contrast, \ImProver\ leads on modularity ($0.143$ vs.\ $0.088$) and outperforms ImProver substantially on dependency ($0.206$ vs.\ $0.047$), suggesting that iterative structural supervision can learn refactoring behaviors that are not reliably induced by prompting alone.

\subsubsection{Per-Iteration Improvement}

\begin{figure*}
    \centering

\begin{subfigure}{0.49\linewidth}
    \centering
    \includegraphics[width=1\linewidth]{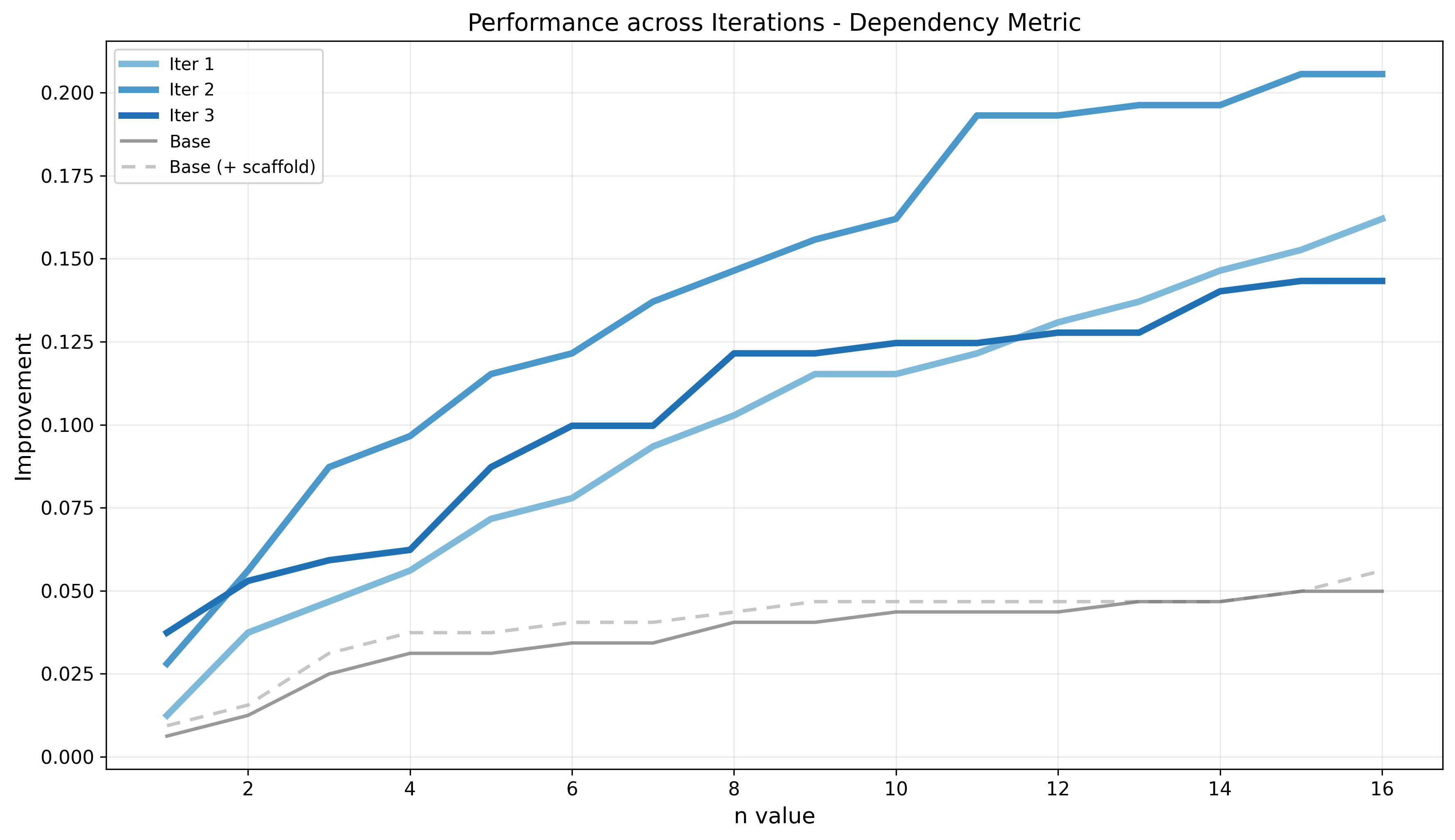}
\end{subfigure}
\hfill

\begin{subfigure}{0.49\linewidth}
    \centering
    \includegraphics[width=1\linewidth]{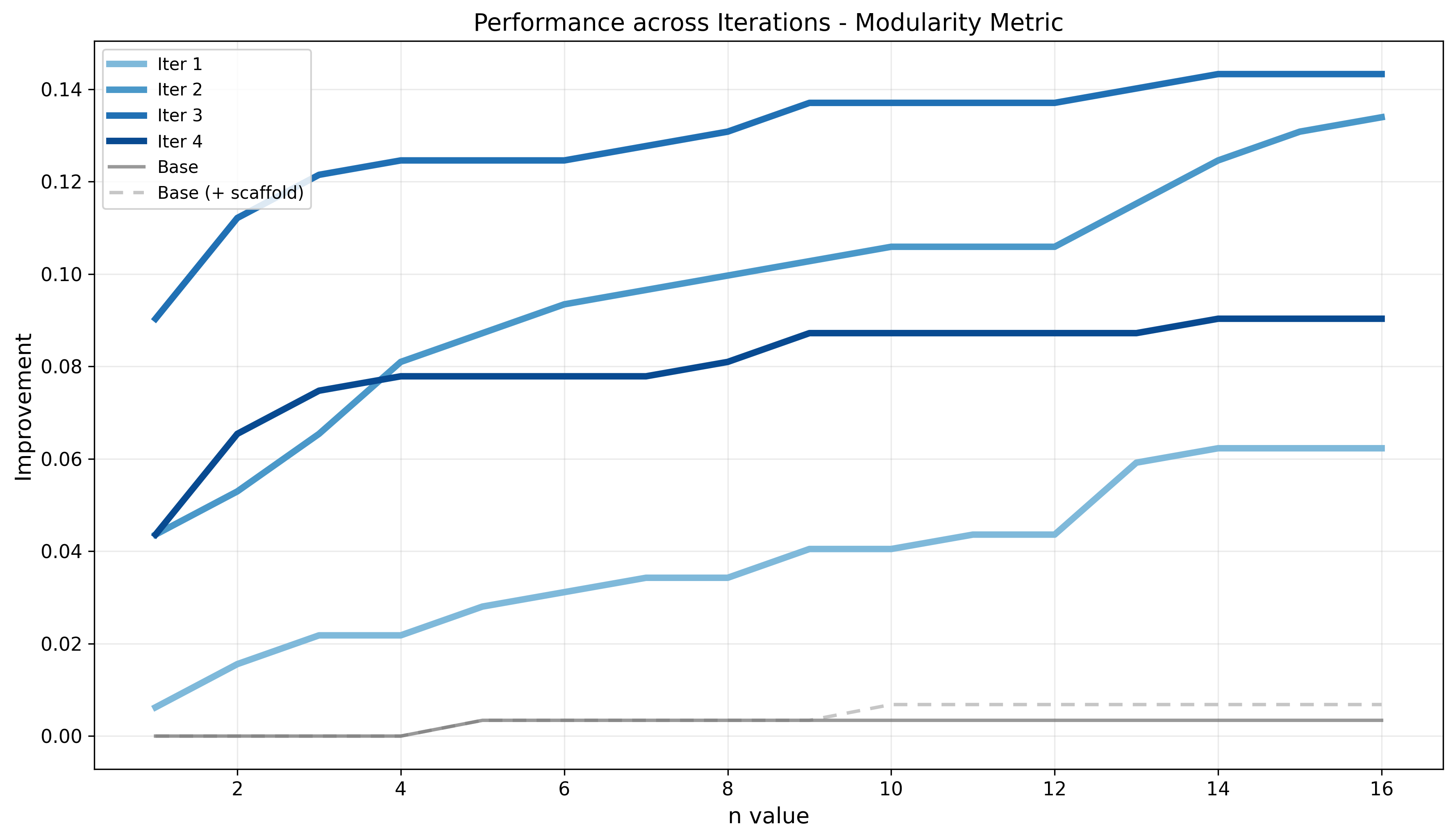}
\end{subfigure}
\hfill
\begin{subfigure}{0.49\linewidth}
    \centering
    \includegraphics[width=1\linewidth]{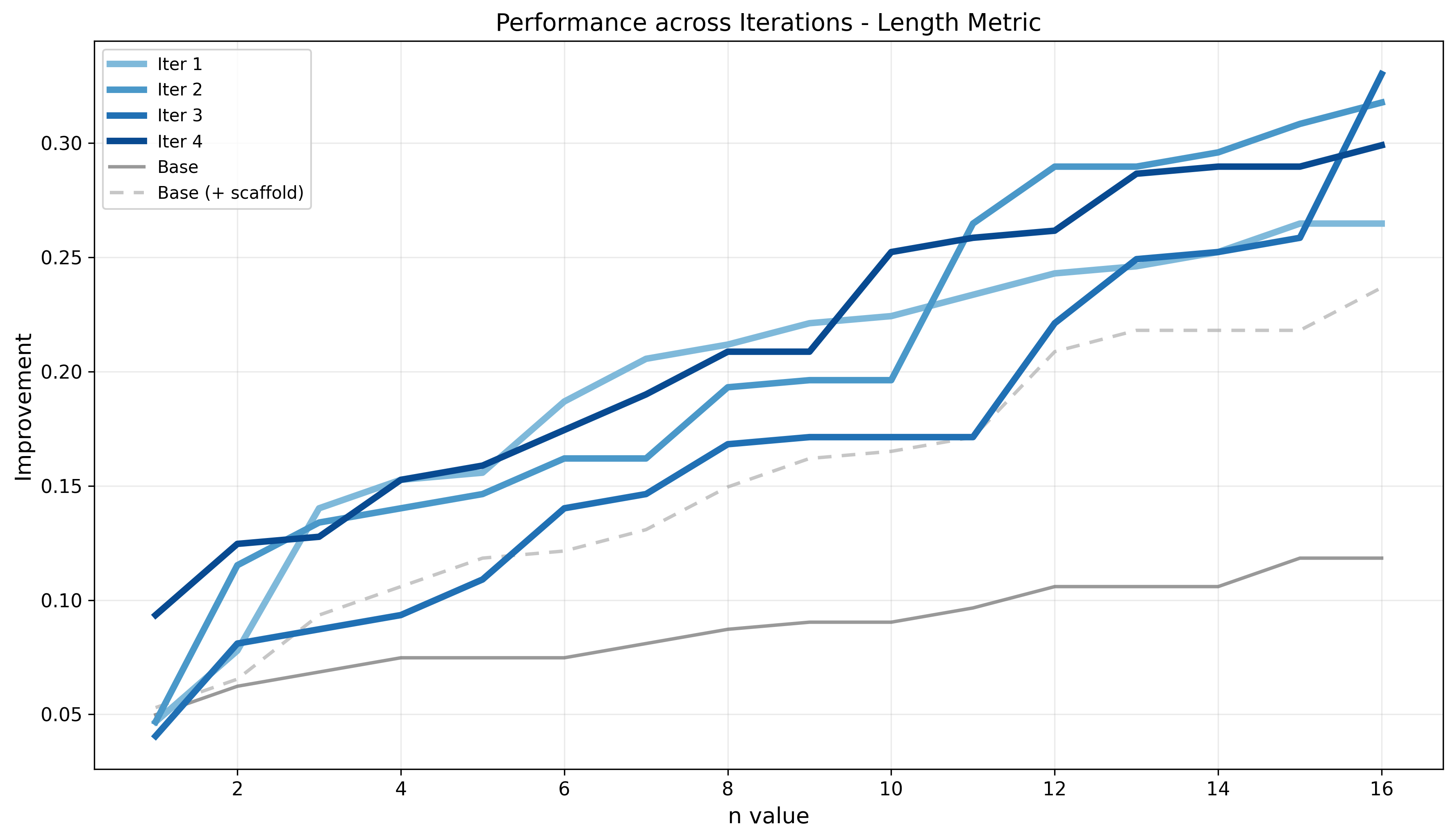}
\end{subfigure}
 \caption{Performance of \ImProver\ as a function of sample budget across training iterations. Gains are largest in early iterations, with saturation by iterations 2–3 and mild regression thereafter.}
 \label{fig:all_iterations}
\end{figure*}

Table~\ref{tab:iter} shows that most improvement occurs within the first two or three IRPO rounds. Dependency peaks at iteration 2, while length and modularity peak at iteration 3, so we select checkpoints separately by metric. These gains suggest that the model is learning useful refactoring behavior from its own filtered generations, and that the replay-buffered preference data remains informative for multiple rounds. Indeed, as shown by the full best@$n$ curves (Figure~\ref{fig:all_iterations}), the trained checkpoints dominate the untrained and scaffold-only baselines across nearly all sample budgets, indicating that the gains are not caused by a single high-sample outlier.

Improvement is not indefinite, however, as later rounds plateau or regress. We interpret this as indicating saturation rather than instability, as once the most common and highest-yield refactoring patterns have been absorbed, later rounds appear to produce fewer genuinely novel improvements and increasingly focus on narrower or noisier examples.

\subsubsection{Effect of Neurosymbolic Scaffolding}
\label{sec:scaffold_power}

Table~\ref{tab:scaffold} isolates the effect of the scaffold $\Psi$ from training. Namely, across nearly all evaluated generators and metrics, adding the scaffold $\Psi$ improves mean best@16 performance, often by a large margin. For example, on length, we observe \texttt{DeepSeek-R1 7B} improves from $0.118$ to $0.236$, \texttt{GPT-5-mini} from $0.330$ to $0.632$, and \texttt{GPT-5-high} from $0.660$ to $0.875$. Dependency also improves for most systems, while modularity gains are smaller but generally positive. This suggests that the scaffold is not merely compensating for weak base models; rather, it changes the representation of the task in a way that makes better rewrites easier to discover under a fixed sampling budget. Overall, the scaffold appears to expand the set of useful rewrites the model can reliably attempt.
As such, by exposing goal-state information, relevant context, and informal abstractions, it helps both small and large models move beyond surface-level editing toward more substantive proof refactoring by improving models' abilities to search over valid higher-value rewrites under a fixed sampling budget.

\begin{table*}[t]
\centering
\caption{\textbf{Scaffold Evaluations.} Effect of neurosymbolic scaffolding. Mean improvement at best@16 with and without the scaffold $\Psi$, across model families and metrics.}
\small
\setlength{\tabcolsep}{5pt}
\begin{tabular}{l|cc|cc|cc}
\toprule
\multirow{2}{*}{\textbf{Model}} & \multicolumn{2}{c|}{\textbf{Length}} & \multicolumn{2}{c|}{\textbf{Mod.}} & \multicolumn{2}{c}{\textbf{Dep.}} \\
& \textbf{Base} & \textbf{Scaffold} & \textbf{Base} & \textbf{Scaffold} & \textbf{Base} & \textbf{Scaffold} \\
\midrule
DeepSeek-R1 7B & 0.118 & \textbf{0.236} & 0.003 & \textbf{0.007} & 0.050 & \textbf{0.056} \\
DeepSeek-R1 14B & 0.140 & \textbf{0.202} & \textbf{0.037} & 0.024 & 0.093 & \textbf{0.106} \\
DeepSeek-R1 (671B) & 0.308 & \textbf{0.355} & 0.055 & \textbf{0.100} & 0.153 & \textbf{0.209} \\
GPT-4o & 0.336 & \textbf{0.396} & 0.034 & \textbf{0.092} & 0.050 & \textbf{0.075} \\
GPT-oss-120B & 0.321 & \textbf{0.508} & 0.075 & \textbf{0.092} & 0.181 & \textbf{0.406} \\
GPT-5-nano & 0.087 & \textbf{0.296} & 0.065 & \textbf{0.069} & 0.106 & \textbf{0.108} \\
GPT-5-mini & 0.330 & \textbf{0.632} & 0.109 & \textbf{0.123} & 0.203 & \textbf{0.267} \\
GPT-5-chat & 0.346 & \textbf{0.576} & 0.118 & \textbf{0.153} & 0.046 & \textbf{0.087} \\
GPT-5-high & 0.660 & \textbf{0.875} & 0.120 & \textbf{0.183} & 0.208 & \textbf{0.315} \\
\bottomrule
\end{tabular}
\label{tab:scaffold}
\end{table*}

\subsubsection{Additional Analyses}
\label{app:statistical_analyses}

\paragraph{Improvement vs Accuracy}
\label{sec:acc}

Mean improvement alone does not distinguish between broad, reliable gains and a smaller set of high-reward successes.
We therefore also report compilation accuracy $\mathcal{A}$ and improved accuracy $\mathcal{A}_\mu^+$, where $\mathcal{A}_\mu^+$ measures the fraction of test problems for which the model produces a compiling proof that has a strictly positive metric improvement score.

Across both models and training iterations, optimization tends to raise $\mathcal{A}_\mu^+$ faster than $\mathcal{A}$.
This is especially visible in the iteration study (Table~\ref{tab:iter_acc}).
For dependency minimization, the base model begins with high compilation accuracy but low improved accuracy ($0.754$ and $0.037$ respectively), indicating that it often produces valid rewrites without meaningfully reducing dependency footprint.
After training, improved accuracy rises sharply, peaking at $0.106$ in iteration 2, while compilation accuracy drops to $0.464$.
This reflects a central tradeoff of proof refactoring: larger structural edits are more likely to yield real gains when they succeed, but they also create more opportunities for compilation failure.

The same pattern appears in cross-model comparisons.
For example, \texttt{GPT-5-nano} attains very high compilation accuracy on dependency ($0.894$) but only moderate improved accuracy ($0.065$), whereas \ImProver\ attains lower compilation accuracy ($0.368$) but slightly higher improved accuracy ($0.069$).
This suggests that conservative models often preserve correctness by making safer edits, while specialized optimizers are more willing to attempt riskier transformations that improve the target metric when successful.
Accordingly, we view $\mathcal{A}$ and $\mathcal{A}_\mu^+$ as complementary: $\mathcal{A}$ measures stability, while $\mathcal{A}_\mu^+$ more directly captures whether the system is solving the optimization problem rather than merely preserving compilability.

\begin{table}
\centering
\caption{\textbf{Accuracy Evaluations.} Compilation accuracy and improved accuracy comparison amongst the best@16 mean improvement samples across all three metrics. Each entry is reported as $\mathcal{A}_\mu^+ / \mathcal{A}$, where $\mathcal{A}_\mu^+$ is improved accuracy and $\mathcal{A}$ is compilation accuracy.}
\small
\setlength{\tabcolsep}{5pt}
\begin{tabular}{l|ccc}
\hline
\textbf{Model} & \textbf{Length} & \textbf{Mod.} & \textbf{Dep.} \\
\hline
DS-R1 7B      & 0.062 / 0.617 & 0.003 / 0.570 & 0.037 / 0.754 \\
DS-R1 14B     & 0.075 / 0.660 & 0.034 / \textbf{0.775} & 0.065 / 0.567 \\
DS-R1 671B  & 0.162 / 0.536 & 0.048 / 0.536 & 0.109 / 0.480 \\
\hline
ImProver 2 & 0.131 / 0.657 & 0.065 / 0.579 & 0.069 / 0.368 \\
\hline
ImProver            & 0.171 / 0.560 & 0.068 / 0.597 & 0.031 / 0.249 \\
GPT-4o              & 0.158 / 0.595 & 0.031 / 0.464 & 0.028 / 0.227 \\
GPT-oss        & 0.171 / 0.692 & 0.068 / 0.477 & 0.097 / 0.676 \\
GPT-5-nano          & 0.044 / 0.461 & 0.053 / 0.757 & 0.065 / \textbf{0.894} \\
GPT-5-mini          & 0.159 / 0.623 & 0.085 / 0.525 & \textbf{0.111} / 0.834 \\
GPT-5-chat          & 0.165 / 0.586 & 0.084 / 0.455 & 0.025 / 0.185 \\
GPT-5-high          & \textbf{0.264} / \textbf{0.757} & \textbf{0.092} / 0.656 & 0.094 / 0.753 \\
\hline
\end{tabular}
\label{tab:acc}
\end{table}

\begin{table}
\centering
\caption{\textbf{Accuracy Progression.} Compilation accuracy and improved accuracy progression amongst the best@16 mean improvement samples across IRPO training iterations. Each entry is reported as $\mathcal{A}_\mu^+ / \mathcal{A}$, where $\mathcal{A}_\mu^+$ is improved accuracy and $\mathcal{A}$ is compilation accuracy.}
\small
\setlength{\tabcolsep}{5pt}
\begin{tabular}{l|ccc}
\hline
\textbf{Model} & \textbf{Length} & \textbf{Mod.} & \textbf{Dep.} \\
\hline
Base      & 0.062 / 0.617 & 0.003 / 0.570 & 0.037 / 0.754 \\
\hline
Iter. 1 & 0.125 / 0.720 & 0.044 / \textbf{0.679} & 0.093 / 0.417 \\
Iter. 2 & \textbf{0.131} / 0.679 &\textbf{ 0.121} / 0.614 & \textbf{0.106} / \textbf{0.464} \\
Iter. 3 & 0.121 / \textbf{0.682} & 0.118 / 0.579 & 0.069 / 0.368 \\
Iter. 4 & \textbf{0.131} / 0.657 & 0.065 / 0.579 & N/A \\

\hline
\end{tabular}
\label{tab:iter_acc}
\end{table}

\paragraph{Per-repository heterogeneity}
\label{sec:repo_hetero}

\begin{table}
\centering
\caption{\textbf{Average improvement by project and metric.}}
\label{tab:project_improvement}
\small
\begin{tabular}{l|ccc}
\hline
\textbf{Project} & \textbf{Length} & \textbf{Dependency} & \textbf{Modularity} \\
\hline
HepLean    & \textbf{1.283} & 0.379 & 0.030 \\
ConNF      & 0.420 & 0.278 & 0.048 \\
Seymour    & 0.199 & 0.100 & \textbf{0.359} \\
FLT        & 0.131 & 0.133 & 0.066 \\
Foundation & 0.082 & 0.015 & 0.219 \\
Carleson    & 0.048 & 0.188 & 0.095 \\
Mathlib    & 0.016 & \textbf{0.306} & 0.163 \\
\hline
\end{tabular}
\end{table}

Performance varies substantially across repositories (Table~\ref{tab:project_improvement}), suggesting that optimization opportunity may be mediated by project-specific proof style, theorem difficulty distributions, and domain.
For example, Mathlib exhibits very small length gains but relatively strong dependency and modularity improvements, consistent with a library whose proofs are often already concise but still admit structural refactoring.
By contrast, HepLean and ConNF show much larger length and dependency gains, suggesting that these corpora contain more opportunities for proof compression and simplification.
We treat these repository-level results as descriptive rather than definitive, since they likely reflect both stylistic differences and variation in theorem composition across projects.

\subsection{Overview of Additional Experiments}
\label{sec:additional_exp}

\paragraph{Ablation Studies}
Appendix~\ref{app:ablations} separates the scaffold channels and reports the hyperparameter searches used during training. The ablation shows that goal-state traces, informalizations, and retrieved context each help (albiet with a majority of improvement stemming from the chain-of-states annotations), while the grid searches motivate per-iteration and per-metric hyperparameter selection and evolution.

% \paragraph{Additional Analyses}
% Appendix~\ref{app:statistical_analyses} extends the analysis from Section~\ref{sec:main_results}, reporting full best@$n$ curves, compilation and improved accuracy trade-offs, intra-family parameter-scaling results, frontier best@$n$ comparisons, and repository-level heterogeneity of performance.

\paragraph{Qualitative Examples}
Appendix~\ref{app:qualitative} gives representative optimized proofs from the MiniCTX-v2 dataset for all three objectives, including provenance and old/new metric scores. Moreover, we additionally include a case study of evaluation on AI-generated proofs (Alphaproof on IMO 2024 \citep{AlphaProof}) with a higher compute budget.

\section{Limitations and Future Work}

In this work, we posit and study a set of particular structural metrics, which may not necessarily align with corpus maintainers' subjective preferences of proof quality. Namely, the dependency and modularity metrics measure the number of explicitly named dependencies and effective spawned goals respectively, which are structural formal objects and therefore may carry discrepancies between raw dependency scores and maintainer preferences. We do not include a maintainer preference study, and future work may wish to address this gap by engineering maintainer preferences directly (perhaps through informal, LLM-based metrics). Additionally, future work may also wish to study the effects of training dataset optimization on downstream prover performance.

Additionally, we study single-step rewriting rather than full agentic systems. The scaffold can be exposed as a tool, but we do not evaluate Codex- or Claude-Code-style agents in this particular work as this work primarily focuses on the development and evaluation of our training pipeline and scaffold, and as such, we prioritize the robust evaluation of single-step rewriting. However, we do find that stronger optimization often lowers compilation accuracy \ref{sec:acc}, suggesting that such agents and agentic iterative repair loops are potential methods to balance this trade-off.

\section{Conclusion}

We have introduced \ImProver, a pipeline for boosting the formal proof-optimization ability of small language models. We have demonstrated the utility of our neurosymbolic augmentations to many varieties of models, showcased iterative self-improvement using our replay buffer architecture, and introduced two novel metrics (as well as revisiting an old one) for practical proof improvement. We have found that our system leads all evaluated unscaffolded systems on proof modularity, is competitive with the strongest frontier models on dependency, and matches mid-tier frontier models on length --- all from a 7B base model.

\section*{Acknowledgements}
% [Redacted during anonymous review]
This research is partially supported by the DARPA expMath program through the DARPA CMO contract number HR0011262E028 and NSF Grant DMS-2434614. 
We would like to thank Dr. Patrick Shafto, expMath Program Manager, for useful technical discussions.
% Work partially supported by NSF Grant DMS-2434614 and a gift from Convergent Research.

% \section*{Impact Statement}
% This paper presents work whose goal is to advance the field of Machine
% Learning. There are many potential societal consequences of our work, none
% which we feel must be specifically highlighted here.

\newpage

\bibliography{neurips_2026}
\bibliographystyle{plainnat}

\newpage
\appendix

\section{Formal Definition of the Modularity Metric}
\label{app:modularity_metric}

This appendix presents a precise, proof-theoretic and type-theoretic definition of the modularity metric $\mu_{\mathrm{mod}}$ used throughout the paper, matching its concrete implementation in Lean~4. We proceed from first principles, beginning with Lean's elaboration semantics, and culminate in the exact algorithm used to count effective spawned goals.

\subsection{Lean Elaboration Semantics}

Lean elaborates tactic proofs by incrementally constructing and solving metavariables. At any point in elaboration, the system maintains a metavariable context $M = (\Delta, \sigma)$ where $\Delta$ is a finite set of metavariable declarations of the form $?m : (\Gamma_{?m} \vdash A_{?m})$, with local context $\Gamma_{?m}$ and target type $A_{?m}$.
Similarly, $\sigma$ is a partial assignment mapping metavariables to terms.

A goal is an unassigned metavariable $?m \in \Delta \setminus \mathrm{dom}(\sigma)$. At elaboration step $i$, Lean maintains a list of \emph{active goals} $\mathcal{A}_i$.

Each tactic step $\tau_i$:
\begin{enumerate}
\item Focuses a goal $?m_i \in \mathcal{A}_i$,
\item Produces an assignment $\sigma_{i+1}(?m_i) = t_i$,
\item Potentially introduces new metavariables corresponding to subgoals.
\end{enumerate}

The set of newly created metavariables is exactly the set of free metavariables occurring in $t_i$ after assignment.

\subsection{Direct Children vs.\ Spawned Goals}

Let $\mathrm{Children}(\tau_i)$ denote the set of metavariables that occur free in $t_i$ and therefore represent \emph{direct subgoals} of the focused goal $?m_i$.

However, Lean tactics may surface additional obligations that are \emph{not} direct children of $?m_i$. These arise from nested tactic blocks (e.g.\ \texttt{have}, \texttt{calc}, \texttt{by}), automation introducing auxiliary lemmas, goal defocus/refocus patterns, tactics that internally elaborate subproofs, etc.

To capture this distinction, define:
\[
\mathrm{Current}_i = \{?m_i\} \cup \mathrm{Children}(\tau_i),
\]
and let $S_i$ be the set of all metavariables that have previously appeared as children:
$S_{i+1} = S_i \cup \mathrm{Children}(\tau_i)$.

\begin{definition}[Spawned Goals]
The \emph{spawned goals} of step $\tau_i$ are defined as:
\[
\mathrm{Spawned}(\tau_i)
= \bigl(\mathrm{Current}_i \setminus \mathrm{Children}(\tau_i)\bigr) \setminus S_i.
\]
\end{definition}

Intuitively, these are goals that first appear at step $i$, are not logical subgoals of the focused goal, correspond to independent subproof obligations.

\subsection{Proof Graph over Steps}

We represent a proof as a directed graph over steps rather than metavariables.

Let steps be indexed by $i=1,\dots,T$. We define:
\begin{itemize}
\item A \emph{normal edge} $i \to j$ if the goal solved at step $j$ is a direct child of step $i$.
\item A \emph{spawned edge} $i \rightsquigarrow j$ if the goal solved at step $j$ was spawned at step $i$.
\end{itemize}

This yields a labeled directed forest
\[
(V, E_{\mathrm{normal}}, E_{\mathrm{spawned}}),
\]
with $E_{\mathrm{spawned}} \subseteq  E_{\mathrm{normal}}$. We refer to this as the proof tree.

\subsection{Canonical Goal Representation}

To prevent adversarial or spurious inflation of modularity, goals are compared modulo definitional equality, $\alpha$-equivalence, and universal abstraction.

Each goal is assigned a canonical representation consisting of:
\begin{itemize}
\item A base target hash,
\item A sorted list of hypothesis type hashes (proof-relevant hypotheses only),
\item A set of sequent variant hashes, obtained by progressively discharging $\forall$-binders and implications,
\item A set of target-only variant hashes, used for wrapper detection.
\end{itemize}

All hashes are computed after $\beta\delta\iota$-normalization, metadata erasure, binder name scrubbing (for $\alpha$-invariance), and replacement of free variables by deterministic canonical constants.

This representation ensures that goals differing only by irrelevant syntactic structure or parameter order are identified.

\subsection{Duplicate and Wrapper Detection}

A spawned goal is considered duplicate and discarded if either:
\begin{enumerate}
\item Its sequent variant hash matches any previously seen goal (global duplicate), or
\item Its target is a definitional wrapper of its parent goal, i.e. $\text{target}(g_{\text{child}}) \in \text{targetVariants}(g_{\text{parent}})$ or vice versa.

\end{enumerate}

This eliminates the potential for many types of reward hacking cases, such as trivial $\forall$-introductions, restatements of the parent goal, and redundant lemma spawning.

\subsection{Nontriviality Filter}

Let $T_g$ be the subtree of steps rooted at a spawned goal $g$.

We require $|T_g| > 2$, ensuring that the goal is not solved immediately, and namely, the goal is not discharged by a single automation step.

This empirically excludes trivial goals solvable by tactics such as
\texttt{simp}, \texttt{tauto}, \texttt{linarith}, \texttt{ring}, \texttt{aesop}, or \texttt{grind}.

\subsection{Effectiveness via Fixed-Point Semantics}

Let each spawned goal $g$ introduce a set of proof variables $\mathsf{Intro}(g)$ corresponding to hypotheses unavailable in the parent context.

We define a spawned subtree rooted at $g$ to be effective if its introduced hypotheses are used in the main (non-spawned) proof, or in another spawned subtree that is itself effective.

Formally, define a monotone operator $\Phi$ on sets of spawned roots:
% \[
% \Phi(S) =
% \{ g \mid
% \mathsf{Intro}(g) \text{ is used outside all spawned subtrees}
% \newline \vee\
% \exists g' \in S,\, g' \neq g,\ \mathsf{Intro}(g) \text{ used in subtree } g'
% \}.
% \]
\begin{align*}
\Phi(S) =
\{ g \mid
\mathsf{Intro}(g) \text{ is used outside all spawned subtrees}
\newline \\
\vee\
\exists g' \in S,\, g' \neq g,\ \mathsf{Intro}(g) \text{ used in subtree } g'
\}.
\end{align*}

\begin{definition}[Effective Spawned Goals]
The set of effective spawned goals is the least fixed point of $\Phi$.
\end{definition}

This is computed by standard fixed-point iteration, guaranteed to terminate since the set of spawned roots is finite.

\subsection{Modularity Metric}

\begin{definition}[Modularity Metric]
Given a verified proof $y$ of $(c,x)$, let $\mathcal{E}(y)$ be the set of effective spawned goals. The modularity metric is defined as:
\[
\mu_{\mathrm{mod}}(c,x,y) = |\mathcal{E}(y)|.
\]
\end{definition}

\section{Context Extraction}
\label{app:context_extraction}

% JA: What are "spans"? Do you mean start and end locations in the source file? It would help to explain.
% (Tate) I clarified this a bit

To facilitate extraction of relevant context from the proof environment, we consider the Lean 4 concrete syntax tree (CST) and abstract syntax tree (AST). The CST preserves surface tokens and references locations in source code; the AST resolves names, binds variables, and canonicalizes declarations (e.g., definitions/theorems/structure/etc. nodes). Context extraction uses both:
\begin{itemize}
\item CST view (textual reachability): harvest all surface identifiers and their source spans that occur in $x$ and in the proof text of $y_0$.
\item AST view (semantic reachability): resolve those identifiers to fully qualified symbols under the language's environment (imports, namespaces, instances, modules); record declaration categories (e.g., definition, theorem) and provenance (module/package).
\end{itemize}

% JA: This next paragraph, especially the definition of $S(c,x,y_0)$, is tough going. What the heck is "Reach"? It's o.k. to keep all the technical stuff, but it would help a lot to say something that conveys the key ideas and intuitions.
% (Tate): I moved this down here from the main body because of how much technical spam was in it. I've added a nice intuitive explanation where it was before.

\paragraph{Graphs and the slice.}
From the AST we build two standard graphs:
(i) an import DAG over files/modules, and
(ii) an entity graph whose vertices are declarations (constants, lemmas, definitions) with edges for semantic references (uses in types/bodies).
Let $\mathsf{touch}(x,y_0)$ be the set of AST nodes directly referenced by the CST identifiers collected from $x$ and $y_0$ (optionally augmented by a dynamic trace of proof states, if available). The context slice is the subgraph
\[S(c,x,y_0) = \mathrm{Reach}\big(G_{\mathrm{ent}},\big(\mathsf{touch}(x,y_0)\big)\big)\]
optionally restricted by the import DAG to a budgeted neighborhood. We then serialize each element of $S$ with metadata as to its object type (e.g. lemma, definition, etc.), as well as a stable snippet of its source content.

Because selection is driven by AST resolution, $\Psi_{\text{ctx}}$ is invariant to superficial edits (whitespace, formatting) and robust to local refactors ($\alpha$-renaming within a module). The slice size grows with the reachable subgraph, not raw file size, yielding a compact, minimal, and deterministic bundle of proof context that is read-only with respect to the program state.

\section{Replay Buffer and Dataset Construction Details}
\label{app:replay_buffer}

We provide a more detailed overview of the replay buffer algorithm. Given the problem set $\mathcal{P} = \mathcal{P}_{\mathrm{train}} \cup \mathcal{P}_{\mathrm{test}}$ and the current iteration $t$ model $G_t$, we run generation as described in \S\ref{sec:generation} to obtain $n$ candidate proofs per problem in $\mathcal{P}_{\mathrm{train}}$. This yields the raw (no-replay) dataset:
\begin{align*}
    \mathcal{D}^{(t)}_{\mathrm{nr}}&=\{(c_T,x_T,y_{T,0},\{(y_{T,i},\widehat{s}_{T,i})\}_{i=1}^{n})\}_{T \in \mathcal{P}_{\mathrm{train}}}\\
    &=\{(c_T,x_T,y_{T,0},\mathcal{Y}_T^{(t)})\}_{T \in \mathcal{P}_{\mathrm{train}}}
\end{align*}
where $\widehat{s}_{T,i} = S_\mu(y_{T,i},y_{T,0}\|c_T,x_T)$ is the improvement score of candidate $y_{T,i}$ over the baseline $y_{T,0}$.

We then construct the solutions (replay) dataset $\mathcal{D}^{(t)}_{\mathrm{re}}$ by post-processing $\mathcal{D}^{(t)}_{\mathrm{nr}}$ together with the previous iteration's already post-processed dataset $\mathcal{D}^{(t-1)}_{\mathrm{re}}$. The procedure is parameterized by a target replay proportion $\rho\in[0,1]$, replay mode \texttt{mode}\,$\in\{\texttt{mark},\texttt{join},\texttt{replace}\}$,  an improvement-rate cap $\pi_{\max}\in[0,1]$, and a minimum gap $\gamma \in [0,1]$.

% JA: The previous paragraph mentions a replay mode. I could not find a description of how, when, and where it is used.
% (Tate): I believe this is another hyperparameter that we forgot to introduce. I will mention this earlier.

\paragraph{Improvement rate and eligibility.}
Additionally, for problem $T$, define its improvement rate at iteration $j$ by
\[
\pi_T^{(j)}=\frac{1}{n}\sum_{i=1}^n \mathbb{I}\!\left[\text{v}(c_T,x_T,y_{T,i}^{(j)})=1\ \wedge\ \widehat{s}_{T,i}^{(j)}>0\right].
\]
A problem is replay-eligible at iteration $t$ iff it had at least one improved, compiling solution in some previous iteration $j<t$. We maintain a reservoir $\mathcal{E}$ of replay-eligible problems with preference for ``easy'' items (largest $\pi_T^{(j)}$ across $j<t$).

\paragraph{Replay buffer construction}

We proceed now to construct the post-replay buffer dataset in two main steps:

\begin{enumerate}
    \item \textit{Mark:} We first mark problems in $\mathcal{D}^{(t)}_{\mathrm{nr}}$ as \textsc{replay} or \textsc{frontier} so that the fraction of replay equals $\rho$:
\begin{enumerate}
\item Start with the subset of problems that are replay-eligible; if their fraction exceeds $\rho$, downsample them by removing highest-$\pi_T$ items (i.e. the ``easiest'') until the replay fraction is $\rho$ (cap at $\pi_{\max}$).
\item If their fraction is below $\rho$, replace uniformly chosen frontier items by items sampled from the reservoir $\mathcal{E}$ (taken from $\mathcal{D}^{(t-1)}_{\mathrm{re}}$) until the replay fraction reaches $\rho$ as best as possible. This keeps dataset size fixed while achieving the target mix.
\end{enumerate}
After this step, every item in the dataset is tagged as \textsc{replay} or \textsc{frontier} and the target mix is satisfied.

\item \textit{Merge}

% \paragraph{Step 2 — \textsf{join} or \textsf{replace}.}
For each item marked \textsc{replay} with key $(c_T,x_T,y_{T,0})$, find its counterpart in $\mathcal{D}^{(t-1)}_{\mathrm{re}}$, say with candidate (multi)set $\mathcal{\widetilde{Y}}^{(t-1)}_T$. Then, we case on \texttt{mode} as follows:
\begin{itemize}
\item \texttt{join}: set the current candidates to the union
$\mathcal{\widetilde{Y}}^{(t)}_T = \mathcal{Y}^{(t)}_T \cup \mathcal{\widetilde{Y}}^{(t-1)}_T$
(deduplicated by normalized proof text). This increases candidate diversity for IRPO.
\item \texttt{replace}: set
$\mathcal{\widetilde{Y}}^{(t)}_T = \mathcal{\widetilde{Y}}^{(t-1)}_T$,
i.e., overwrite the current candidates by the previous iteration's.
\end{itemize}
If \texttt{mode} is \texttt{mark}, skip this step (no candidate-level splice). In other words, set $\mathcal{\widetilde{Y}}^{(t)}_T = \mathcal{Y}^{(t)}_T$

\end{enumerate}

With this, we obtain the replay dataset:
\[
\mathcal{D}^{(t)}_{\mathrm{re}}
=\Big\{\,\big(c_T,x_T,y_{T,0},\ \mathcal{\widetilde{Y}}^{(t)}_T\big)\ \Big|\ T\in\mathcal{P}_{\mathrm{train}}\Big\}
\]

\paragraph{Filtering}

Finally, we post-process $\mathcal{D}^{(t)}_{\mathrm{re}}$ by filtering out low-quality candidates and separating the samples into winner ($W$) and loser ($L$) sets. Specifically, we filter the high improvement rate problems (those which were sufficiently ``easy'' for the model to improve on many candidates) by removing problems $T$ with $\pi_T^{(t)} > \pi_{\max}$. Namely, we define $\widetilde{\mathcal{P}}^{(t)} = \mathcal{P}_{\mathrm{train}} \setminus \{T \mid \pi_T^{(t)} > \pi_{\max}\}$ to be the filtered problem set.

Then, for each problem $T$ in $\widetilde{\mathcal{P}}^{(t)}$, we partition $\mathcal{\widetilde{Y}}^{(t)}_T = W_T^{(t)} \cup L_T^{(t)}$ such that:
\begin{itemize}
\item $W_T^{(t)} = \{y \in \mathcal{\widetilde{Y}}^{(t)}_T \mid \text{v}(c_T,x_T,y) = 1 \land \widehat{s}_{T,y} > \delta^{(t)}\}$, where $\delta^{(t)}$ is the $\gamma$-th percentile of all scores $\{\widehat{s}_{T,i} : T \in \widetilde{\mathcal{P}}^{(t)}, y_i \in \mathcal{\widetilde{Y}}^{(t)}_T\}$.
    \item $L_T^{(t)} = \mathcal{\widetilde{Y}}^{(t)}_T \setminus W_T^{(t)}$.
\end{itemize}
In other words, winners are those candidates that compile and have an improvement score above the $\gamma$-th percentile threshold across all samples in the dataset, while losers are the rest.

With this, we finalize the filtered post-replay buffer dataset as:
\[
\mathcal{D}_{\text{fil}}^{(t)} = \Big\{(c_T,x_T,y_{T,0},W_T^{(t)},L_T^{(t)}) \mid T \in \widetilde{\mathcal{P}}^{(t)}\Big\}
\]

\paragraph{IRPO Dataset}

As IRPO is a preference-based training method, we now convert $\mathcal{D}_{\text{fil}}^{(t)}$ into a set of preference pairs. This process is parameterized by two hyperparameters $W,L \in \mathbb{N}$, which control the number of winners and losers to sample per problem, respectively.

First, for a given problem $T$ in $\widetilde{\mathcal{P}}^{(t)}$, we deduplicate $W_T^{(t)}$ by equal scores $\widehat{s}_{T,i} = \widehat{s}_{T,j}$ and deduplicate $L_T^{(t)}$ by string equality.

With this, we form two families of preference pairs:
\[
\mathfrak{P}^{\ell\to w}_T=\{(b,g): b\in L_T^{(t)},\ g\in W_T^{(t)}\}\]\[
\mathfrak{P}^{w\to w}_T=\{(g',g): g,g'\in W_T^{(t)},\ \widehat{s}_{T,g}>\widehat{s}_{T,g'}\}.
\]

And with this, we form our IRPO training dataset as the following set:
\[\mathcal{D}_{\text{IRPO}}^{(t)} = \bigcup_{T \in \widetilde{\mathcal{P}}^{(t)}} (\mathfrak{P}^{\ell\to w}_T \cup \mathfrak{P}^{w\to w}_T)\]
Which can be reinterpreted elementwise as a collection of tuples $(c_T,x_T,y_{T,0},y_{T,\ell},y_{T,w})$.

\begin{algorithm}
  \caption{Preference Pair Creation}
  \label{alg:pref_pair_creation}
  \begin{algorithmic}
    \STATE {\bfseries Input:} filtered dataset $\mathcal{D}_{\text{fil}}^{(t)}$, hyperparameters $W \in \mathbb{N}$ and $L \in \mathbb{N}$
    \STATE Initialize $\mathcal{D}_{\text{IRPO}}^{(t)} = [ \ ]$
    \FOR{$T$ {in} $\mathcal{D}_{\text{fil}}^{(t)}$}
        \STATE De-duplicate $W_T^{(t)}$ by total improvement score
        \STATE De-duplicate $L_T^{(t)}$ by string equality
        \STATE Discard or duplicate elements uniformly at random until $|W_T^{(t)}|=W$ and $|L_T^{(t)}|=L$
        \FOR{$w_1$ in $W_T^{(t)}$}
            \FOR{$w_2$ in $W_T^{(t)}$}
                \IF{$\mu(c, x, w_1) > \mu(c, x, w_2)$}
                    \STATE $\mathcal{D}_{\text{IRPO}}^{(t)} := (w_1, w_2) \texttt{::} \mathcal{D}_{\text{IRPO}}^{(t)}$
                \ENDIF
            \ENDFOR
        \ENDFOR
        \FOR{$w$ in $W_T^{(t)}$}
            \FOR{$l$ in $L_T^{(t)}$}
                \STATE $\mathcal{D}_{\text{IRPO}}^{(t)} := (w, l) \texttt{::} \mathcal{D}_{\text{IRPO}}^{(t)}$
            \ENDFOR
        \ENDFOR
    \ENDFOR
  \end{algorithmic}
\end{algorithm}

\section{Chain-of-States Implementation}
\label{app:annotation_cos}
Given a verified proof $y_0$, let its tactic steps be indexed by $i=1,\dots,T$. From the \texttt{InfoTree} we obtain for each step $(\texttt{goalsBefore}_i,\texttt{goalsAfter}_i)$ and a pretty-printed tactic. We define
\[
\Psi_{\text{cos}}(y_0)\ =\ \big\langle\,(\texttt{goalsBefore}_i,\ \tau_i,\ \texttt{goalsAfter}_i)\,\big\rangle_{i=1}^T,
\]
and serialize each triple as a short snippet where goal lists are pretty-printed into comments adjacent to $\tau_i$. This turns hidden kernel states into explicit cues the model can condition on.

\section{System Configuration}
\label{app:config}

% \subsection{Evaluation}

\subsection{Compute Resources}

All local model training and evaluation runs were performed on a machine with 8 NVIDIA L40S GPUs. Local evaluations used this same GPU pool for batched best@$n$ generation and Lean verification. Closed-source and hosted API baselines were accessed through OpenRouter; for these models, compute was provided by the API provider rather than by our local hardware. We report the inference token costs used for the API baselines in Table~\ref{tab:frontier_and_intrafamily}.

\subsection{Training}
\label{app:training_config}

Training uses the same fixed systems configuration across metrics and IRPO rounds, while the main optimization hyperparameters are selected separately for each metric and iteration by validation grid search. In particular, the preference gap, replay-buffer ratio, learning rate, and related selection thresholds are tuned per round; the resulting searches are reported in Figures~\ref{fig:grid_len_all} and~\ref{fig:grid_dep_all}. Unless otherwise stated, each training run uses the following fixed configuration:

\begin{itemize}
    \item micro-batch size: 1 example per GPU, with gradient accumulation 1, for an effective batch size of 8 examples across 8 GPUs;
    \item epochs: 1;
    \item optimizer: \texttt{adamw\_torch};
    \item learning-rate scheduler: cosine, with 5 warmup steps;
    \item weight decay: 0.0 and maximum gradient norm: 1.0;
    \item precision and memory settings: bf16, FlashAttention-2, gradient checkpointing, and DeepSpeed ZeRO-3 CPU offload.
\end{itemize}

\begin{figure*}
\centering
\begin{subfigure}{0.6\linewidth}
    \includegraphics[width=\textwidth]{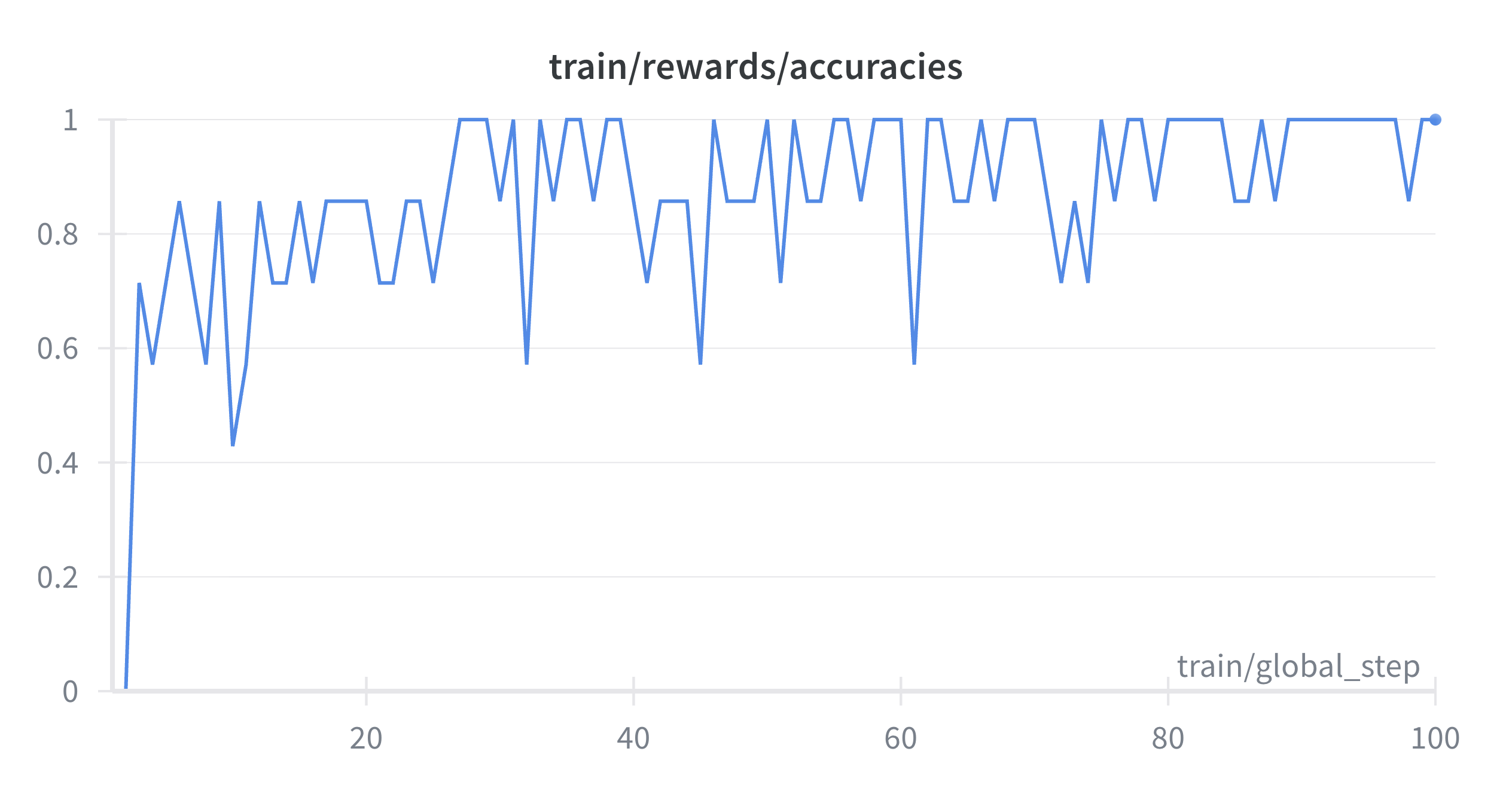}
    \caption{Training accuracy over time on iteration 3 of the dependency metric.}
    \label{fig:IRPO_Dependency_Iter_3_Acc}
\end{subfigure}
\hfill
\begin{subfigure}{0.6\linewidth}
    \includegraphics[width=\textwidth]{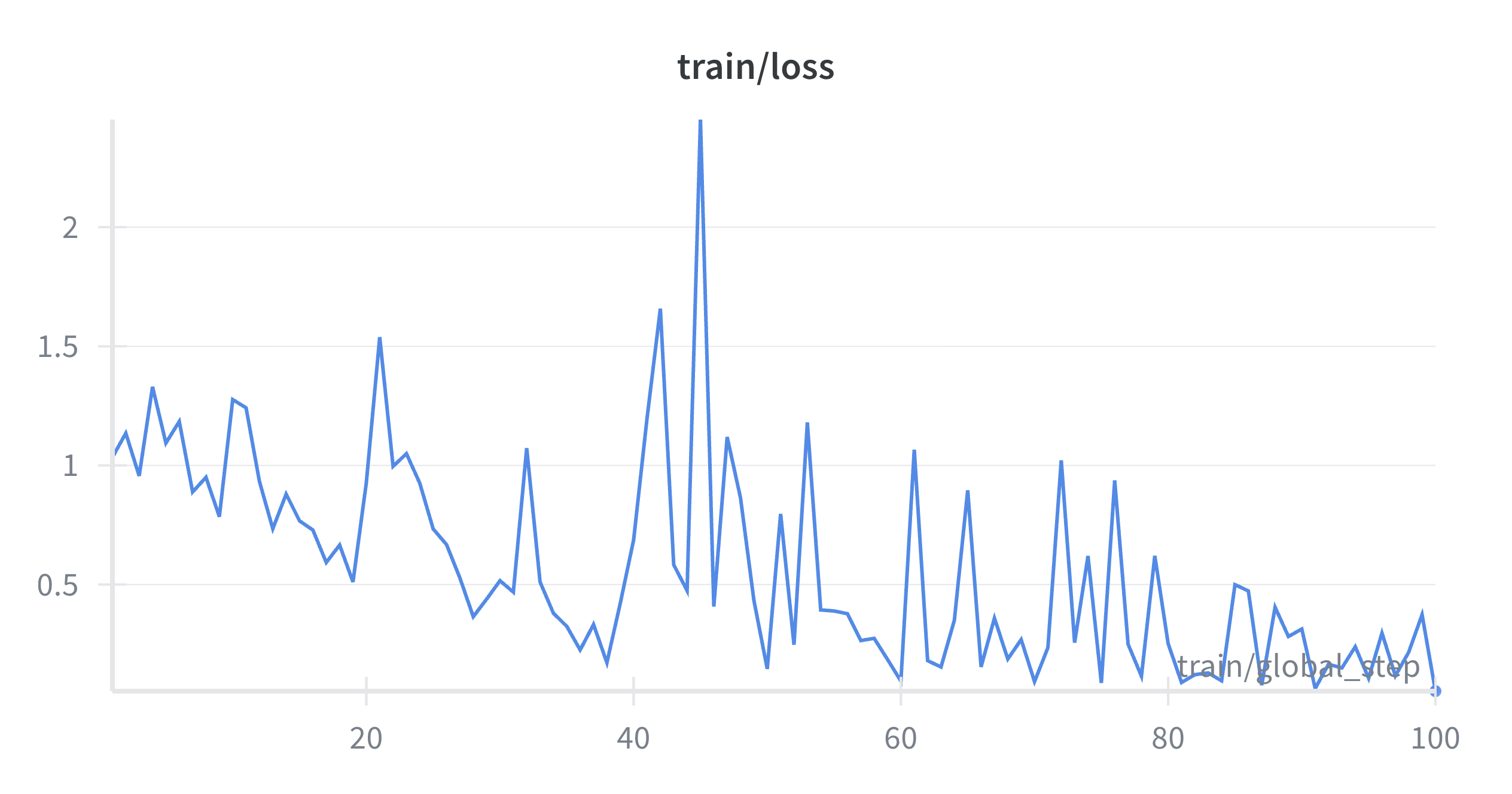}
    \caption{Training loss over time on iteration 3 of the dependency metric. Loss exhibits an overall downward trend. }
    \label{fig:IRPO_Dependency_Iter_3_Loss}
\end{subfigure}
\hfill
\begin{subfigure}{0.6\linewidth}
    \includegraphics[width=\textwidth]{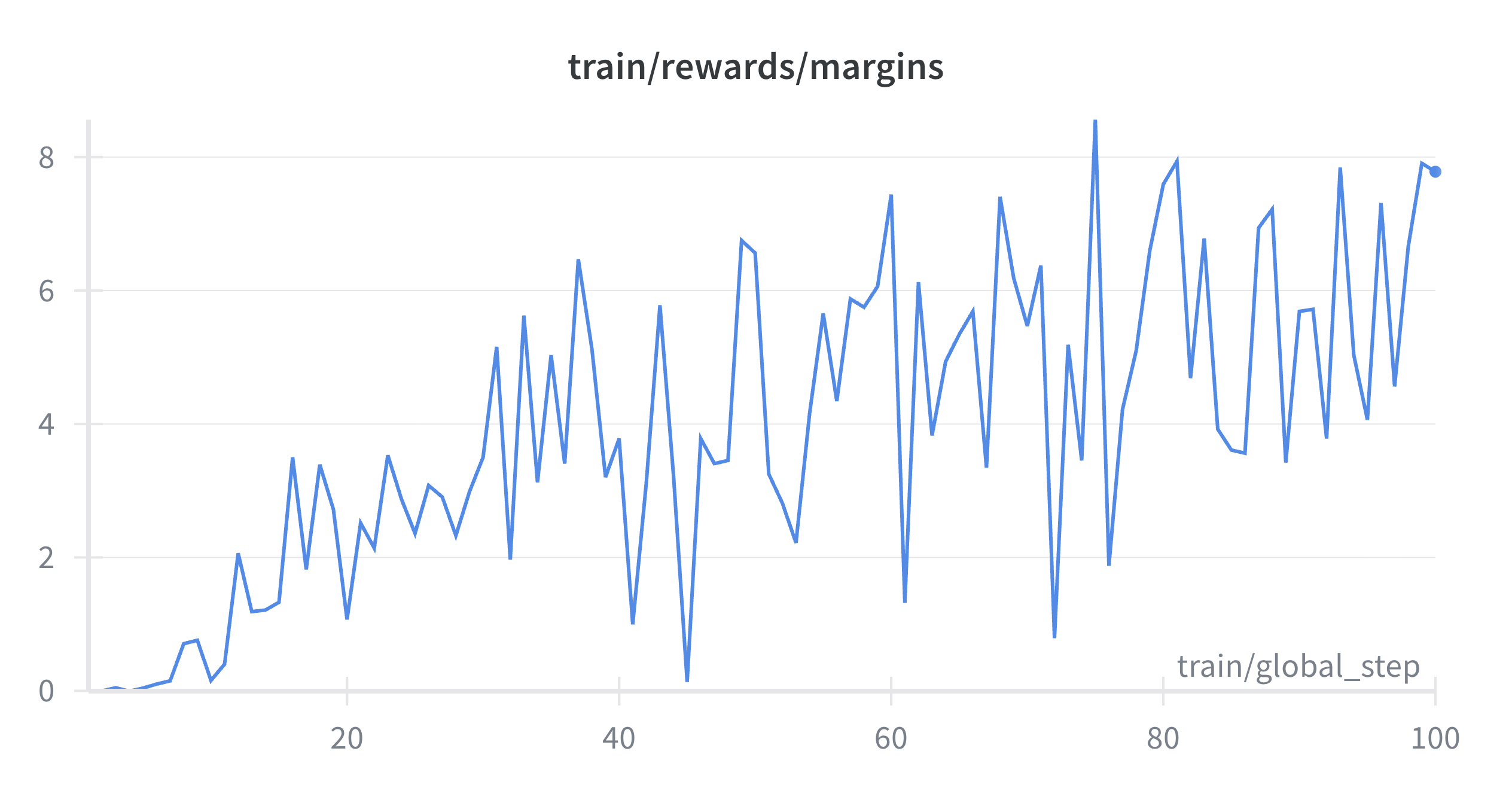}
    \caption{Average margin over time during training on iteration 3 of the dependency metric. }
    \label{fig:IRPO_Dependency_Iter_3_Margin}
\end{subfigure}
\caption{Training statistics during iteration 3 of the dependency metric. Model shows stable performance despite overall regression on this iteration. }

\end{figure*}

Figure~\ref{fig:IRPO_Dependency_Iter_3_Acc}--\ref{fig:IRPO_Dependency_Iter_3_Margin} shows representative training traces for the third dependency-optimization round.

\subsection{System Prompts}

During generation, \ImProver \ is prompted by combining the following prompts: depending on the metric, we take one of the \textbf{length}, \textbf{modularity}, or \textbf{dependency} prompts, and append the \textbf{annotation}, \textbf{context}, and \textbf{examples} prompts along with the relevant neurosymbolic augmentation in the correct locations.

\paragraph{Length:}

\tiny\texttt{You are an expert Lean4 theorem rewriting assistant. Shorten the current Lean4 theorem (wrapped in <CURRENT>...</CURRENT>) to be as short as possible in length - measured in the number of tactics in the proof - while also ensuring that the output is still a correct proof of the theorem. Be sure to output your final response as a Lean4 theorem wrapped in <IMPROVED>...</IMPROVED> tags, as shown in the example. Namely, only return the statement and proof of the current theorem in Lean4 code, wrapped in <IMPROVED>...</IMPROVED> tags. Do not include any other text or comments.}

\paragraph{Modularity:}

\tiny\texttt{You are an expert Lean4 theorem rewriting assistant. Given a Lean4 theorem enclosed in <CURRENT>...</CURRENT> tags, rewrite the theorem to be as modular and declarative as possible. Modularity is defined by the number of independent, meaningful subproofs, measured by the occurrence of tactics that spawn new goals (such as 'have' statements, case splits, automation tactics, or 'calc' blocks) -- with the caveat that these subproofs must be nontrivial and contribute to the overall proof structure. That means any sort of duplicate, unused, or trivial spawned goals will be ignored and/or penalized; this includes trivial modifications to spawned goals such as changing binders into forall statements. etc. Your objective is to maximize the number of these useful, nontrivial, and interesting spawned subproofs, while ensuring that the theorem remains correct and is clearly structured and readable. Optimize and rewrite the proof structure based on genuine sub-arguments, not superficial goal spawning. Validate that the revised theorem remains correct and that improvements in modularity are nontrivial and significant. In your output, only provide the improved statement and proof, wrapped in <IMPROVED>...</IMPROVED> tags, with no additional text or comments. Under no circumstances should you create artificial or superficial modularity in order to optimize for or maximize a reward metric; prioritize exclusively genuine mathematical quality and proof clarity over any gamified optimization.}

\paragraph{Dependency:}

\tiny\texttt{You are an expert Lean4 theorem rewriting assistant. Rewrite the current Lean4 theorem (wrapped in <CURRENT>...</CURRENT>) to be as independent of external theorems and lemmas as possible. Namely, you aim to rewrite the proof to minimize the number of external dependencies - while also ensuring that the output is still a correct proof of the theorem. Be sure to output your final response as a Lean4 theorem wrapped in <IMPROVED>...</IMPROVED> tags, as shown in the example. Namely, only return the statement and proof of the current theorem in Lean4 code, wrapped in <IMPROVED>...</IMPROVED> tags. Do not include any other text or comments.}

\paragraph{Annotation:}

\tiny\texttt{A version of the current theorem with the goal states annotated has also been provided for reference (wrapped in <ANNOTATED>...</ANNOTATED>). Namely, the goal states have been interleaved between tactics as comments to help you better understand the proof and ensure the correctness of your response. Do not include such state comments in your final response.}

\paragraph{Context:}

\tiny\texttt{The proof context, with relevant definitions and theorems, has additionally been provided to help you better understand the proof and ensure the correctness of your response. It is wrapped in <CONTEXT>...</CONTEXT>, with each item wrapped in <ITEM>...</ITEM>.}

\paragraph{Examples:}

\tiny\texttt{Here are some examples of such optimization, as wrapped in <EXAMPLES>...</EXAMPLES>. Note that these examples are for illustrative purposes only and should not be copied directly. Instead, use them to understand the kind of optimization expected and apply similar techniques to the current theorem, using these positive examples as an intuition and guidance on what kinds of optimizations you may do on your current target theorems. Additionally, you will also be provided with a negative example which will be marked as such. Use it to understand common pitfalls and avoid them in your response. Use both the positive and negative examples to guide your optimization of the current theorem.}

\subsection{Autoinformalization}
\label{app:informal}

\normalsize During the generation round at iteration $t=0$, we create informal statements of each theorem for use in neurosymbolic augmentation. This is carried out by prompting the base model $G_0$ with the following:

\tiny\texttt{You are an expert informalizer of formal mathematics to natural language. Namely, given a formal theorem and proof in Lean4,
you will generate an informalized statement of this same theorem in natural language, as well as (2) an informalized, natural language version of the same formal proof
that is aligned with the informal statement. Namely, when informalizing the proof, you should convert each tactic of the formal proof into a natural language step in the informal proof, and thereby, your informal proof should be written as a sequence of steps. Consider the following example:}

\textit{(examples omitted)}

\tiny\texttt{Now, with these examples in mind, it is now your turn to informalize the following formal statement and proof, which is wrapped in <FORMAL>...</FORMAL> tags.}

\tiny\texttt{You may think and reason as much as you want, but ensure that your final answer for (1): the informal statement is wrapped in <STATEMENT>...</STATEMENT> tags, and (2): the informal proof is wrapped in <PROOF>...</PROOF> tags.
Your final answer should have both a <STATEMENT>...</STATEMENT> tag and a <PROOF>...</PROOF> tag, and if there is no formal proof provided in the input, you may simply output <PROOF></PROOF> for the proof after informalizing the statement (i.e. if you are given a theorem without a proof, or a definition/class/etc.).
Input:
<FORMAL>}

\textit{(formal statement of proof is placed here)}

\tiny\texttt{</FORMAL>}

\normalsize The final informal statements are then parsed out for use in neurosymbolic augmentation.

% \newpage
\onecolumn

\section{Additional Experiments and Analyses}
\label{app:additional_exp_and_analyses}

\subsection{Ablation Study}
\label{app:ablations}

In addition to experiments presented in \ref{sec:main_results}, we study the effect of each source of neurosymbolic augmentation on model performance, finding that each added channel increases length-metric performance (Figure \ref{fig:neuro_ablations}). Additionally, we perform hyperparameter searches at each iteration for both the length (Figure \ref{fig:grid_len_all}) and dependency metric (Figure \ref{fig:grid_dep_all}).

\subsubsection{Neurosymbolic Ablations}

\begin{figure}[th]
    \centering
    \includegraphics[width=0.85\linewidth]{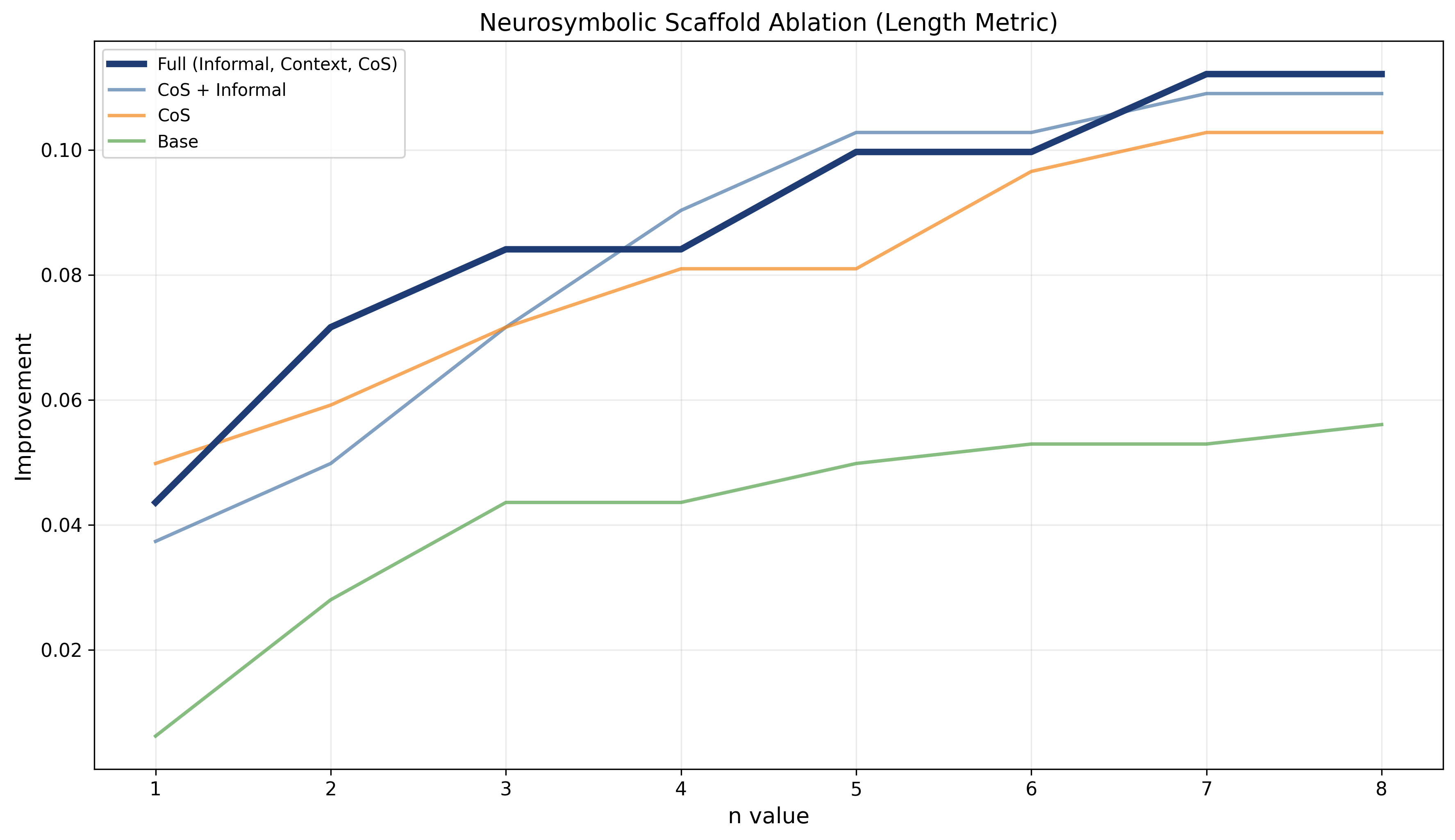}
    \caption{Effect of neurosymbolic augmentation on base model performance (\texttt{DeepSeek-R1-Distill-Qwen-7B}) on the length metric, vs. number of samples generated. Each source of augmentation shows noticeable improvement in average score on some $n$ values. }
    \label{fig:neuro_ablations}
\end{figure}

Figure~\ref{fig:neuro_ablations} shows that the unaugmented base model is consistently the weakest generator across sample budgets. Adding chain-of-states information alone produces a substantial improvement, indicating that exposing intermediate Lean goals gives the model useful local structure for proof rewriting. Adding informalized statements and proofs marginally improves performance across much of the curve, and the full scaffold, which additionally includes retrieved context, reaches the best performance at larger values of $n$. However, we observe that the majority of this improvement comes from the initial chain-of-states annotation. The gap is especially clear at best@8, where the full scaffold roughly doubles the improvement of the unaugmented base model. These results support the claim that neurosymbolic augmentation is not merely a prompting detail, but a central part of making proof optimization learnable and sample-efficient for small models.

\subsubsection{Grid Search Results}

\begin{figure}[t]
\centering
\captionsetup{skip=4pt}

\begin{subfigure}[t]{\linewidth}
  \centering
  \includegraphics[width=0.65\linewidth]{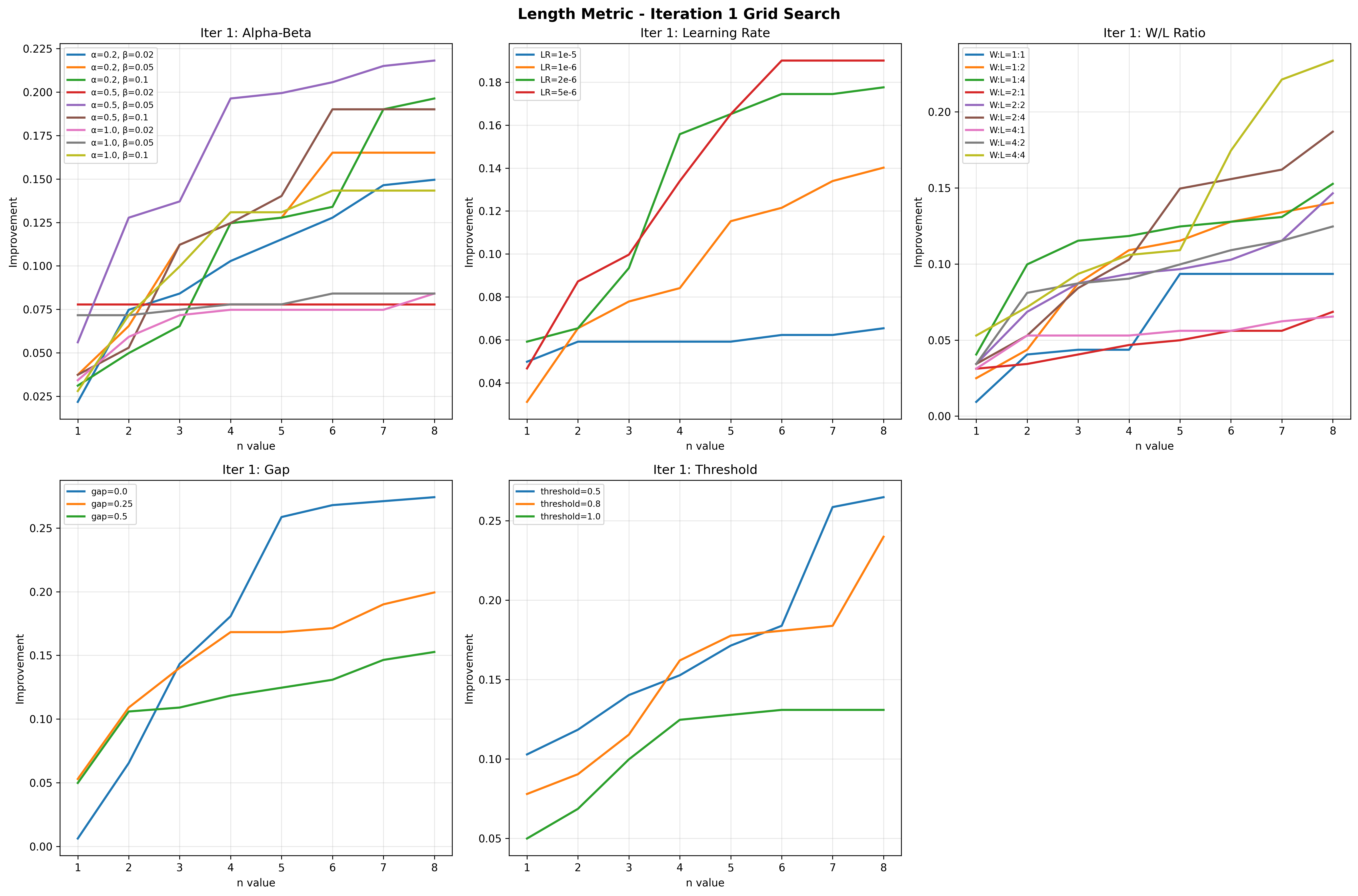}
  % \caption{Iteration 1.}
\end{subfigure}

% \vspace{-6pt}

\begin{subfigure}[t]{\linewidth}
  \centering
  \includegraphics[width=0.65\linewidth]{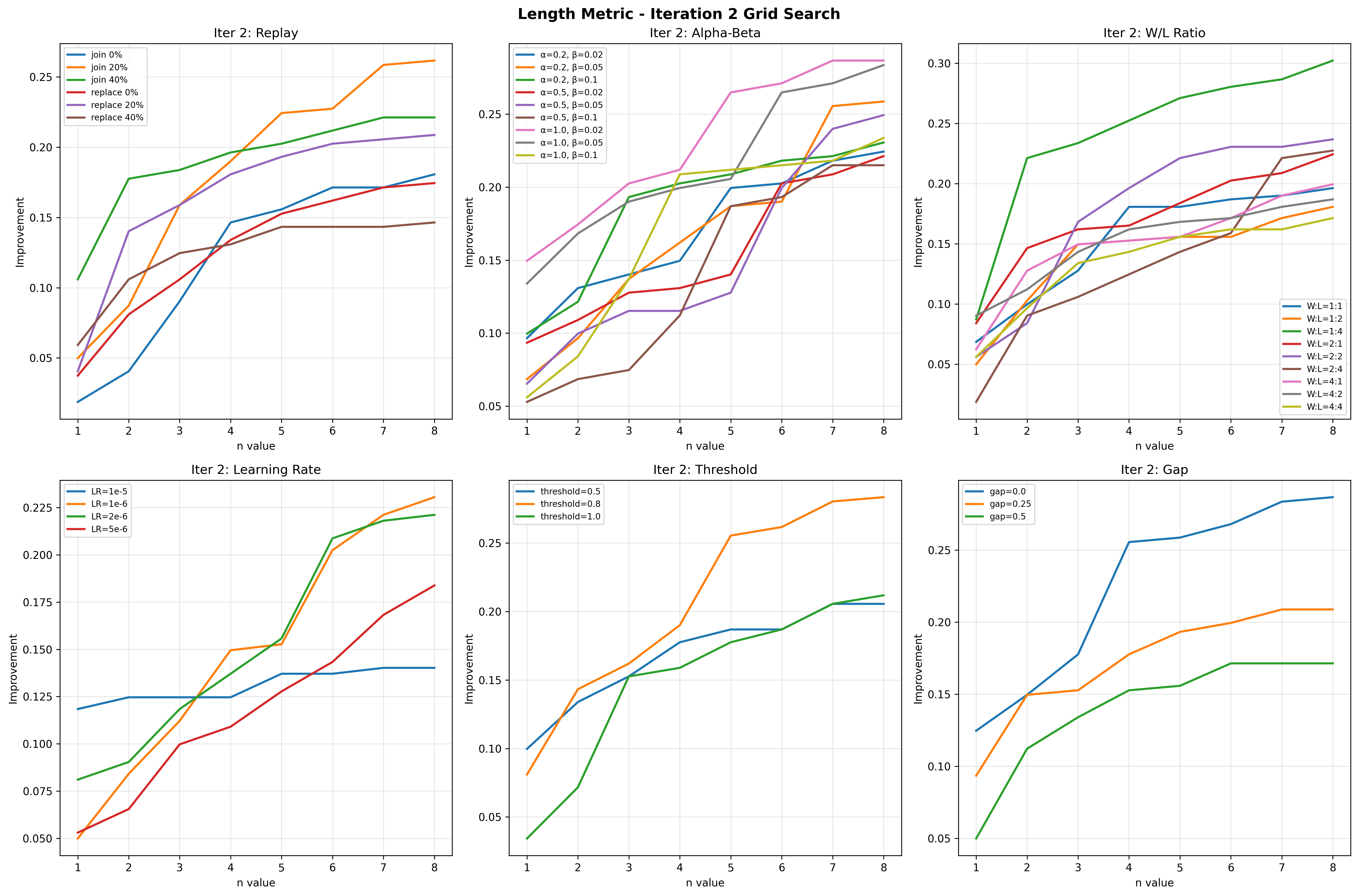}
  % \caption{Iteration 2.}
\end{subfigure}

% \vspace{-6pt}

\begin{subfigure}[t]{\linewidth}
  \centering
  \includegraphics[width=0.65\linewidth]{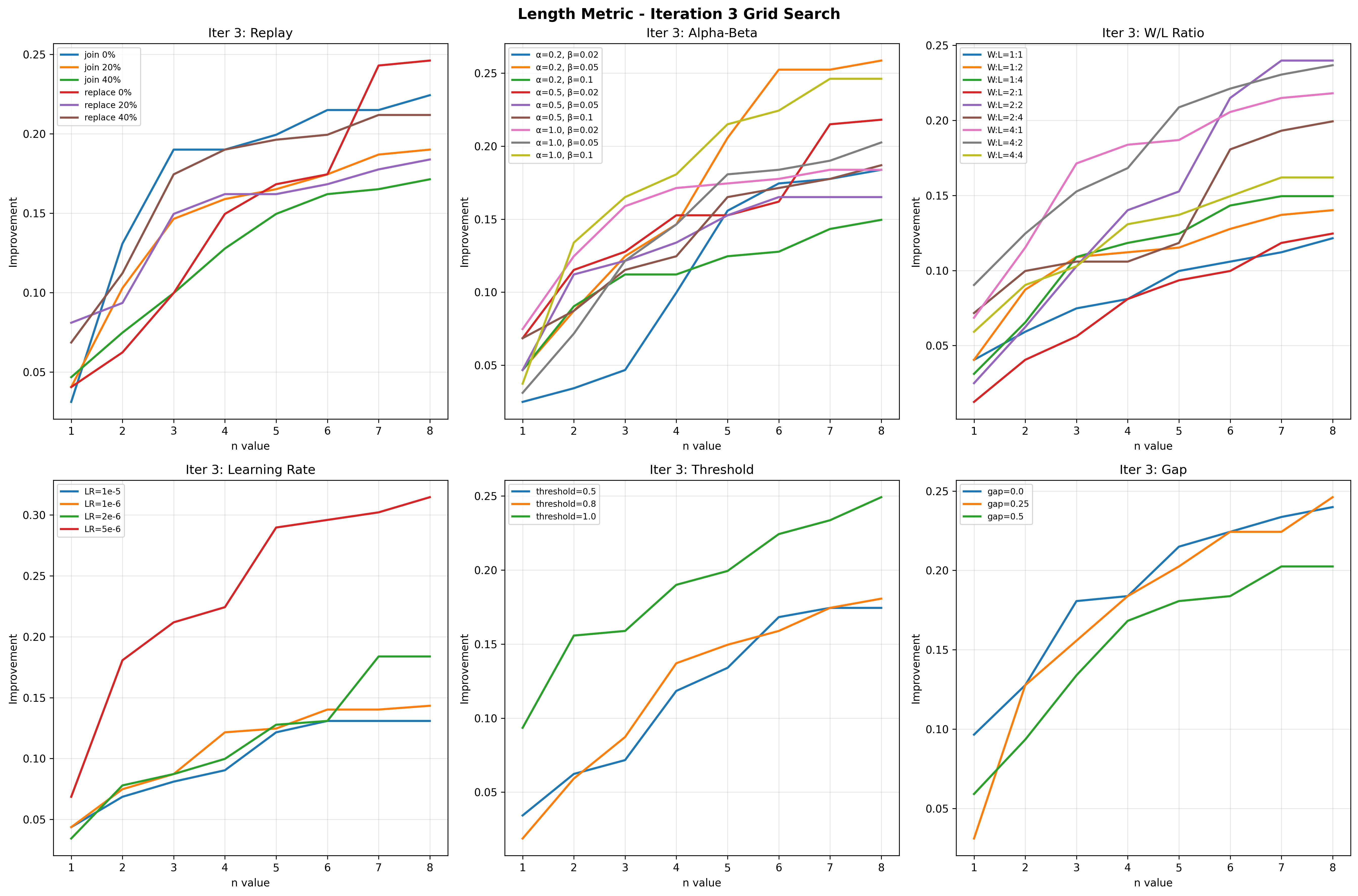}
  % \caption{Iteration 3.}
\end{subfigure}

\caption{Hyperparameter grid searches for the \textbf{length} metric across iterations 1--3.}
\label{fig:grid_len_all}
\end{figure}
\FloatBarrier
% \newpage

\begin{figure}[t]
\centering
\captionsetup{skip=4pt}

\begin{subfigure}[t]{\linewidth}
  \centering
  \includegraphics[width=0.65\linewidth]{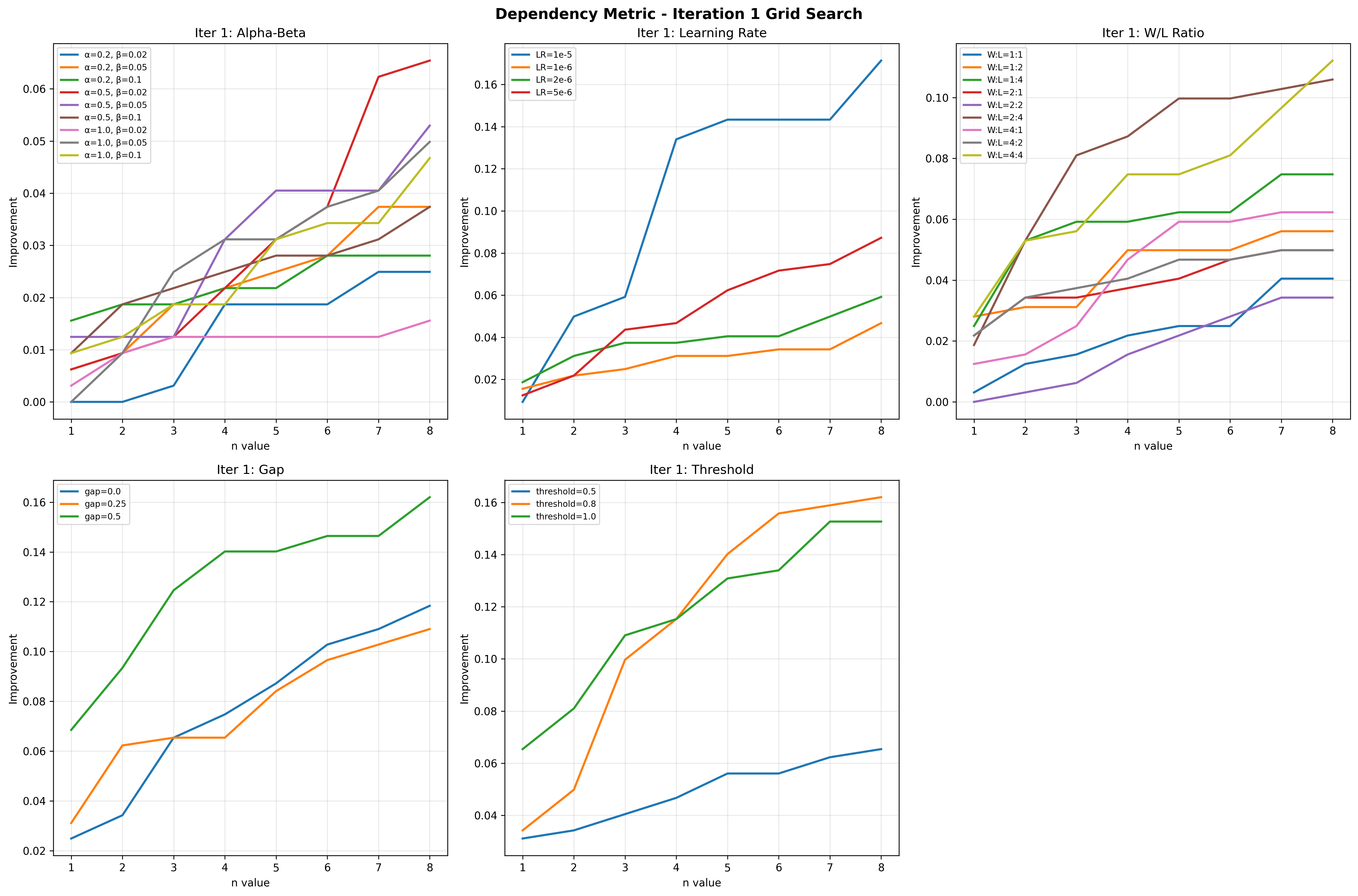}
  % \caption{Iteration 1.}
\end{subfigure}

% \vspace{-6pt}

\begin{subfigure}[t]{\linewidth}
  \centering
  \includegraphics[width=0.65\linewidth]{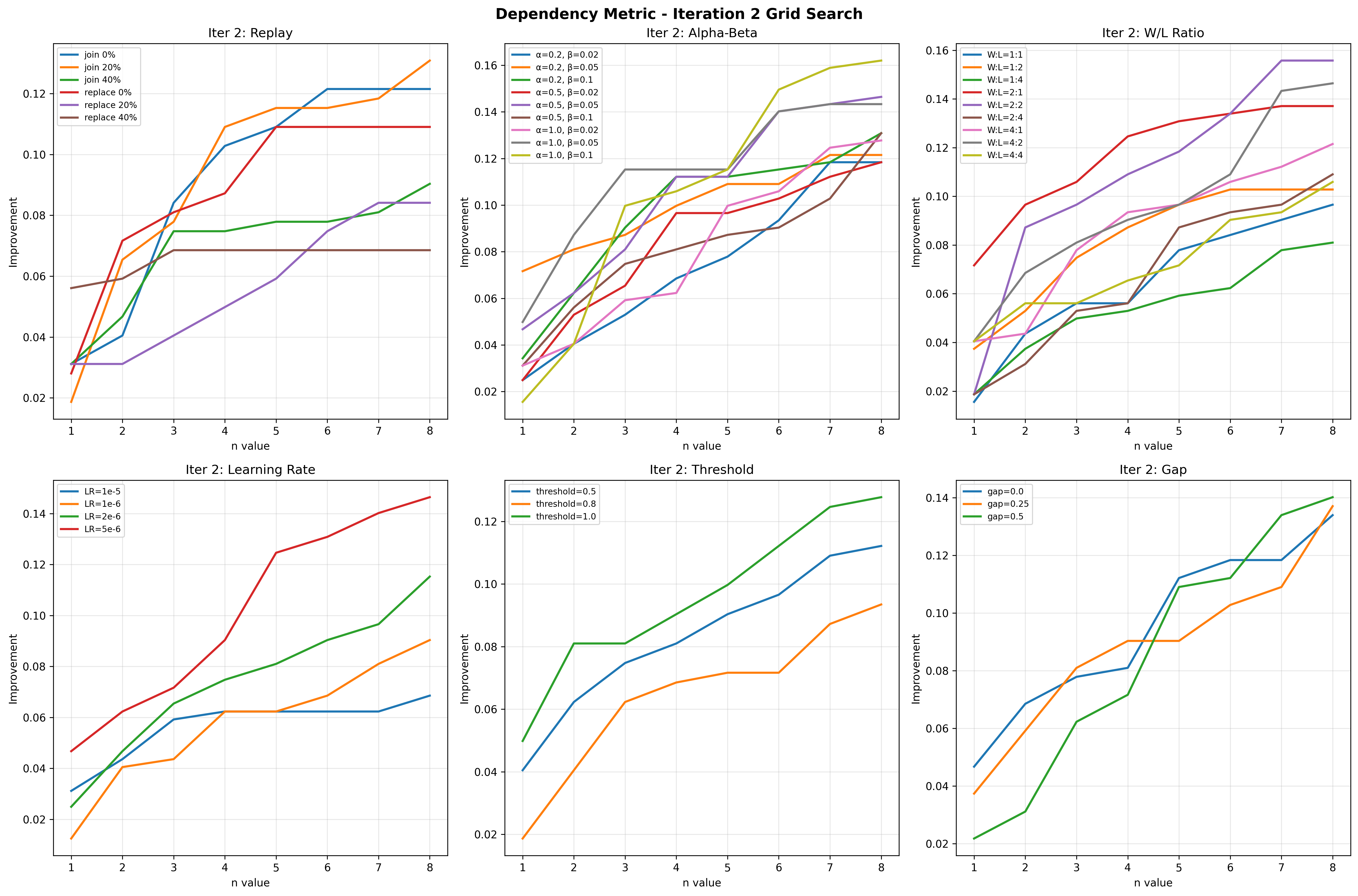}
  % \caption{Iteration 2.}
\end{subfigure}

% \vspace{-6pt}

\begin{subfigure}[t]{\linewidth}
  \centering
  \includegraphics[width=0.65\linewidth]{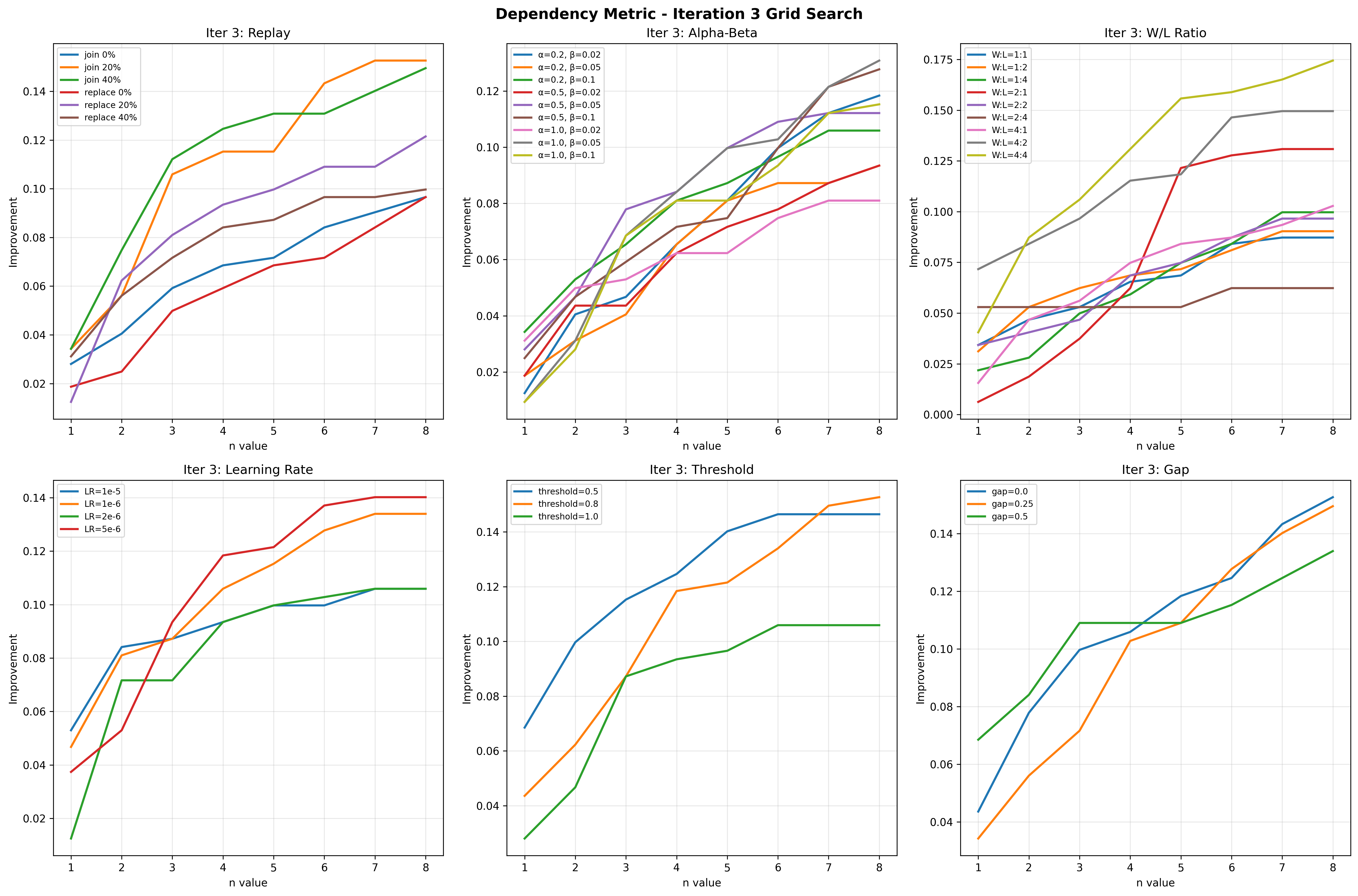}
  % \caption{Iteration 3.}
\end{subfigure}

\caption{Hyperparameter grid searches for the \textbf{dependency} metric across iterations 1--3.}
\label{fig:grid_dep_all}
\end{figure}

\FloatBarrier

Figures~\ref{fig:grid_len_all} and~\ref{fig:grid_dep_all} show that the training pipeline is sensitive to optimization hyperparameters, but that several qualitative patterns are stable. Increasing the sample budget generally improves scores for nearly all settings, so the relative comparisons are not artifacts of a single best@$n$ choice. For length optimization, lower preference gaps and moderate replay settings tend to perform well, suggesting that proof-shortening benefits from keeping a broad set of successful rewrites in the training signal. For dependency optimization, larger filtering gaps and more selective thresholds are often stronger, consistent with the metric rewarding more targeted structural changes. Learning-rate and winner/loser-ratio effects vary across iterations, which is why we tune these settings separately for each metric and training round rather than reusing a single global configuration.

\subsection{Qualitative Examples}
\label{app:qualitative}

Figures~\ref{fig:qual-dep-coatom}--\ref{fig:qual-mod-alphaproof-p6} show representative examples from the held-out evaluation set. For each example, we report the source module and declaration together with the original and optimized metric values. The AlphaProof qualitative case studies use best@64 sampling rather than the best@16 budget used for the main MiniCTX-v2 evaluations. The original AlphaProof IMO 2024 release targets Lean v4.10.0; for these examples, we manually bumped the AlphaProof proofs to Lean v4.17.0. The bumped version is available at [Redacted], and the original release is available at \url{https://storage.googleapis.com/deepmind-media/DeepMind.com/Blog/imo-2024-solutions/index.html}.

\subsubsection{Dependency Optimization}

% Mathlib.Order.Atoms
% SetLike.isCoatom_iff
% 5.0
% 2.0
% -3.0
{
\setlength{\columnsep}{0.2cm}
\begin{paracol}{2}
{
\textbf{Original (human-written)}
\begin{lstlisting}[basicstyle=\tiny\ttfamily]
theorem isCoatom_iff [OrderTop A] {K : A} :
    IsCoatom K ↔ K ≠ T ∧ ∀ H g, K ≤ H → g ∉ K → g ∈ H → H = T := by
  simp_rw [IsCoatom, lt_iff_le_not_le, SetLike.not_le_iff_exists,
    and_comm (a := _ ≤ _), and_imp, exists_imp, ← and_imp, and_comm]
\end{lstlisting}
}
\switchcolumn
{
\textbf{\ImProver \ (dependency-optimized)}
\begin{lstlisting}[basicstyle=\tiny\ttfamily]
theorem isCoatom_iff [OrderTop A] {K : A} :
    IsCoatom K ↔ K ≠ T ∧ ∀ H g, K ≤ H → g ∉ K → g ∈ H → H = T := by
  constructor <;> intro h
  <;> simp_all [IsCoatom, lt_iff_le_not_le, SetLike.not_le_iff_exists]
  <;> tauto
\end{lstlisting}
}

\end{paracol}
\captionof{figure}{\ImProver \ optimizes a proof for dependencies, reducing explicit dependencies by 3. }
\label{fig:qual-dep-coatom}
}
% \end{figure}

Figure~\ref{fig:qual-dep-coatom} comes from \texttt{Mathlib.Order.Atoms}, declaration \texttt{SetLike.isCoatom\_iff}. The original proof has explicit dependency count $5$, while the optimized proof has count $2$. The rewrite replaces a long \texttt{simp\_rw} chain naming several transformations with a case split, broader simplification, and propositional reasoning via \texttt{tauto}; this reduces the explicit dependency footprint while preserving the same theorem statement.

% ConNF.External.Basic
% ConNF.mem_cross_iff
% 2.0
% 0.0
% -2.0

{
    \setlength{\columnsep}{0.2cm}

    \begin{paracol}{2}
{
\textbf{Original (human-written)}
\begin{lstlisting}[basicstyle=\tiny\ttfamily]
theorem mem_cross_iff (x y : TSet γ) :
    ∀ a, a ∈' cross hβ hγ hδ x y ↔ ∃ b c, a = ⟨b, c⟩' ∧ b ∈' x ∧ c ∈' y := by
  intro a
  rw [cross, mem_inter_iff, vCross_spec]
  constructor
  · rintro ⟨h₁, b, c, rfl, h₂⟩
    simp only [op_mem_converse_iff, vCross_spec, op_inj] at h₁
    obtain ⟨b', c', ⟨rfl, rfl⟩, h₁⟩ := h₁
    exact ⟨b, c, rfl, h₁, h₂⟩
  · rintro ⟨b, c, rfl, h₁, h₂⟩
    simp only [op_mem_converse_iff, vCross_spec, op_inj]
    exact ⟨⟨c, b, ⟨rfl, rfl⟩, h₁⟩, ⟨b, c, ⟨rfl, rfl⟩, h₂⟩⟩
\end{lstlisting}
\switchcolumn
\textbf{\ImProver \ (dependency-optimized)}
\begin{lstlisting}[basicstyle=\tiny\ttfamily]
theorem mem_cross_iff (x y : TSet γ) :
    ∀ a, a ∈' cross hβ hγ hδ x y ↔ ∃ b c, a = ⟨b, c⟩' ∧ b ∈' x ∧ c ∈' y := by
  intro a
  -- Use the definition of cross and simplify the membership conditions directly
  constructor <;> intro h
  -- First direction: Assume membership in cross, construct the pair
  <;> simp [cross] at h ⊢
  -- Second direction: Decompose the pair existence claim and verify conditions
  <;> aesop
  -- Handle remaining simple cases with basic reasoning
\end{lstlisting}
}
\end{paracol}
\captionof{figure}{\ImProver \ optimizes a proof for dependencies, reducing explicit dependencies by 2. }
\label{fig:qual-dep-cross}
% \end{figure}
}

Figure~\ref{fig:qual-dep-cross} comes from \texttt{ConNF.External.Basic}, declaration \texttt{ConNF.mem\_cross\_iff}. The dependency count decreases from $2$ to $0$. The optimized proof avoids explicitly naming the original rewrite lemmas and instead unfolds \texttt{cross} and delegates the remaining elementary cases to automation. This illustrates a common dependency-optimization behavior: replacing brittle named rewrite sequences with more local simplification and search.

{
\noindent\begin{minipage}[t]{0.49\textwidth}
\textbf{Original (AlphaProof)}
\begin{lstlisting}[style=leanfullproof]
theorem imo_2024_p6
    (IsAquaesulian : (ℚ → ℚ) → Prop)
    (IsAquaesulian_def : ∀ f, IsAquaesulian f ↔
      ∀ x y, f (x + f y) = f x + y ∨ f (f x + y) = x + f y) :
    IsLeast {(c : ℤ) | ∀ f, IsAquaesulian f → {(f r + f (-r)) | (r : ℚ)}.Finite ∧
      {(f r + f (-r)) | (r : ℚ)}.ncard ≤ c} 2 := by
  exists@?_
  ·
    useλu b=>if j:u 0=0then by_contra λc=>?_ else ?_
    ·
      suffices:({J|∃k,u k+u (-k)= J}) ⊆{0}
      · simp_all[this.antisymm]
      rintro - ⟨a, rfl⟩
      contrapose! c
      simp_all
      suffices:{U|∃examples6 : ℚ, u examples6 + u (-examples6)= U} ⊆{0,(u (a : Rat)+ (u<|@@↑(( (-a ))))) } ..
      · constructor
        · exact ( Set.toFinite ( _) ).subset (by simpa using this)
        · exact (Set.ncard_le_ncard (by simpa using this)).trans (Set.ncard_pair (Ne.symm (↑ ( (c)) ) )).le
      rintro-⟨hz, rfl⟩
      induction b @hz a
      ·
        have:=b (-a)$ hz+u a
        have:=b hz hz
        simp_all[add_comm]
        have:=b (-hz) (hz+u ↑(hz))
        simp_all[ add_assoc, C]
        induction this
        ·
          simp_all
          have:=b hz (hz+(u a+u (-a)))
          have:=b (hz+(u a+u (-a)))$ hz+(u a+u (-a))
          use .inr$ by_contra$ by hint
        have:=b hz$ hz+(u hz+u (-hz))
        cases b (hz+(u hz+u (-hz)))$ hz+(u hz+u (-hz))with|_=>hint
      have:=b (-hz) (u hz+a)
      have:=b$ -a
      specialize this (u hz+a)
      simp_all[ ←add_assoc]
      have:=b 0
      have:=b
      specialize b a a
      simp_all[add_comm]
      have:=(this<| -a) (↑a + (((u a))): (↑_ :((( _) ) ) )) ..
      simp_all[add_assoc]
      cases this
      ·
        simp_all
        contrapose! IsAquaesulian_def
        simp_all
        exfalso
        have:=this a (a+(u hz+u ( -hz)))
        simp_all[Ne.symm,Bool]
        have:=‹∀congr_arg G,_› (a+(u hz+u (-hz)))$ a+(u ↑hz+u ↑( -hz) )
        simp_all
      have:=this a (a +(u a+u (-a)))
      cases‹forall Jd S,_› (a+(u a+u (-a))) ( a + (u a +u ↑(-a)))with| _ =>hint
    simp_all
    cases b 0 0with|_=>exact absurd (b 0$ (0+(1 *(@(u ↑.((0) )))))^ 01: ↑ ((_)) ) (id$ (by(cases ( b (u 0) ( (u 0)))with|_ => continuity)))
  rintro K V
  specialize V $ λ N=>-N+2 *Int.ceil N
  specialize( V $ (IsAquaesulian_def _).mpr _)
  · simp_rw [ ←eq_sub_iff_add_eq']
    ring_nf
    use mod_cast@?_
    norm_num[<-add_mul,Int.ceil_eq_iff]
    useλc K=>(em _).imp (⟨by linarith[Int.ceil_lt_add_one c,Int.le_ceil K],.⟩) (by repeat use by linarith[.,Int.le_ceil c,or,Int.ceil_lt_add_one$ K])
  simp_all[Int.ceil_neg, ←add_assoc]
\end{lstlisting}
\end{minipage}\hfill
\begin{minipage}[t]{0.49\textwidth}
\textbf{\ImProver \ (dependencies-optimized)}
\begin{lstlisting}[style=leanfullproof]
theorem imo_2024_p6
    (IsAquaesulian : (ℚ → ℚ) → Prop)
    (IsAquaesulian_def : ∀ f, IsAquaesulian f ↔
      ∀ x y, f (x + f y) = f x + y ∨ f (f x + y) = x + f y) :
    IsLeast {(c : ℤ) | ∀ f, IsAquaesulian f → {(f r + f (-r)) | (r : ℚ)}.Finite ∧
      {(f r + f (-r)) | (r : ℚ)}.ncard ≤ c} 2 := by
  exists@?_
  · useλu b=>if j:u 0=0then by_contra λc=>?_ else ?_
    · suffices:({J|∃k,u k+u (-k)= J}) ⊆{0}
      · simp_all[this.antisymm]
      rintro - ⟨a, rfl⟩
      contrapose! c
      simp_all
      suffices:{U|∃examples6 : ℚ, u examples6 + u (-examples6)= U} ⊆{0,(u (a : Rat)+ (u<|@@↑(( (-a ))))) } ..
      · constructor
        · exact ( Set.toFinite ( _) ).subset (by simpa using this)
        · exact (Set.ncard_le_ncard (by simpa using this)).trans (Set.ncard_pair (Ne.symm (↑ ( (c)) ) )).le
      rintro-⟨hz, rfl⟩
      induction b @hz a
      · have:=b (-a)$ hz+u a
        have:=b hz hz
        simp_all[add_comm]
        have:=b (-hz) (hz+u ↑(hz))
        simp_all[ add_assoc, C]
        induction this
        · simp_all
          have:=b hz (hz+(u a+u (-a)))
          have:=b (hz+(u a+u (-a)))$ hz+(u a+u (-a))
          use .inr$ by_contra$ by hint
        have:=b hz$ hz+(u hz+u (-hz))
        cases b (hz+(u hz+u (-hz)))$ hz+(u hz+u (-hz))with|_=>hint
      have:=b (-hz) (u hz+a)
      have:=b$ -a
      specialize this (u hz+a)
      simp_all[ ←add_assoc]
      have:=b 0
      have:=b
      specialize b a a
      simp_all[add_comm]
      have:=(this<| -a) (↑a + (((u a))): (↑_ :((( _) ) ) )) ..
      simp_all[add_assoc]
      cases this
      · simp_all
        contrapose! IsAquaesulian_def
        simp_all
        exfalso
        have:=this a (a+(u hz+u ( -hz)))
        simp_all[Ne.symm,Bool]
        have:=‹∀congr_arg G,_› (a+(u hz+u (-hz)))$ a+(u ↑hz+u ↑( -hz) )
        simp_all
      have:=this a (a +(u a+u (-a)))
      cases‹forall Jd S,_› (a+(u a+u (-a))) ( a + (u a +u ↑(-a)))with| _ =>hint
    simp_all
    cases b 0 0with|_=>exact absurd (b 0$ (0+(1 *(@(u ↑.((0) )))))^ 01: ↑ ((_)) ) (id$ (by(cases ( b (u 0) ( (u 0)))with|_ => continuity)))
  rintro K V
  specialize V $ λ N=>-N+2 *Int.ceil N
  specialize( V $ (IsAquaesulian_def _).mpr _)
  · simp_rw [ ←eq_sub_iff_add_eq']
    ring_nf
    use mod_cast@?_
    norm_num[<-add_mul,Int.ceil_eq_iff]
    useλc K=>(em _).imp (⟨by linarith[Int.ceil_lt_add_one c,Int.le_ceil K],.⟩) (by repeat use by linarith[.,Int.le_ceil c,or,Int.ceil_lt_add_one$ K])
  let S : Set ℚ := {x | ∃ r : ℚ, (-r + 2 * ↑⌈r⌉) + (-(-r) + 2 * ↑⌈-r⌉) = x}
  have hsub : ({(0 : ℚ), 2} : Set ℚ) ⊆ S := by
    intro x hx
    simp at hx
    rcases hx with rfl | rfl
    · exact ⟨(-1 : ℚ), by norm_num⟩
\end{lstlisting}
\end{minipage}

% \clearpage

% \noindent\begin{minipage}[t]{0.49\textwidth}
% \textbf{Original (continued)}
% \begin{lstlisting}[style=leanfullproof]

% \end{lstlisting}
% \end{minipage}\hfill
% \begin{minipage}[t]{0.49\textwidth}
% \textbf{\ImProver \ (dependencies-optimized; continued)}
% \begin{lstlisting}[style=leanfullproof]

% \end{lstlisting}
% \end{minipage}

\clearpage
\noindent\begin{minipage}[t]{0.49\textwidth}
\textbf{Original (continued)}
\begin{lstlisting}[style=leanfullproof]
  suffices:2<=V.1.toFinset.card
  · let M:=V.1.toFinset
    have h2nat : 2 ≤ ({x | ∃ r : ℚ, 2 * ↑⌈r⌉ + -(2 * ↑⌊r⌋) = x} : Set ℚ).ncard := by
      rwa [Set.ncard_eq_toFinset_card _ V.1]
    have h2int : (2 : ℤ) ≤ ({x | ∃ r : ℚ, 2 * ↑⌈r⌉ + -(2 * ↑⌊r⌋) = x} : Set ℚ).ncard := by
      exact_mod_cast h2nat
    exact h2int.trans V.2
  use Finset.one_lt_card.2$ by
    refine ⟨0, V.1.mem_toFinset.2 ?_, 2, V.1.mem_toFinset.2 ?_, by norm_num⟩
    · exact ⟨-1, by norm_num⟩
    · exact ⟨(1 / 2 : ℚ), by norm_num⟩
\end{lstlisting}
\end{minipage}\hfill
\begin{minipage}[t]{0.49\textwidth}
\textbf{\ImProver \ (dependencies-optimized; continued)}
\begin{lstlisting}[style=leanfullproof]
    · exact ⟨(1 / 2 : ℚ), by norm_num⟩
  have h2nat : 2 ≤ S.ncard := by
    have hp : ({(0 : ℚ), 2} : Set ℚ).ncard = 2 := by
      simpa using (Set.ncard_pair (by norm_num : (0 : ℚ) ≠ 2))
    calc
      2 = ({(0 : ℚ), 2} : Set ℚ).ncard := hp.symm
      _ ≤ S.ncard := Set.ncard_le_ncard hsub V.1
  have h2int : (2 : ℤ) ≤ S.ncard := by
    exact_mod_cast h2nat
  exact h2int.trans V.2
\end{lstlisting}
\end{minipage}

\captionof{figure}{\ImProver \ optimizes an AlphaProof proof for dependencies, reducing explicit dependencies by 3.}
\label{fig:qual-dep-alphaproof-p6}
}

Figure~\ref{fig:qual-dep-alphaproof-p6} comes from \texttt{AlphaProof.P6}, declaration \texttt{imo\_2024\_p6}. The score changes from $18$ to $15$. The optimized proof keeps the same global proof search structure but removes a few explicit named dependencies by relying more on local facts and consolidated arithmetic/set-cardinality reasoning. However, overall, this case demonstrates negligible impact by ImProver2 on such a large-scale, complex problem.

\subsubsection{Length Optimization}
% HepLean.PerturbationTheory.FieldOpFreeAlgebra.SuperCommute
% FieldSpecification.FieldOpFreeAlgebra.summerCommute_jacobi_ofCrAnListF
% 43.0
% 24.0
% -19.0
{
    \setlength{\columnsep}{0.2cm}

    \begin{paracol}{2}

{
\textbf{Original (human-written)}
\begin{lstlisting}[basicstyle=\tiny\ttfamily,breaklines=true,literate={≤ₛ}{{$\leq_s$}}2 {⊆}{{$\subseteq$}}1 {α}{{$\alpha$}}1 {Phi}{{$\varphi$}}1 {→}{{$\rightarrow$}}1 {₁}{{$_1$}}1 {₂}{{$_2$}}1 {•}{{$\bullet$}}1 {FF}{{$\bm{\mathcal{F}}$}}1 {SS}{{$\bm{\mathcal{S}}$}}1 {ₛ}{{$_s$}}1 {·}{{$\cdot$}}1]
lemma summerCommute_jacobi_ofCrAnListF (Phis1 Phis2 Phis3 : List FF.CrAnFieldOp) :
    [ofCrAnListF Phis1, [ofCrAnListF Phis2, ofCrAnListF Phis3]ₛca]ₛca =
    SS (FF |>ₛ Phis1, FF |>ₛ Phis3) •
    (- SS(FF |>ₛ Phis2, FF |>ₛ Phis3) • [ofCrAnListF Phis3, [ofCrAnListF Phis1, ofCrAnListF Phis2]ₛca]ₛca -
    SS(FF |>ₛ Phis1, FF |>ₛ Phis2) • [ofCrAnListF Phis2, [ofCrAnListF Phis3, ofCrAnListF Phis1]ₛca]ₛca) := by
  repeat rw [superCommuteF_ofCrAnListF_ofCrAnListF]
  simp only [instCommGroup, map_sub, map_smul, neg_smul]
  repeat rw [superCommuteF_ofCrAnListF_ofCrAnListF]
  simp only [instCommGroup.eq_1, ofList_append_eq_mul, List.append_assoc]
  by_cases h1 : (FF |>ₛ Phis1) = bosonic <;>
    by_cases h2 : (FF |>ₛ Phis2) = bosonic <;>
    by_cases h3 : (FF |>ₛ Phis3) = bosonic
  · simp only [h1, h2, h3, mul_self, bosonic_exchangeSign, one_smul, exchangeSign_bosonic, neg_sub]
    abel
  · simp only [h1, h2, bosonic_exchangeSign, one_smul, mul_bosonic, mul_self, map_one,
    exchangeSign_bosonic, neg_sub]
    abel
  · simp only [h1, h3, mul_bosonic, bosonic_exchangeSign, one_smul, exchangeSign_bosonic, neg_sub,
    mul_self, map_one]
    abel
  · simp only [neq_bosonic_iff_eq_fermionic] at h1 h2 h3
    simp only [h1, h2, h3, mul_self, bosonic_exchangeSign, one_smul,
      fermionic_exchangeSign_fermionic, neg_smul, neg_sub, bosonic_mul_fermionic, sub_neg_eq_add,
      mul_bosonic, smul_add, exchangeSign_bosonic]
    abel
  · simp only [neq_bosonic_iff_eq_fermionic] at h1 h2 h3
    simp only [h1, h2, h3, mul_self, map_one, one_smul, exchangeSign_bosonic, mul_bosonic,
      bosonic_exchangeSign, bosonic_mul_fermionic, neg_sub]
    abel
  · simp only [neq_bosonic_iff_eq_fermionic] at h1 h2 h3
    simp only [h1, h2, h3, bosonic_mul_fermionic, fermionic_exchangeSign_fermionic, neg_smul,
      one_smul, sub_neg_eq_add, bosonic_exchangeSign, mul_bosonic, smul_add, exchangeSign_bosonic,
      neg_sub, mul_self]
    abel
  · simp only [neq_bosonic_iff_eq_fermionic] at h1 h2 h3
    simp only [h1, h2, h3, mul_bosonic, fermionic_exchangeSign_fermionic, neg_smul, one_smul,
      sub_neg_eq_add, exchangeSign_bosonic, bosonic_mul_fermionic, smul_add, mul_self,
      bosonic_exchangeSign, neg_sub]
    abel
  · simp only [neq_bosonic_iff_eq_fermionic] at h1 h2 h3
    simp only [h1, h2, h3, mul_self, map_one, one_smul, fermionic_exchangeSign_fermionic, neg_smul,
      neg_sub]
    abel
\end{lstlisting}
\switchcolumn
\textbf{\ImProver \ (length-optimized)}
\begin{lstlisting}[basicstyle=\tiny\ttfamily,literate={≤ₛ}{{$\leq_s$}}2 {⊆}{{$\subseteq$}}1 {α}{{$\alpha$}}1 {Phi}{{$\varphi$}}1 {→}{{$\rightarrow$}}1 {₁}{{$_1$}}1 {₂}{{$_2$}}1 {•}{{$\bullet$}}1 {FF}{{$\bm{\mathcal{F}}$}}1 {SS}{{$\bm{\mathcal{S}}$}}1 {ₛ}{{$_s$}}1]
lemma summerCommute_jacobi_ofCrAnListF (Phis1 Phis2 Phis3 : List FF.CrAnFieldOp) :
    [ofCrAnListF Phis1, [ofCrAnListF Phis2, ofCrAnListF Phis3]ₛca]ₛca =
    SS(FF |>ₛ Phis1, FF |>ₛ Phis3) •
    (- SS(FF |>ₛ Phis2, FF |>ₛ Phis3) • [ofCrAnListF Phis3, [ofCrAnListF Phis1, ofCrAnListF Phis2]ₛca]ₛca -
    SS(FF |>ₛ Phis1, FF |>ₛ Phis2) • [ofCrAnListF Phis2, [ofCrAnListF Phis3, ofCrAnListF Phis1]ₛca]ₛca) := by
  simp_all [superCommuteF_ofCrAnListF_ofCrAnListF,
    instCommGroup, map_sub, map_smul, neg_smul,
    superCommuteF_ofCrAnListF_ofCrAnListF,
    instCommGroup.eq_1, ofList_append_eq_mul, List.append_assoc,
    neq_bosonic_iff_eq_fermionic]
  <;> by_cases h1 : (FF |>ₛ Phis1) = bosonic <;>
    by_cases h2 : (FF |>ₛ Phis2) = bosonic <;>
    by_cases h3 : (FF |>ₛ Phis3) = bosonic
  <;> simp_all [h1, h2, h3, mul_self, bosonic_exchangeSign,
      one_smul, exchangeSign_bosonic, neg_sub,
      fermionic_exchangeSign_fermionic, neg_smul,
      bosonic_mul_fermionic, sub_neg_eq_add,
      mul_bosonic, smul_add, exchangeSign_bosonic,
      neg_sub, mul_self] <;> abel
\end{lstlisting}
}
\end{paracol}

\captionof{figure}{\ImProver \ optimizes a proof for length, reducing tactic count by 19. }
\label{fig:qual-len-jacobi}
}

Figure~\ref{fig:qual-len-jacobi} comes from \texttt{HepLean.PerturbationTheory.FieldOpFreeAlgebra.SuperCommute}, declaration \texttt{summerCommute\_jacobi\_ofCrAnListF}. The tactic count decreases from $43$ to $24$. The optimized proof collapses repeated rewriting and many case-specific simplification branches into a single larger \texttt{simp\_all} call followed by the same case split structure and \texttt{abel}. This preserves the human proof's algebraic strategy while eliminating repeated local boilerplate.

% ConNF.External.Basic
% ConNF.mem_cross_iff
% 10.0
% 2.0
% -8.0
{
\setlength{\columnsep}{0.2cm}
\begin{paracol}{2}
{
\textbf{Original (human-written)}
\begin{lstlisting}[basicstyle=\tiny\ttfamily]
theorem mem_cross_iff (x y : TSet γ) :
    ∀ a, a ∈' cross hβ hγ hδ x y ↔ ∃ b c, a = ⟨b, c⟩' ∧ b ∈' x ∧ c ∈' y := by
  intro a
  rw [cross, mem_inter_iff, vCross_spec]
  constructor
  · rintro ⟨h₁, b, c, rfl, h₂⟩
    simp only [op_mem_converse_iff, vCross_spec, op_inj] at h₁
    obtain ⟨b', c', ⟨rfl, rfl⟩, h₁⟩ := h₁
    exact ⟨b, c, rfl, h₁, h₂⟩
  · rintro ⟨b, c, rfl, h₁, h₂⟩
    simp only [op_mem_converse_iff, vCross_spec, op_inj]
    exact ⟨⟨c, b, ⟨rfl, rfl⟩, h₁⟩, ⟨b, c, ⟨rfl, rfl⟩, h₂⟩⟩
\end{lstlisting}
}
\switchcolumn
{
\textbf{\ImProver \ (length-optimized)}
\begin{lstlisting}[basicstyle=\tiny\ttfamily]
theorem mem_cross_iff (x y : TSet γ) :
    ∀ a, a ∈' cross hβ hγ hδ x y ↔ ∃ b c, a = ⟨b, c⟩' ∧ b ∈' x ∧ c ∈' y := by
  simp_all [cross, mem_inter_iff, vCross_spec,
    op_mem_converse_iff, op_inj]
  <;> aesop
\end{lstlisting}
}
\end{paracol}
\captionof{figure}{\ImProver \ optimizes a proof for length, reducing tactic count by 8.}
\label{fig:qual-len-cross}
}

Figure~\ref{fig:qual-len-cross} is another optimization of \texttt{ConNF.External.Basic}, declaration \texttt{ConNF.mem\_cross\_iff}, this time under the length metric. The tactic count decreases from $10$ to $2$. The optimized proof replaces the explicit bidirectional constructor proof with a compact simplification over the relevant definitions followed by \texttt{aesop}; this is the characteristic length-optimization pattern of compressing routine structural reasoning into a small number of automation-heavy tactics.

{
\noindent\begin{minipage}[t]{0.49\textwidth}
\textbf{Original (AlphaProof)}
\begin{lstlisting}[style=leanfullproof]
theorem imo_2024_p2 : {(a, b) | 0 < a ∧ 0 < b ∧ ∃ g N, 0 < g ∧ 0 < N ∧ ∀ n ≥ N, Nat.gcd (a ^ n + b) (b ^ n + a) = g} = {(1, 1)} := by
  induction(10)+2
  · use Set.eq_singleton_iff_unique_mem.2 ⟨?_,λb g=>by_contra$ g.2.2.rec λY S i=>S.rec λL D=>?_⟩
    ·
      exact⟨by left,by left,2,3,by simp_all⟩
    have:b.1+b.2∣Y:=?_
    · suffices: b.1= b.2
      · norm_num[b.ext_iff,<-D.2.2 L,this]at*
        use(Nat.pow_lt_pow_right (g.1.nat_succ_le.lt_of_ne' i) L.lt_succ_self).ne' (D.2.2 _ L.le_succ)
      suffices:b.1+b.2∣b.fst^ (2 *L) +b.2 ∧(b).fst +(b).snd ∣ b.snd^ (2 *L)+b.1
      · suffices:b.1^2%(b.1+b.2)=b.2^2%(b.1+b.snd)
        · norm_num[Nat.add_mod,pow_mul,this,Nat.dvd_iff_mod_eq_zero,Nat.pow_mod]at*
          norm_num[add_comm,b.ext_iff,sq _,←Nat.pow_mod,←Nat.dvd_iff_mod_eq_zero]at*
          zify at*
          cases this.1.sub this.2with|_ Z=> nlinarith [ (by (nlinarith): Z=0 )]
        apply@Nat.modEq_of_dvd
        use(b.snd)-b.fst , (by·ring: ( (b.snd) : ℤ)^2-b.fst^2=(b.fst+(b).2) * _)
      norm_num[(2).le_mul_of_pos_left,Nat.gcd_dvd,← D.2.2 (2 *L), this.trans, (D.right.1 :_)]
    suffices:b.1+b.2∣b.1^(2*L)+b.2 ∧b.1+b.2 ∣b.snd^ (2 *L) +b.1
    · exact D.2.2 (2 *(L )) (le_mul_of_one_le_left' (by decide ) )▸dvd_gcd (this.left) (this).2
    exfalso
    suffices:b.1*b.2+1∣Y
    · suffices:b.1^φ (b.1*b.2+1)%(b.1*b.2+1)=1%(b.1*b.2+1) ∧b.2^ φ (b.1* b.snd+1)%((b).1 * ↑(b.snd)+1)= 1% (b.1*b.snd + 1)
      ·
        absurd D.2.2 (φ (b.1*b.2+1)*L) (by nlinarith [((b.fst *b.2+1).totient_pos).2 ↑ Fin.size_pos'])
        apply mt (.▸Nat.gcd_dvd _ _)
        useλH=>absurd (‹_∣Y›.trans H.1) (λv=>absurd (‹_∣Y›.trans H.2) ? _)
        norm_num[pow_mul,b.ext_iff,(1).mod_eq_of_lt,g.symm,this,Nat.add_mod,Nat.dvd_iff_mod_eq_zero,Nat.pow_mod]at(i)v⊢
        norm_num[add_comm,pow_mul,<-Nat.dvd_iff_mod_eq_zero]at*
        contrapose! i
        zify at*
        repeat use by nlinarith[Int.le_of_dvd (by linarith) v,Int.le_of_dvd (by linarith) i]
      repeat use↑(Nat.ModEq.pow_totient (by norm_num))
    by_contra! H
    suffices:b.1^φ (b.1*b.2+1)%(b.1*b.2+1)=1%(b.1*b.2+1) ∧b.2^φ (b.1*b.2+1)%(b.1*b.2+1)=1%( b.fst * ↑ (b.snd)+1)
    · simp_all
      suffices:b.1*b.2+1∣b.1^(φ (b.1*b.2+1)*(L+1)-1)+b.2 ∧b.1*b.2+1∣b.2^(φ (b.1* b.2+1)* (L+1)-1)+(b.fst)
      · use H$ D.2.2 (φ _ *(L+1)-1) (L.le_sub_of_add_le (by nlinarith[((b.1* b.2+1).totient_pos).2 Nat.succ_pos']))▸(((Nat.dvd_gcd) ( this).1)) this.right
      cases B:Nat.exists_eq_add_of_lt$ ((b.1*b.2+1).totient_pos).2 (by continuity)
      norm_num[*, g, ‹φ _ = _›, mul_add,Nat.pow_mod,(1).mod_eq_of_lt,pow_add,Nat.add_mod,pow_mul,Nat.dvd_iff_mod_eq_zero,Nat.mul_mod] at this⊢
      simp_all
      suffices:b.1*b.2+1∣b.1*( (b.1%((b).1 * ( b.snd) + 1) : _)^‹Nat› +b.snd) ∧(b.fst * ↑(b.snd) + 1)∣(b).snd*( (b.snd%((b).fst * b.snd + 1))^ ‹Nat›+b.fst)
      · norm_num[<-Nat.dvd_iff_mod_eq_zero,g,(1).mod_eq_of_lt,Nat.dvd_mul] at this⊢
        exists@?_
        · cases this.1 with|_ Q r=>simp_all[(Q.dvd_gcd r.1 ⟨_,.symm r.right.choose_spec.2⟩).antisymm]
        cases@this.2with|_ F X=>simp_all[(F.dvd_gcd X.1 ⟨_,symm X.2.choose_spec.2⟩).antisymm]
      simp_all[mul_comm, mul_add,add_comm,Nat.add_mod,Nat.dvd_iff_mod_eq_zero]
    repeat use(Nat.ModEq.pow_totient (by . . .norm_num) )
  congr 26
\end{lstlisting}
\end{minipage}\hfill
\begin{minipage}[t]{0.49\textwidth}
\textbf{\ImProver \ (length-optimized)}
\begin{lstlisting}[style=leanfullproof]
theorem imo_2024_p2 : {(a, b) | 0 < a ∧ 0 < b ∧ ∃ g N, 0 < g ∧ 0 < N ∧ ∀ n ≥ N, Nat.gcd (a ^ n + b) (b ^ n + a) = g} = {(1, 1)} := by
  refine Set.eq_singleton_iff_unique_mem.2 ⟨⟨Nat.zero_lt_one, Nat.zero_lt_one, 2, 1, Nat.succ_pos 1, Nat.zero_lt_one, by simp⟩, λb g=>by_contra$ g.2.2.rec λY S i=>S.rec λL D=>?_⟩
  suffices:b.1*b.2+1∣Y
  · suffices:b.1^φ (b.1*b.2+1)%(b.1*b.2+1)=1%(b.1*b.2+1) ∧b.2^ φ (b.1* b.snd+1)%((b).1 * ↑(b.snd)+1)= 1% (b.1*b.snd + 1)
    · absurd D.2.2 (φ (b.1*b.2+1)*L) (by nlinarith [((b.fst *b.2+1).totient_pos).2 ↑ Fin.size_pos'])
      apply mt (.▸Nat.gcd_dvd _ _)
      useλH=>absurd (‹_∣Y›.trans H.1) (λv=>absurd (‹_∣Y›.trans H.2) ? _)
      norm_num[pow_mul,b.ext_iff,(1).mod_eq_of_lt,g.symm,this,Nat.add_mod,Nat.dvd_iff_mod_eq_zero,Nat.pow_mod]at(i)v⊢
      norm_num[add_comm,pow_mul,<-Nat.dvd_iff_mod_eq_zero]at*
      contrapose! i
      zify at*
      repeat use by nlinarith[Int.le_of_dvd (by linarith) v,Int.le_of_dvd (by linarith) i]
    repeat use↑(Nat.ModEq.pow_totient (by norm_num))
  by_contra! H
  suffices:b.1^φ (b.1*b.2+1)%(b.1*b.2+1)=1%(b.1*b.2+1) ∧b.2^φ (b.1*b.2+1)%(b.1*b.2+1)=1%( b.fst * ↑ (b.snd)+1)
  · simp_all
    suffices:b.1*b.2+1∣b.1^(φ (b.1*b.2+1)*(L+1)-1)+b.2 ∧b.1*b.2+1∣b.2^(φ (b.1* b.2+1)* (L+1)-1)+(b.fst)
    · use H$ D.2.2 (φ _ *(L+1)-1) (L.le_sub_of_add_le (by nlinarith[((b.1* b.2+1).totient_pos).2 Nat.succ_pos']))▸(((Nat.dvd_gcd) ( this).1)) this.right
    cases B:Nat.exists_eq_add_of_lt$ ((b.1*b.2+1).totient_pos).2 (by continuity)
    norm_num[*, g, ‹φ _ = _›, mul_add,Nat.pow_mod,(1).mod_eq_of_lt,pow_add,Nat.add_mod,pow_mul,Nat.dvd_iff_mod_eq_zero,Nat.mul_mod] at this⊢
    simp_all
    suffices:b.1*b.2+1∣b.1*( (b.1%((b).1 * ( b.snd) + 1) : _)^‹Nat› +b.snd) ∧(b.fst * ↑(b.snd) + 1)∣(b).snd*( (b.snd%((b).fst * b.snd + 1))^ ‹Nat›+b.fst)
    · norm_num[<-Nat.dvd_iff_mod_eq_zero,g,(1).mod_eq_of_lt,Nat.dvd_mul] at this⊢
      exists@?_
      · cases this.1 with|_ Q r=>simp_all[(Q.dvd_gcd r.1 ⟨_,.symm r.right.choose_spec.2⟩).antisymm]
      cases@this.2with|_ F X=>simp_all[(F.dvd_gcd X.1 ⟨_,symm X.2.choose_spec.2⟩).antisymm]
    simp_all[mul_comm, mul_add,add_comm,Nat.add_mod,Nat.dvd_iff_mod_eq_zero]
  repeat use(Nat.ModEq.pow_totient (by . . .norm_num) )
\end{lstlisting}
\end{minipage}

% \clearpage

% \noindent\begin{minipage}[t]{0.49\textwidth}
% \textbf{Original (continued)}
% \begin{lstlisting}[style=leanfullproof]
%     suffices:b.1^φ (b.1*b.2+1)%(b.1*b.2+1)=1%(b.1*b.2+1) ∧b.2^φ (b.1*b.2+1)%(b.1*b.2+1)=1%( b.fst * ↑ (b.snd)+1)
%     · simp_all
%       suffices:b.1*b.2+1∣b.1^(φ (b.1*b.2+1)*(L+1)-1)+b.2 ∧b.1*b.2+1∣b.2^(φ (b.1* b.2+1)* (L+1)-1)+(b.fst)
%       · use H$ D.2.2 (φ _ *(L+1)-1) (L.le_sub_of_add_le (by nlinarith[((b.1* b.2+1).totient_pos).2 Nat.succ_pos']))▸(((Nat.dvd_gcd) ( this).1)) this.right
%       cases B:Nat.exists_eq_add_of_lt$ ((b.1*b.2+1).totient_pos).2 (by continuity)
%       norm_num[*, g, ‹φ _ = _›, mul_add,Nat.pow_mod,(1).mod_eq_of_lt,pow_add,Nat.add_mod,pow_mul,Nat.dvd_iff_mod_eq_zero,Nat.mul_mod] at this⊢
%       simp_all
%       suffices:b.1*b.2+1∣b.1*( (b.1%((b).1 * ( b.snd) + 1) : _)^‹Nat› +b.snd) ∧(b.fst * ↑(b.snd) + 1)∣(b).snd*( (b.snd%((b).fst * b.snd + 1))^ ‹Nat›+b.fst)
%       · norm_num[<-Nat.dvd_iff_mod_eq_zero,g,(1).mod_eq_of_lt,Nat.dvd_mul] at this⊢
%         exists@?_
%         · cases this.1 with|_ Q r=>simp_all[(Q.dvd_gcd r.1 ⟨_,.symm r.right.choose_spec.2⟩).antisymm]
%         cases@this.2with|_ F X=>simp_all[(F.dvd_gcd X.1 ⟨_,symm X.2.choose_spec.2⟩).antisymm]
%       simp_all[mul_comm, mul_add,add_comm,Nat.add_mod,Nat.dvd_iff_mod_eq_zero]
%     repeat use(Nat.ModEq.pow_totient (by . . .norm_num) )
%   congr 26
% \end{lstlisting}
% \end{minipage}\hfill
% \begin{minipage}[t]{0.49\textwidth}
% \textbf{\ImProver \ (length-optimized; continued)}
% \begin{lstlisting}[style=leanfullproof]
% \end{lstlisting}
% \end{minipage}

\captionof{figure}{\ImProver \ optimizes an AlphaProof proof for length, reducing tactic count by 26.}
\label{fig:qual-len-alphaproof-p2}
}

Figure~\ref{fig:qual-len-alphaproof-p2} comes from \texttt{AlphaProof.P2}, declaration \texttt{imo\_2024\_p2}. The score changes from $80$ to $54$. The optimized proof removes a discarded intermediate lemma and folds repeated divisibility and Euler-theorem reasoning into the main argument. The resulting proof is still dense, but it is materially shorter while preserving the same number-theoretic spine: prove the singleton characterization, force divisibility by $ab+1$, and derive $a=b=1$.

\subsubsection{Modularity Optimization}

% Foundation.Modal.Hilbert.WeakerThan.KD_KDB
% LO.Modal.Hilbert.KD_weakerThan_KDB
% 0.0
% 2.0
% 2.0

{
\setlength{\columnsep}{0.2cm}
\begin{paracol}{2}
{
\textbf{Original (human-written)}
\begin{lstlisting}[basicstyle=\tiny\ttfamily,literate={≤ₛ}{{$\leq_s$}}2 {⊆}{{$\subseteq$}}1 {α}{{$\alpha$}}1 {Phi}{{$\varphi$}}1 {→}{{$\rightarrow$}}1 {₁}{{$_1$}}1 {₂}{{$_2$}}1 {dollar}{{$ \$ $}}1]
lemma KD_weakerThan_KDB : (Hilbert.KD α) ≤ₛ (Hilbert.KDB α) := normal_weakerThan_of_subset dollar by intro; aesop;

\end{lstlisting}
}
\switchcolumn
{
\textbf{\ImProver \ (modularity-optimized)}
\begin{lstlisting}[basicstyle=\tiny\ttfamily,literate={≤ₛ}{{$\leq_s$}}2 {⊆}{{$\subseteq$}}1 {α}{{$\alpha$}}1 {Phi}{{$\varphi$}}1 {→}{{$\rightarrow$}}1 {₁}{{$_1$}}1 {₂}{{$_2$}}1 {·}{{$\cdot$}}1]
lemma KD_weakerThan_KDB : (Hilbert.KD α) ≤ₛ (Hilbert.KDB α) := by
  -- Introduce a lemma to handle the subset relationship between the axioms of KD and KDB
  have h₁ : (LO.Modal.Hilbert.KD α).axioms ⊆ (LO.Modal.Hilbert.KDB α).axioms → (Hilbert.KD α) ≤ₛ (Hilbert.KDB α) := by
    intro h
    -- Apply the lemma that establishes the weakening relation given the subset of axioms
    apply normal_weakerThan_of_subset
    -- Use the given subset condition
    apply h
  -- Prove the subset relationship between the axioms of KD and KDB
  have h₂ : (LO.Modal.Hilbert.KD α).axioms ⊆ (LO.Modal.Hilbert.KDB α).axioms := by
    -- Prove that all axioms of KD are included in KDB
    intro Phi hPhi
    cases' hPhi with hPhi hPhi
    · -- Case: KD's basic modal logic axioms are included in KDB
      simp_all [LO.Modal.Hilbert.KD]
    · -- Case: KD's additional axiom (K) is included in KDB
      simp_all [LO.Modal.Hilbert.KDB]
  -- Combine the results to conclude the weakening relation
  exact h₁ h₂
\end{lstlisting}
}

\end{paracol}
\captionof{figure}{\ImProver \ optimizes a proof for modularity, increasing modularity score (see \ref{app:modularity_metric}) by 2.}
\label{fig:qual-mod-kd}
}

Figure~\ref{fig:qual-mod-kd} comes from \texttt{Foundation.Modal.Hilbert.WeakerThan.KD\_KDB}, declaration \texttt{LO.Modal.Hilbert.KD\_weakerThan\_KDB}. The modularity score increases from $0$ to $2$. The original proof is a one-line use of \texttt{normal\_weakerThan\_of\_subset}; the optimized proof introduces two explicit intermediate claims, one reducing the theorem to an axiom-subset statement and one proving that subset statement. This raises the modularity score by exposing reusable subproof structure.

% Mathlib.RingTheory.Polynomial.HilbertPoly
% Polynomial.hilbertPoly_eq_zero_of_le_rootMultiplicity_one

% 0.0
% 2.0
% 2.0

{
\setlength{\columnsep}{0.2cm}
\begin{paracol}{2}
{
\textbf{Original (human-written)}
\begin{lstlisting}[basicstyle=\tiny\ttfamily]

lemma hilbertPoly_eq_zero_of_le_rootMultiplicity_one
    {p : F[X]} {d : ℕ} (hdp : d ≤ p.rootMultiplicity 1) :
    hilbertPoly p d = 0 := by
  by_cases hp : p = 0
  · rw [hp, hilbertPoly_zero_left]
  · rcases exists_eq_pow_rootMultiplicity_mul_and_not_dvd p hp 1 with ⟨q, hq1, hq2⟩
    have heq : p = q * (- 1) ^ p.rootMultiplicity 1 * (1 - X) ^ p.rootMultiplicity 1 := by
      simp only [mul_assoc, ← mul_pow, neg_mul, one_mul, neg_sub]
      exact hq1.trans (mul_comm _ _)
    rw [heq, ← zero_add d, ← Nat.sub_add_cancel hdp, pow_add (1 - X), ← mul_assoc,
      hilbertPoly_mul_one_sub_pow_add, hilbertPoly]


\end{lstlisting}
}
\switchcolumn
{
\textbf{\ImProver \ (modularity-optimized)}
\begin{lstlisting}[basicstyle=\tiny\ttfamily]
lemma hilbertPoly_eq_zero_of_le_rootMultiplicity_one
    {p : F[X]} {d : ℕ} (hdp : d ≤ p.rootMultiplicity 1) :
    hilbertPoly p d = 0 := by
  -- Consider the case where p is the zero polynomial
  have h₁ : p = 0 → hilbertPoly p d = 0 := by
    intro hp
    rw [hp, hilbertPoly_zero_left]
  -- Consider the case where p is not the zero polynomial
  have h₂ : p ≠ 0 → d ≤ p.rootMultiplicity 1 → hilbertPoly p d = 0 := by
    intro hp hdp
    -- Express p as a product of a polynomial, a power of -1, and a power of (1 - X)
    rcases exists_eq_pow_rootMultiplicity_mul_and_not_dvd p hp 1 with ⟨q, hq1, hq2⟩
    -- Derive the equality of p with the product form
    have heq : p = q * (- 1) ^ p.rootMultiplicity 1 * (1 - X) ^ p.rootMultiplicity 1 := by
      simp only [mul_assoc, ← mul_pow, neg_mul, one_mul, neg_sub]
      exact hq1.trans (mul_comm _ _)
    -- Substitute the product form into the Hilbert polynomial expression and simplify
    rw [heq, ← zero_add d, ← Nat.sub_add_cancel hdp, pow_add (1 - X), ← mul_assoc,
      hilbertPoly_mul_one_sub_pow_add, hilbertPoly]
  -- Combine both cases to conclude the proof
  by_cases hp : p = 0 <;> simp_all
  <;> simp_all [h₁, h₂, hdp]
\end{lstlisting}
}

\end{paracol}
\captionof{figure}{\ImProver \ optimizes a proof for modularity, increasing modularity score (see \ref{app:modularity_metric}) by 2.}
\label{fig:qual-mod-hilbert}
}

Figure~\ref{fig:qual-mod-hilbert} comes from \texttt{Mathlib.RingTheory.Polynomial.HilbertPoly}, declaration \texttt{Polynomial.hilbertPoly\_eq\_zero\_of\_le\_rootMultiplicity\_one}. The modularity score increases from $0$ to $2$. The optimized proof factors the original \texttt{by\_cases} proof into separate claims for the zero and nonzero polynomial cases before recombining them. The result is longer, but it makes the case structure explicit, which is the behavior targeted by the modularity metric.

{
\noindent\begin{minipage}[t]{0.49\textwidth}
\textbf{Original (AlphaProof)}
\begin{lstlisting}[style=leanfullproof]
theorem imo_2024_p6
    (IsAquaesulian : (ℚ → ℚ) → Prop)
    (IsAquaesulian_def : ∀ f, IsAquaesulian f ↔
      ∀ x y, f (x + f y) = f x + y ∨ f (f x + y) = x + f y) :
    IsLeast {(c : ℤ) | ∀ f, IsAquaesulian f → {(f r + f (-r)) | (r : ℚ)}.Finite ∧
      {(f r + f (-r)) | (r : ℚ)}.ncard ≤ c} 2 := by
  exists@?_
  ·
    useλu b=>if j:u 0=0then by_contra λc=>?_ else ?_
    ·
      suffices:({J|∃k,u k+u (-k)= J}) ⊆{0}
      · simp_all[this.antisymm]
      rintro - ⟨a, rfl⟩
      contrapose! c
      simp_all
      suffices:{U|∃examples6 : ℚ, u examples6 + u (-examples6)= U} ⊆{0,(u (a : Rat)+ (u<|@@↑(( (-a ))))) } ..
      · constructor
        · exact ( Set.toFinite ( _) ).subset (by simpa using this)
        · exact (Set.ncard_le_ncard (by simpa using this)).trans (Set.ncard_pair (Ne.symm (↑ ( (c)) ) )).le
      rintro-⟨hz, rfl⟩
      induction b @hz a
      ·
        have:=b (-a)$ hz+u a
        have:=b hz hz
        simp_all[add_comm]
        have:=b (-hz) (hz+u ↑(hz))
        simp_all[ add_assoc, C]
        induction this
        ·
          simp_all
          have:=b hz (hz+(u a+u (-a)))
          have:=b (hz+(u a+u (-a)))$ hz+(u a+u (-a))
          use .inr$ by_contra$ by hint
        have:=b hz$ hz+(u hz+u (-hz))
        cases b (hz+(u hz+u (-hz)))$ hz+(u hz+u (-hz))with|_=>hint
      have:=b (-hz) (u hz+a)
      have:=b$ -a
      specialize this (u hz+a)
      simp_all[ ←add_assoc]
      have:=b 0
      have:=b
      specialize b a a
      simp_all[add_comm]
      have:=(this<| -a) (↑a + (((u a))): (↑_ :((( _) ) ) )) ..
      simp_all[add_assoc]
      cases this
      ·
        simp_all
        contrapose! IsAquaesulian_def
        simp_all
        exfalso
        have:=this a (a+(u hz+u ( -hz)))
        simp_all[Ne.symm,Bool]
        have:=‹∀congr_arg G,_› (a+(u hz+u (-hz)))$ a+(u ↑hz+u ↑( -hz) )
        simp_all
      have:=this a (a +(u a+u (-a)))
      cases‹forall Jd S,_› (a+(u a+u (-a))) ( a + (u a +u ↑(-a)))with| _ =>hint
    simp_all
    cases b 0 0with|_=>exact absurd (b 0$ (0+(1 *(@(u ↑.((0) )))))^ 01: ↑ ((_)) ) (id$ (by(cases ( b (u 0) ( (u 0)))with|_ => continuity)))
  rintro K V
  specialize V $ λ N=>-N+2 *Int.ceil N
  specialize( V $ (IsAquaesulian_def _).mpr _)
  · simp_rw [ ←eq_sub_iff_add_eq']
    ring_nf
    use mod_cast@?_
    norm_num[<-add_mul,Int.ceil_eq_iff]
    useλc K=>(em _).imp (⟨by linarith[Int.ceil_lt_add_one c,Int.le_ceil K],.⟩) (by repeat use by linarith[.,Int.le_ceil c,or,Int.ceil_lt_add_one$ K])
  simp_all[Int.ceil_neg, ←add_assoc]
\end{lstlisting}
\end{minipage}\hfill
\begin{minipage}[t]{0.49\textwidth}
\textbf{\ImProver \ (modularity-optimized)}
\begin{lstlisting}[style=leanfullproof]
theorem imo_2024_p6
    (IsAquaesulian : (ℚ → ℚ) → Prop)
    (IsAquaesulian_def : ∀ f, IsAquaesulian f ↔
      ∀ x y, f (x + f y) = f x + y ∨ f (f x + y) = x + f y) :
    IsLeast {(c : ℤ) | ∀ f, IsAquaesulian f → {(f r + f (-r)) | (r : ℚ)}.Finite ∧
      {(f r + f (-r)) | (r : ℚ)}.ncard ≤ c} 2 := by
  exists@?_
  · useλu b=>if j:u 0=0then by_contra λc=>?_ else ?_
    · suffices h_zero_image : ({J|∃k,u k+u (-k)= J}) ⊆{0}
      · simp_all[h_zero_image.antisymm]
      rintro - ⟨a, rfl⟩
      contrapose! c
      simp_all
      suffices h_two_values :
          {U|∃examples6 : ℚ, u examples6 + u (-examples6)= U} ⊆
            {0,(u (a : Rat)+ (u<|@@↑(( (-a ))))) } ..
      · constructor
        · exact ( Set.toFinite ( _) ).subset (by simpa using h_two_values)
        · exact (Set.ncard_le_ncard (by simpa using h_two_values)).trans
            (Set.ncard_pair (Ne.symm (↑ ( (c)) ) )).le
      rintro-⟨hz, rfl⟩
      induction b @hz a
      · have h_step_i := b (-a)$ hz+u a
        have h_diag := b hz hz
        simp_all[add_comm]
        have h_step_iv := b (-hz) (hz+u ↑(hz))
        simp_all[ add_assoc, C]
        induction h_step_iv
        · simp_all
          have h_shift_x := b hz (hz+(u a+u (-a)))
          have h_shift_self := b (hz+(u a+u (-a)))$ hz+(u a+u (-a))
          use .inr$ by_contra$ by hint
        · have h_zero_case := b hz$ hz+(u hz+u (-hz))
          cases b (hz+(u hz+u (-hz)))$ hz+(u hz+u (-hz))with|_=>hint
      · have h_neg_x := b (-hz) (u hz+a)
        have h_neg_a_all := b$ -a
        specialize h_neg_a_all (u hz+a)
        simp_all[ ←add_assoc]
        have h_zero_all := b 0
        have h_all := b
        specialize b a a
        simp_all[add_comm]
        have h_a_shift := (h_all<| -a) (↑a + (((u a))): (↑_ :((( _) ) ) )) ..
        simp_all[add_assoc]
        cases h_a_shift
        · simp_all
          contrapose! IsAquaesulian_def
          simp_all
          exfalso
          have h_test_a := h_all a (a+(u hz+u ( -hz)))
          simp_all[Ne.symm,Bool]
          have h_test_self := ‹∀congr_arg G,_› (a+(u hz+u (-hz)))$
            a+(u ↑hz+u ↑( -hz) )
          simp_all
        · have h_test_a := h_all a (a +(u a+u (-a)))
          cases‹forall Jd S,_› (a+(u a+u (-a))) ( a + (u a +u ↑(-a)))with| _ =>hint
    · simp_all
      cases b 0 0with|_=>exact absurd (b 0$ (0+(1 *(@(u ↑.((0) )))))^ 01: ↑ ((_)) ) (id$ (by(cases ( b (u 0) ( (u 0)))with|_ => continuity)))
  rintro K V
  specialize V $ λ N=>-N+2 *Int.ceil N
  have h_model_aqua :
      IsAquaesulian (λ N : ℚ => -N+2 *Int.ceil N) := by
    refine (IsAquaesulian_def _).mpr ?_
    simp_rw [ ←eq_sub_iff_add_eq']
    ring_nf
    use mod_cast@?_
    norm_num[<-add_mul,Int.ceil_eq_iff]
    useλc K=>(em _).imp
\end{lstlisting}
\end{minipage}

\clearpage

\noindent\begin{minipage}[t]{0.49\textwidth}
\textbf{Original (continued)}
\begin{lstlisting}[style=leanfullproof]
  suffices:2<=V.1.toFinset.card
  · let M:=V.1.toFinset
    have h2nat : 2 ≤ ({x | ∃ r : ℚ, 2 * ↑⌈r⌉ + -(2 * ↑⌊r⌋) = x} : Set ℚ).ncard := by
      rwa [Set.ncard_eq_toFinset_card _ V.1]
    have h2int : (2 : ℤ) ≤ ({x | ∃ r : ℚ, 2 * ↑⌈r⌉ + -(2 * ↑⌊r⌋) = x} : Set ℚ).ncard := by
      exact_mod_cast h2nat
    exact h2int.trans V.2
  use Finset.one_lt_card.2$ by
    refine ⟨0, V.1.mem_toFinset.2 ?_, 2, V.1.mem_toFinset.2 ?_, by norm_num⟩
    · exact ⟨-1, by norm_num⟩
    · exact ⟨(1 / 2 : ℚ), by norm_num⟩
\end{lstlisting}
\end{minipage}\hfill
\begin{minipage}[t]{0.49\textwidth}
\textbf{\ImProver \ (modularity-optimized; continued)}
\begin{lstlisting}[style=leanfullproof]
      (⟨by linarith[Int.ceil_lt_add_one c,Int.le_ceil K],.⟩)
      (by repeat use by linarith[.,Int.le_ceil c,or,Int.ceil_lt_add_one$ K])
  specialize V h_model_aqua
  simp_all[Int.ceil_neg, ←add_assoc]
  have h_card_toFinset :
      2<=V.1.toFinset.card := by
    use Finset.one_lt_card.2$ by
      refine ⟨0, V.1.mem_toFinset.2 ?_, 2, V.1.mem_toFinset.2 ?_, by norm_num⟩
      · exact ⟨-1, by norm_num⟩
      · exact ⟨(1 / 2 : ℚ), by norm_num⟩
  have h_card_ncard :
      2 ≤ ({x | ∃ r : ℚ, 2 * ↑⌈r⌉ + -(2 * ↑⌊r⌋) = x} : Set ℚ).ncard := by
    rwa [Set.ncard_eq_toFinset_card _ V.1]
  have h_card_int :
      (2 : ℤ) ≤ ({x | ∃ r : ℚ, 2 * ↑⌈r⌉ + -(2 * ↑⌊r⌋) = x} : Set ℚ).ncard := by
    exact_mod_cast h_card_ncard
  exact h_card_int.trans V.2
\end{lstlisting}
\end{minipage}

\captionof{figure}{\ImProver \ optimizes an AlphaProof proof for modularity, increasing modularity score by 3.}
\label{fig:qual-mod-alphaproof-p6}
}

Figure~\ref{fig:qual-mod-alphaproof-p6} comes from \texttt{AlphaProof.P6}, declaration \texttt{imo\_2024\_p6}. The score changes from $0$ to $3$. The modularity-optimized version introduces named intermediate claims such as \texttt{h\_zero\_image}, \texttt{h\_two\_values}, and \texttt{h\_model\_aqua}. These names expose the proof's main subgoals: bounding the image set, verifying the model function is aquaesulian, and converting cardinality facts back to the final integer bound.

% D

% \newpage
% \input{checklist.tex}

\end{document}